\documentclass[12pt, a4paper]{report}
\usepackage[a4paper, total={6.25in, 8.25in}]{geometry} %%%%%% CHECK MARGIN REQ. 8.25 in?
\usepackage[T1]{fontenc}
\usepackage[latin1]{inputenc}
\usepackage[english]{babel}
\usepackage{siunitx}
\usepackage{graphicx}
\usepackage{tipa} % for the \ark{} command
\usepackage{graphics} % for pdf, bitmapped graphics files
\usepackage{amsmath}
\usepackage{latexsym}
\usepackage{bbm}
\usepackage{amscd}% for commutative diagrams
\usepackage{mathrsfs} %this package is for the script font \mathscr
\usepackage{relsize}
\usepackage{delarray}

\usepackage{graphicx} % for youtube image link

\usepackage{datatool}% Sorted list http://ctan.org/pkg/datatool

\usepackage{xcolor} % highlight text 
\usepackage[nottoc]{tocbibind} % Adds bibliography to TOC
\usepackage{glossaries}
\usepackage{stmaryrd}
\usepackage{pstricks}
\usepackage{theorem}
\usepackage{changepage}
\usepackage{euscript}
\usepackage{textcomp}
\usepackage{esvect}
\usepackage{parskip}
\usepackage{placeins}
\usepackage{subcaption}
\usepackage{array}
\usepackage{delarray}
\usepackage{stmaryrd}
\usepackage{fancyhdr}
\usepackage{graphpap}
\usepackage{makeidx}
\usepackage{enumerate}
\usepackage{esint}
\usepackage{datetime}
\usepackage{caption}
\usepackage{smartdiagram}
\usepackage{amssymb} % for math symbols and environments

\newtheorem{definition}{Definition}
\usesmartdiagramlibrary{additions}
\usepackage{abstract}
\usepackage{color}
\definecolor{igreen}{rgb}{0.0, 0.56, 0.0}
\usepackage{xcolor, colortbl}
\colorlet{gred}{-red!75!green!65!}
\colorlet{mamber}{-red!75!green!15!blue!50!}
\colorlet{grown}{-red!75!blue!20!green}
\colorlet{bled}{-red!85!blue!40!green!45!}
\colorlet{waters}{cyan!25} % Define color for the water
\colorlet{water}{cyan!25!green!20!} % Define color for the water
\definecolor{grin}{HTML}{00F9DE}
\usepackage{rotating}
\providecommand{\keywords}[1]{\textbf{\textit{Keywords---}} #1}

\usepackage{arydshln}
\usepackage{booktabs}
\usepackage{graphicx}
\usepackage{tikz}
\usepackage{tikz-3dplot}
\usetikzlibrary{
arrows,
arrows.meta,
automata,
backgrounds,
calc,
decorations,
decorations.pathmorphing,
decorations.pathreplacing,
decorations.fractals,
external,
fit,
matrix,
petri,
positioning,
shadows,
shapes,
shapes.multipart,
topaths,
intersections
}
\usepackage{eso-pic}
\def\ba{\begin{array}}
\def\ea{\end{array}}
\def\beann{\begin{eqnarray*}}
\def\eeann{\end{eqnarray*}}
\def\bea{\begin{eqnarray}}
\def\eea{\end{eqnarray}}

\makeatletter
\newlength\qvec@height
\newlength\qvec@depth
\newlength\qvec@width
\newcommand{\qvec}[2][]{
    \settoheight{\qvec@height}{$#2$}
    \settodepth{\qvec@depth}{$#2$}
    \settowidth{\qvec@width}{$#2$}
  \def\qvec@arg{#1}
  \raisebox{.2ex}{\raisebox{\qvec@height}{\rlap{% 
    \kern.05em
    \begin{tikzpicture}[scale=1,shorten >=-3pt,shorten <=-3pt]
    \pgfsetroundcap
    \coordinate (Stx) at (.05em,0) ;
		\coordinate (Arx) at (\qvec@width-.05em,0) ;
    \draw[->](Stx) to[bend left] (Arx);
    \end{tikzpicture}
  }}}
  #2
}
\makeatother
\makeatletter
\newlength\pvec@height
\newlength\pvec@depth
\newlength\pvec@width
\newcommand{\pvec}[2][]{
    \settoheight{\pvec@height}{$#2$}
    \settodepth{\pvec@depth}{$#2$}
    \settowidth{\pvec@width}{$#2$}
  \def\pvec@arg{#1}
  \raisebox{.2ex}{\raisebox{\pvec@height}{\rlap{% 
    \kern.05em
    \begin{tikzpicture}[scale=1,shorten >=-3pt,shorten <=-3pt]
    \pgfsetroundcap
    \coordinate (Stx) at (.05em,0) ;
		\coordinate (Arx) at (\pvec@width-.05em,0) ;
    \draw[->](Stx) to[bend right] (Arx);
    \end{tikzpicture}
  }}}
  #2
}
\makeatother
\makeatletter
\newlength\vvec@height%
\newlength\vvec@depth%
\newlength\vvec@width%
\newcommand{\vvec}[2][]{%
  \ifmmode%
    \settoheight{\vvec@height}{$#2$}%
    \settodepth{\vvec@depth}{$#2$}%
    \settowidth{\vvec@width}{$#2$}%
  \else 
    \settoheight{\vvec@height}{#2}%
    \settodepth{\vvec@depth}{#2}%
    \settowidth{\vvec@width}{#2}%
  \fi%
  \def\vvec@arg{#1}%
  \def\vvec@dd{:}%
  \def\vvec@d{.}%
  \raisebox{.2ex}{\raisebox{\vvec@height}{\rlap{%
    \kern.05em%
    \begin{tikzpicture}[scale=1]
    \pgfsetroundcap
    \draw (.05em,0)--(\vvec@width-.05em,0);
    \draw (\vvec@width-.05em,0)--(\vvec@width-.15em, .075em);
    \draw (\vvec@width-.05em,0)--(\vvec@width-.15em,-.075em);
    \ifx\vvec@arg\vvec@d%
      \fill(\vvec@width*.45,.5ex) circle (.5pt);%
    \else\ifx\vvec@arg\vvec@dd%
      \fill(\vvec@width*.30,.5ex) circle (.5pt);%
      \fill(\vvec@width*.65,.5ex) circle (.5pt);%
    \fi\fi%
    \end{tikzpicture}%
  }}}%
  #2%
}
\makeatother
\def\ba{\begin{array}}
\def\ea{\end{array}}
\def\beann{\begin{eqnarray*}}
\def\eeann{\end{eqnarray*}}
\def\bea{\begin{eqnarray}}
\def\eea{\end{eqnarray}}

\usepackage{titlesec}
\usepackage{multirow}
\usepackage{hyperref}
\usepackage{lipsum}
\usepackage{tikz-cd}
\usepackage{float}
\usepackage{titling}
\usepackage{epigraph}
\usepackage[title, titletoc]{appendix}
\titleformat{\chapter}{\normalfont\LARGE}{\thechapter\,$\vert$}{20pt}{\LARGE}{\setcounter{chapter}{0}}
\titlespacing*{\chapter}{0pt}{-70pt}{40pt} %Move chapter titles up
\makeatletter
\newcommand\BackgroundPicturea[3]{
	\setlength{\unitlength}{1pt}
	\put(0,\strip@pt\paperheight){
		\parbox[t]{\paperwidth}{
			\vspace{#2}\hspace{#3}
			\mbox{\includegraphics[scale=0.5]{#1}}
}}}
\newcommand\BackgroundPictureb[3]{
	\setlength{\unitlength}{1pt}
	\put(0,\strip@pt\paperheight){
		\parbox[t]{\paperwidth}{
			\vspace{#2}\hspace{#3}
			\mbox{\includegraphics[scale=0.3]{#1}}
}}}
\makeatother
\usepackage{setspace}
\addto\captionsenglish{
	\renewcommand{\contentsname}%
	{Table of Contents}
	\setcounter{tocdepth}{3}% Include \subsubsection in ToC
	\setcounter{secnumdepth}{3}% Number \subsubsection in ToC
	}

\usepackage{xcolor}
\usepackage{listings}

\definecolor{codegreen}{rgb}{0,0.6,0}
\definecolor{codegray}{rgb}{0.5,0.5,0.5}
\definecolor{codepurple}{rgb}{0.58,0,0.82}
\definecolor{backcolour}{rgb}{0.95,0.95,0.92}

\usepackage{amsfonts}

\newcommand{\sortitem}[1]{%
  \DTLnewrow{list}% Create a new entry
  \DTLnewdbentry{list}{description}{#1}% Add entry as description
}
\newenvironment{sortedlist}{%
  \DTLifdbexists{list}{\DTLcleardb{list}}{\DTLnewdb{list}}% Create new/discard old list
}{%
  \DTLsort{description}{list}% Sort list
  \begin{itemize}%
    \DTLforeach*{list}{\theDesc=description}{%
      \item[] \theDesc}% Print each item, remove [] for bullet
  \end{itemize}%
}
\usepackage{algorithm}
\usepackage{algpseudocode}

\usepackage{pgfgantt}

\usepackage{makecell} %multiline in tables

\usepackage{lscape}
\makeatletter
\renewcommand{\ALG@beginalgorithmic}{\footnotesize}
\makeatother

\usepackage{multicol}
\hypersetup{pdftitle = Project Report, pdfauthor = {First Last}, pdfstartview=FitH, pdfkeywords = essay, pdfpagemode = FullScreen, colorlinks, anchorcolor = black, citecolor = blue, urlcolor = blue, filecolor = green, linkcolor = blue, plainpages = false}
\fancyheadoffset{6pt}
\fancyfootoffset{6pt}
\date{September 2024}

\title{\textbf{Multi-Objective Bayesian Optimization with Independent Tanimoto Kernel Gaussian Processes for Diverse Pareto Front Exploration}}
\author{\\ \Large{Anabel Yong (ID: 23205123)}
\\ Supervisors: Professor Brooks Paige, UCL
\\ \& Dr. Austin Tripp, Cambrige Machine Learning Group 
\\ \& Dr. Layla-Hosseini Gerami @ IgnotaLabs.AI
\\ Faculty of Engineering
\\ Department of Computer Science
\\ 
\\
\\ University College London
\\
A Project Report Presented in Partial Fulfillment of the Degree \\ \textit{MSc Computational Statistics and Machine Learning}
\\ \\
}
\begin{document}
%Adjust logo positions here
% \AddToShipoutPicture*{\parbox[t][\paperheight][t]{\paperwidth}{%
%           \includegraphics[width=\paperwidth]{\BackgroundPicturea{Logos/ucl_long%_logo.png}{3in}{3in}}
%           }}
% \AddToShipoutPicture*{\centering\BackgroundPictureb{Logos/Bentham2011_065_c623d.jpg}{3in}{3.7in}}
\AddToShipoutPictureBG*{%
  \AtPageUpperLeft{%
    \raisebox{-\height}{%
      \includegraphics[width=\paperwidth]{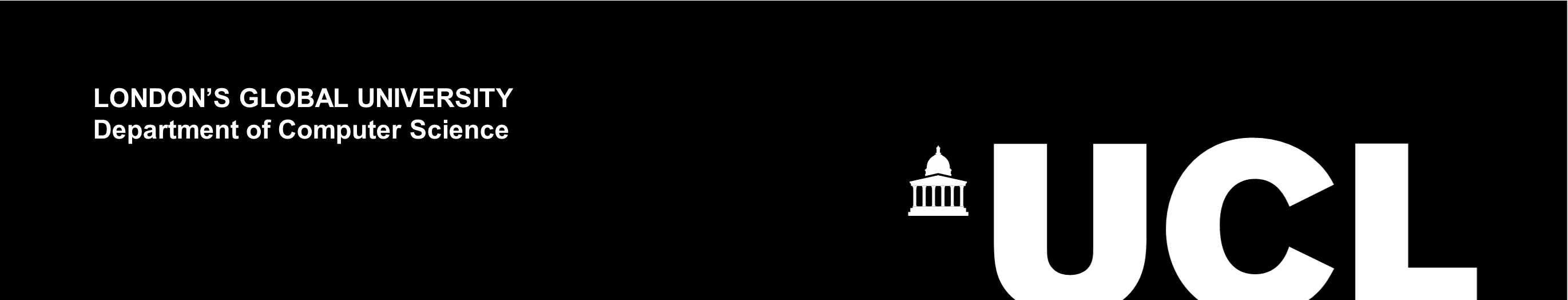}%
    }}
}
\AddToShipoutPicture*{%
      \parbox[t][\paperheight][t]{\paperwidth}{%
          \includegraphics[width=1.2\paperwidth]{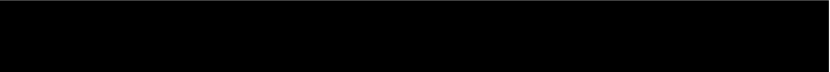}
      }}

\thispagestyle{headings}
\maketitle
\FloatBarrier
\pagenumbering{roman}

\thispagestyle{empty}
\begin{abstract}
We present GP-MOBO, a novel multi-objective Bayesian Optimization algorithm that advances the state-of-the-art in molecular optimization. Our approach integrates a fast minimal package for Exact Gaussian Processes (GPs) capable of efficiently handling the full dimensionality of sparse molecular fingerprints without the need for extensive computational resources. GP-MOBO consistently outperforms traditional methods like GP-BO by fully leveraging fingerprint dimensionality, leading to the identification of higher-quality and valid SMILES. Moreover, our model achieves a broader exploration of the chemical search space, as demonstrated by its superior proximity to the Pareto front in all tested scenarios. Empirical results from the DockSTRING dataset reveal that GP-MOBO yields higher geometric mean values across 20 Bayesian optimization iterations, underscoring its effectiveness and efficiency in addressing complex multi-objective optimization challenges with minimal computational overhead.

% \vspace{-10mm} %To remove added white space after
\end{abstract}

\thispagestyle{empty}
\chapter*{Acknowledgements}
I want to thank my supervisors, Layla-Hosseini Gerami and Brooks Paige for their academic support and pushing me towards academic excellence, and the highest quality of guidance. Thank you Brooks for giving me the ideas for our novel GP-MOBO model. I want to thank Layla for the great emotional, and chemistry support, the Ignota team, and finally, for introducing me to Austin Tripp. My appreciation for Austin Tripp, my external supervisor, spans from his extreme patience regarding this project to his immense knowledge in the realm of drug discovery, Gaussian Processes, and advanced topics in kernel methods. Thank you for making me code everything from scratch. This would not have been a feasible project without Austin. Thank you to the 3 of you for pushing me to my academic limits and being the best supervisors I could ask for during my time at UCL. 

Finally, I dedicate this to Jonathan, who showed me I was capable of doing difficult projects, and for relentlessly believing and supporting me throughout this whole journey this year - even when I felt like giving up. Thank you for being there for me when I am not capable of taking care of myself and making me ramen all the time. Additionally, I want to thank my parents for giving me this opportunity to pursue my degree and providing unconditional support and love despite being so far away. 
\thispagestyle{empty}
\chapter*{Declaration}
I, Anabel Yong,  I declare that the thesis has been composed by myself and that the work has not be submitted for any other degree or professional qualification. I confirm that the work submitted is my own, except where work which has formed part of jointly-authored publications has been included and referenced. The report may be freely copied and distributed provided the source is explicitly acknowledged. \\
%Add signature png here.
% \begin{figure}[H]
% \includegraphics{Logos/signature.PNG}
% \end{figure}
\vspace{-2cm}
\noindent\begin{tabular}{ll}
Anabel Yong & 09/09/24 \\
\makebox[2.5in]{\hrulefill} & \makebox[2.5in]{\hrulefill}\\
\textit{Signature} & \textit{Date}\\
\end{tabular}

\tableofcontents

\thispagestyle{plain}
\listoffigures
\listoftables
%\lstlistoflistings
\listofalgorithms

%\begin{singlespace}
\chapter*{List of Abbreviations}
\begin{sortedlist} %sort alphabetically
  \sortitem{MCS:     Monte Carlo Sampling}
  \sortitem{SOBO:    Single-Objective Bayesian Optimization}
  \sortitem{MPO:     Multi-Property Objectives}
  \sortitem{MOBO:    Multi-Objective Bayesian Optimization}
  \sortitem{ECFP:    Extended-Connectivity Fingerprints}
  \sortitem{GP:      Gaussian Processes}
  \sortitem{GPR:     Gaussian Process Regression}
  \sortitem{NLML:    Negative Log Marginal Likelihood}
  \sortitem{RBF:     Radial Basis Function}
  \sortitem{RKHS:    Reproducing Kernel Hilbert Space}
  \sortitem{SVMs:    Support Vector Machines}
  \sortitem{EI:      Expected Improvement}
  \sortitem{EHVI:    Expected Hypervolume Improvement}
  \sortitem{TDC:     Therapeutics Data Common}
  \sortitem{HV:      Hypervolume}
  \sortitem{MOGP:    Multi-output Gaussian Processes}
  \sortitem{MO-TKGP:  Multi-output Tanimoto Kernel Gaussian Processes}
  \sortitem{GP-MOBO: Gaussian Process Multi-Objective Bayesian Optimization}
  \sortitem{UCB-PT:  Upper Confidence Bound - PyTorch}
  \sortitem{EI-PT:   Expected Improvement - PyTorch}
  \sortitem{KERN-GP-EHVI: Custom Kernel-Gaussian Process-Expected Hypervolume Improvement}
  \sortitem{KERN-GP-EI: Custom Kernel-Gaussian Process-Expected Improvement}
  \sortitem{RS:     Random Sampling}
\end{sortedlist}
% \end{singlespace}
%%%%%%%%%%%%%%%%%%%%%%%%%%%%%%%%%%%%%%%%%%%%%%%%%%%%%%%%%%%%%%%%%%%%%%%%%%%%%%%%
\chapter{Introduction} \label{Chap2}
\pagenumbering{arabic}
\section{Machine Learning Models for Molecular Optimization}
In recent years, machine learning models have gained significant traction in molecular optimization, in the realm of drug discovery. These models offer potential solutions for navigating vast chemical spaces to identify molecules with desirable properties such as high efficacy and low toxicity. Molecular optimization is typically framed as an optimization task over a molecular space $\mathcal{M}$, with the goal of balancing multiple competing objectives - such as pharmacokinetic and pharmacodynamic properties - in real-world applications. The complexity of these problems grows substantially when multiple objectives must be optimized simultaneously\cite{gao_2024_sample}\cite{tripp_2024_diagnosing}. 

Most existing optimization frameworks, particularly those that employ Bayesian Optimization (BO) have been designed for single-objective tasks. This traditional approach involves scalarizing multiple objectives into a single scalar function, which simplifies the optimization process but forces an implicit trade-off between objectives, even when the trade-offs may not be well understood or defined in advance. Real-world problems, however, require multi-objective optimization (MOO), which allows for simultaneous optimization across several criteria without pre-defining their trade-offs. In this thesis, we investigate the effectiveness of a simple multi-objective Bayesian Optimization (MOBO) approach using independent Gaussian Processes (GPs) and an Expected Hypervolume Improvement (EHVI) acquisition function. Despite the simplicity of this setup, it has not been thoroughly explored in the context of molecular optimization, and we hypothesize that it could yield competitive performance in comparison to more complex models. 

\subsection{Motivation for Multi-Objective Bayesian Optimization}
Single-objective optimization frameworks dominate the literature, where multiple objectives are scalarized into a single composite score. This scalarization, however, introduces limitations, as it implicitly specifies a fixed trade-off between objectives. For example, optimizing for $a=1, b=2$ assumes this trade-off is preferable to $a=2,b=1$, yet in many practical cases, the preferred trade-off is not known a priori. In contrast, multi-objective Bayesian optimization (MOBO) seeks to approximate the Pareto frontier, where multiple trade-offs between objectives are discovered rather than being predefined. 

Scalarized Bayesian Optimization(BO) models for molecular optimization have performed adequately in some settings but they are ill-suited to complex, real-world tasks, particularly in drug discovery, where trade-offs between properties like efficacy and toxicity are often unknown. MOBO offers a more flexible and comprehensive solution by modeling each independently, allowing researchers to explore the trade-off space more freely. 

Our motivation stems from this gap in the literature: many works focus on single-objective BO, while real-world problems are inherently multi-objective. We aim to demonstrate that a simpler MOBO approach, usign independent Gaussian Processes (GP) for each molecular objective, can be highly effective for molecular optimization. This research explores whether the MOBO acquisition function, called EHVI, which is well-established for finding Pareto-optimal solutions, combined with independent GPs, can outperform or at least match the performance of finding optimal chemical compounds shown in this benchmark paper by Gao et al(2022) \cite{gao_2024_sample}.

\section{Related Work}
\subsection{Molecular Optimization}
Molecular optimization in drug discovery has seen significant progress through the integration of machine learning. Generative models, such as Variational Autoencoders (VAEs) \cite{kusner_2017_grammar}\cite{jin_2019_junction}, and Generative Adversarial Networks (GANs)\cite{decao_2022_molgan}, have demonstrated promise in proposing novel molecules. These VAEs \cite{quadrant_2019_all}\cite{karlova_2021_molecular}\cite{tazhigulov_2022_molecular}\cite{bradshaw_2020_barking}\cite{jin_2019_junction}\cite{maus_2022_local} typically map molecular structures to a latent space and apply single-objective BO techniques, to find optimal solutions. Despite their potential, these models often require large datasets and suffer from high computational costs. Fitting a GP in the high-dimensional latent space of a VAE is challenging, as the latent space is often complex and not well-suited for direct GP application. The lack of smoothness and continuity in the VAE's latent space can further hinder the GP's ability to make accurate predictions, complicating the balance between exploration and exploitation in Bayesian Optimization (BO) tasks \cite{grosnit_2021_highdimensional}. Moreover, modeling such latent spaces often requires a large number of samples, making the optimization process inefficient, especially for high-dimensional tasks \cite{btepage_2021_gaussian}.

Reinforcement learning frameworks like REINVENT\cite{olivecrona_2017_molecular} or GFlowNets\cite{bengio_2021_gflownet}, as well as genetic algorithms, have also been proposed to explore the molecular space. However, these methods face challenges in efficiently balancing exploration and exploitation, particularly when generating invalid SMILES strings or exploring sub-optimal regions of the chemical space. For example, genetic algorithms\cite{brown_2019_guacamol}\cite{gao_2022_amortized}\cite{jensen_2019_a} rely on random mutations of known molecules, which can lead to inefficient search processes \cite{tripp_2024_diagnosing}.

Gaussian Process Bayesian Optimization (GP-BO)\cite{josmiguelhernndezlobato_2017_parallel} has emerged as a popular framework for molecular optimization. GP-BO excels at modeling uncertainty, making it suitable for applications with sparse or expensive-to-acquire data \cite{josmiguelhernndezlobato_2017_parallel}\cite{tripp_2023_a}. However, many of these models, including GP-BO, focus on single-objective optimization, using scalarization techniques to combine objectives into a single metric. While scalarization simplifies the optimization process, it pre-defines trade-offs between objectives, which may not be well-understood in real-world drug discovery scenarios. This limitation is particularly problematic in multi-objective optimization tasks where the goal is to find the Pareto frontier.

While many previous work have focused on large complex models, we demonstrate that a straightforward method, free from scalarization, can effectively handle multi-objective tasks, achieving competitive performance with current state-of-the-art GP-BO.

\subsection{Multi-Objective Bayesian Optimization}
Multi-objective Bayesian Optimization (MOBO) extends the principles of Bayesian optimization problems involving multiple conflicting objectives, aiming to efficiently approximate the Pareto front- the set of non-dominated solutions representing the best trade-offs among objectives \cite{knowles_2006_parego}. Unlike single-objective optimization, MOBO does not require scalarization of objectives, avoiding the need to predefined trade-offs, which is particularly advantageous when these trade-offs are unknown or difficult to specify in advance. 

Early approaches to MOBO often relied on scalarization techniques, such as weighted sums or utility functions, to combine multiple objectives into a single objective function\cite{marler_2009_the}. However, these methods inherently require the specification of weights or parameters, which can bias the search towards certain regions of the Pareto front and may not capture true diversity of optimal solutions. 

To overcome these limitations, researchers have developed acquisition functions specifically designed for multi-objective settings. One prominent example is the Expected Hypervolume Improvement (EHVI) acquisition function \cite{emmerich_2008_the}\cite{yang_2019_efficient}, which quantifies the expected increase in the hypervolume bounded by the current Pareto front and a reference point. EHVI guides the optimization process toward solutions that contribute most to improving the Pareto front, effectively balancing exploration and exploitation

Another notable acquisition function is the Pareto Expected Improvement (PEI), which extends the concept of Expected Improvement from single-objective optimization to multi-objective contexts\cite{keane_2006_statistical}. PEI evaluates the expected improvement over the current Pareto front, promoting diversity in the discovered solutions. Methods like Predictive Entropy Search for Multi-objective Optimization (PESMO)\cite{hernandezlobato_2016_predictive} and Multi-objective Upper Confidence Bound (MOUCB)\cite{daulton_2020_differentiable} have also been proposed to efficiently navigate the trade-offs between objectives.

In terms of modeling approaches, both independent Gaussian Processes (GPs) for each objective and multi-output GPs that capture correlations between objectives have been employed \cite{bonilla_2007_multitask}\cite{alvarez_2024_kernels}. While multi-output GPs can model inter-objective dependencies, they often come with increased computational complexity, especially in high-dimensional settings. Independent GPs offer a simpler alternative, with each GP modeling an individual objective, making them scalable and easier to implement.

MOBO has been successfully applied in various domains, including engineering design optimization\cite{falkner_2018_bohb}, hyperparameter tuning in machine learning models \cite{frazier_2008_a}, and materials science for discovering new compounds with desired properties \cite{gmezbombarelli_2018_automatic}. Despite these advances, the application of MOBO in molecular optimization remains relatively underexplored. Most molecular optimization studies have focused on single-objective problems or have used scalarization methods when dealing with multiple objectives \cite{olivecrona_2017_molecular}\cite{gmezbombarelli_2018_automatic}. This gap suggests a missed opportunity to fully exploit the capabilities of MOBO in discovering diverse and Pareto-optimal molecules.

Recent efforts have started to bridge this gap. For example, Hernandez-Lobato et al.(2016)\cite{miguelhernndezlobato_2016_a} proposed a general framework for constrained Bayesian optimization using information-based search strategies, which can be adapted to multi-objective scenarios. However, comprehensive studies that systematically apply MOBO techniques - particularly those utilizing simple and scalable models like independent GPs with EHVI - to molecular optimization tasks are still lacking.

\section{Novelty Aspects and Contributions of this Paper}
Our work aims to address this deficiency by investigating the effectiveness of a straightforward MOBO approach in molecular optimization. By employing independent GPs for each objective and leveraging the EHVI acquisition function, we seek to efficiently approximate the Pareto front without the need for complex modeling techniques or large training datasets. This approach not only simplifies the implementation but also has the potential to uncover a more diverse set of optimal molecules, better reflecting the multifaceted objectives inherent in drug discovery and other chemical optimization problems. 

Although Mehta et al(2022)\cite{mehta_2022_momemes} has acknowledged this issue and proposed a multi-objective Bayesian optimization setup by multiplying single-objective acquisition function, the focus of their approach differs from ours. They coined their method the "multi-objective Bayesian optimization acquisition function", yet our approach integrates a more robust and interpretable framework to capture trade-offs. Additionally, the only other notable implementation of surrogate model-based MOBO for molecular optimization is from MIT Coley's Group with their MolPAL \cite{graff_2021_accelerating}, which extends from a single-objective setup using message passing neural networks (MPNNs) as surrogate models. While their MOBO approach showed improvement over scalarization, it did not achieve competitive performance against single-objective MolPAL, ranking 13th in Gao's benchmark (see Table 5 of \cite{gao_2024_sample}). This highlights the need for more effective solutions like ours, which is benchmarked against Tripp et al.'s GP-BO (2021)\cite{tripp_2023_a}, one of the best-performing model in molecular optimization. 

In addition, we introduce \texttt{KERN-GP}, a kernel-only Gaussian Process package that enables the use of exact Tanimoto coefficients, retaining full molecular fingerprint dimensionality without the need for projection to lower dimensions. This allows us to better capture the intricacies of molecular structures, contributing to more accurate predictions and robust optimization results. We also incorporate MinMax Kernels to handle count fingerprints, providing enhanced flexibility and computational efficiency \cite{tripp_2024_diagnosing}\cite{ralaivola_2005_graph}.

Our GP-MOBO framework is benchmarked against Tripp et al's GP-BO (2021)\cite{tripp_2023_a}, the 4th best-performing model in molecular optimization\cite{gao_2024_sample}. By expanding GP-BO into a multi-objective framework, we demonstrate that even with a relatively simple setup, our model outperforms state-of-the-art methods in terms of Pareto front diversity and solution quality. This represents a significant improvement in the field of multi-objective molecular optimization, offering a scalable and interpretable alternative to more complex models.

The subsequent chapter will provide some cheminformatics preliminaries, theoretical background on Multi-Output Gaussian Processes, Tanimoto and MinMax Kernels, and multi-objective Bayesian optimization. This will be followed by a detailed methodology and experimental design to validate the performance of our GP-MOBO framework. Our results showcase the superiority of GP-MOBO over GP-BO, and the discussion will outline key takeaways and potential future directions to further enhance our approach.

%%%%%%%%%%%%%%%%%%%%%%%%%%%%%%%%%%%%%%%%%%%%%%%%%%%%%%%%%%%%%%%%%%%%%%%%%%%%%%%%
\chapter{Background} \label{Chap3}
\section{GUACAMOL's Molecular Property Objectives (MPOs)}
\label{section:guacamol-mpo}
Multi-Property Objectives (MPOs) are a set of goals used in drug discovery to find compounds that satisfy several properties simultaneously. In this particular well-known benchmark dataset, GUACAMOL \cite{brown_2019_guacamol}, MPOs are designed to replicate the complex criteria that real-world drug candidates must meet. These objectives often encompass a variety of properties such as chemical similarity, pharmacokinetic properties and structural constraints. In Table \ref{tab:MPO-table}, some of these MPOs are as detailed below. 
\begin{table}[H]
\centering
\caption{Examples of Goal-Directed Benchmarks in GUACAMOL Dataset}
\label{tab:MPO-table}
\begin{tabular}{|l|l|l|l|l|}
\hline
\textbf{Benchmark Name} & \textbf{Scoring} & \textbf{Mean} & \textbf{Scoring Function(s)} & \textbf{Modifier} \\ \hline
Osimertinib MPO & top-1 & geom & sim(osimertinib, FCFP4) & Thresholded(0.8) \\ \cline{2-2} \cline{4-5}
 & top-10 &  & sim(osimertinib, ECFP6) & MinGaussian(0.85, 2) \\ \cline{2-2} \cline{4-5}
 & top-100 &  & TPSA & MaxGaussian(100, 2) \\ \cline{4-5}
 &  &  & logP & MinGaussian(4, 2) \\ \hline
Fexofenadine MPO & top-1 & geom & sim(fexofenadine, AP) & Thresholded(0.8) \\ \cline{2-2} \cline{4-5}
 & top-10 &  & logP & MinGaussian(4, 2) \\ \cline{2-2} \cline{4-5}
 & top-100 &  & TPSA & MaxGaussian(90, 2) \\ \hline
Ranolazine MPO & top-1 & geom & sim(ranolazine, AP) & Thresholded(0.7) \\ \cline{2-2} \cline{4-5}
 & top-10 &  & logP & MinGaussian(4, 2) \\ \cline{2-2} \cline{4-5}
 & top-100 &  & TPSA & MaxGaussian(95, 20) \\ \cline{4-5}
 &  &  & number of fluorine atoms & Gaussian(1, 1) \\ \hline
\end{tabular}
\end{table}
The Osimertinib MPO is focused on finding molecules that are similar to Osimertinib (an anti-cancer drug) while meeting additional constraints such as logP and Topological Polar Surface Area (TPSA). The similarity to the target molecule is combined with other property scores using the geometric mean. These score modifiers shown in Table \ref{tab:MPO-table}, details how each modifier influences the scoring function in multi-objective optimization problems. How these score modifiers evaluate molecules are mostly detailed in Brown's GUACAMOL paper \cite{brown_2019_guacamol}. 

All generative machine learning models, to date, utilize this scalarization by geometric mean optimization setup to make the single-objective optimization problem tractable \cite{gao_2024_sample}. However, this can mask trade-offs between objectives and makes it harder to achieve a truly balanced solution. Here, in this research, will be the first investigation into separating Guacamol MPO objectives and working with these scoring functions for our algorithm. How did we do it without scalarizing the geometric mean? This is detailed in our Methodology and Experimental Design (Section \ref{section:gp-mobo}) below. 

\section{SMILES}
\begin{figure}[H]
    \centering
    \includegraphics[width=0.4\linewidth]{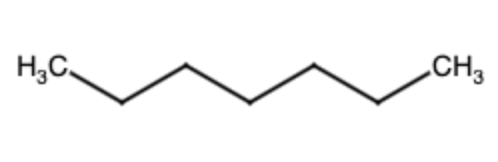}
    \caption{Molecular Structure for SMILES string \textcolor{red}{"CCCCCCC"}.}
    \label{fig:SMILES}
\end{figure}

The Simplified Molecular-Input Line-Entry System (SMILES) offers a textual string format for encoding molecular structures, as introduced by Anderson et al.(1987)\cite{anderson_1987_smiles} and further developed by Weininger(1988)\cite{weininger_1988_smiles}. Examples of SMILES are illustrated in Figure \ref{fig:SMILES}.
SMILES represents molecules as a sequence of atoms and bonds using short ASCII strings, where atoms are denoted by their chemical symbols. The advantage of using SMILES is its interpretability, allowing for direct integration with machine learning models. However, as acknowledged by Taleongpong and Paige (2024)\cite{taleongpong_2024_improving}, SMILES lack structural invariance, where the same molecule can be represented by different strings due to variations in atom ordering and molecular conformation. 

\section{Therapeutics Data Common Oracles}
Oracles, provided by Therapeutics Data Commons (TDC)\cite{huang_2021_therapeutics}, are functions or models that evaluate specific molecular properties represented by SMILES strings. These predictive models are commonly used in drug discovery tasks to generate or optimize molecules for desirable properties such as high docking scores, low toxicity, or bioavailability.
\begin{figure}[H]
    \centering
    \includegraphics[width=1.1\linewidth]{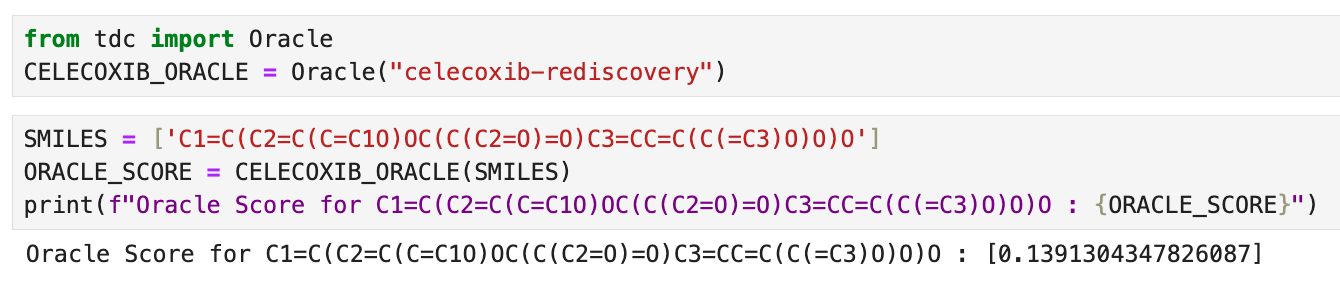}
    \caption[\textbf{Use of an Oracle from Therapeutics Data Common(TDC) to evaluate SMILES corresponding to a molecular structure for its similarity to the celecoxib molecule:}]{\textbf{Use of an Oracle from Therapeutics Data Common(TDC) to evaluate SMILES corresponding to a molecular structure for its similarity to the celecoxib molecule:} The Oracle function is tailored for "celecoxib-rediscovery" task, which returns a numerical score that quantifies simlarity of molecule to desired properties of celecoxib.}
    \label{fig:oracle-functions}
\end{figure}
The TDC package includes several oracles that evaluate physiochemical properties (e.g., logP, logD, QED, molecular weight) and toxicity (e.g., hERG inhibition). Many of these oracle functions, such as those used in Guacamol MPO tasks(Table \ref{tab:MPO-table}), are readily available for use in molecular optimization efforts.

\section{Molecular Fingerprints}
\label{section:fingerprints}
Additionally, SMILES are also an integral part of generating molecular fingerprints. Molecular fingerprints represent chemical compounds as fixed-length vectors suitable for machine learning models\cite{rogers_2010_extendedconnectivity}. These fingerprints capture the presence or absence of specific molecular substructures or properties, facilitating applications such as virtual screening, and similarity searching. Various types of molecular fingerprints exist, including structural key-based\cite{durant_2002_reoptimization}, path-based \cite{hinselmann_2011_jcompoundmapper}, circular \cite{rogers_2010_extendedconnectivity} and pharmacophore \cite{mah_2006_the} fingerprints. However, in this research, we will solely focus on Morgan fingerprinting \cite{rogers_2010_extendedconnectivity}, a circular fingerprinting technique. Among these, Morgan fingerprints have gained prominence due to their robustness and flexibility in capturing molecular features relevant to various cheminformatics tasks. 

\subsection{Morgan/ECFP Fingerprints}
Extended Connectivity Fingerprints (ECFP)\cite{rogers_2010_extendedconnectivity}, represent a sophisticated method for encoding molecular structures. This approach assigns unique numeric identifiers to each atom in a molecule, which are iteratively updated based on the identifiers fingerprint's diameter, a parameter that defines the extent of atomic neighborhoods considered during the fingerprinting process. For instance, ECFP6 fingerprints, which are widely used, involve three iterations of updating atom identifiers, corresponding to a diameter of six bonds in the molecular graph \cite{rogers_2010_extendedconnectivity}\cite{hassan_2006_cheminformatics}.

\subsection{Generation of Molecular Fingerprints}
\begin{figure}[H]
    \centering
    \includegraphics[width=1.1\linewidth]{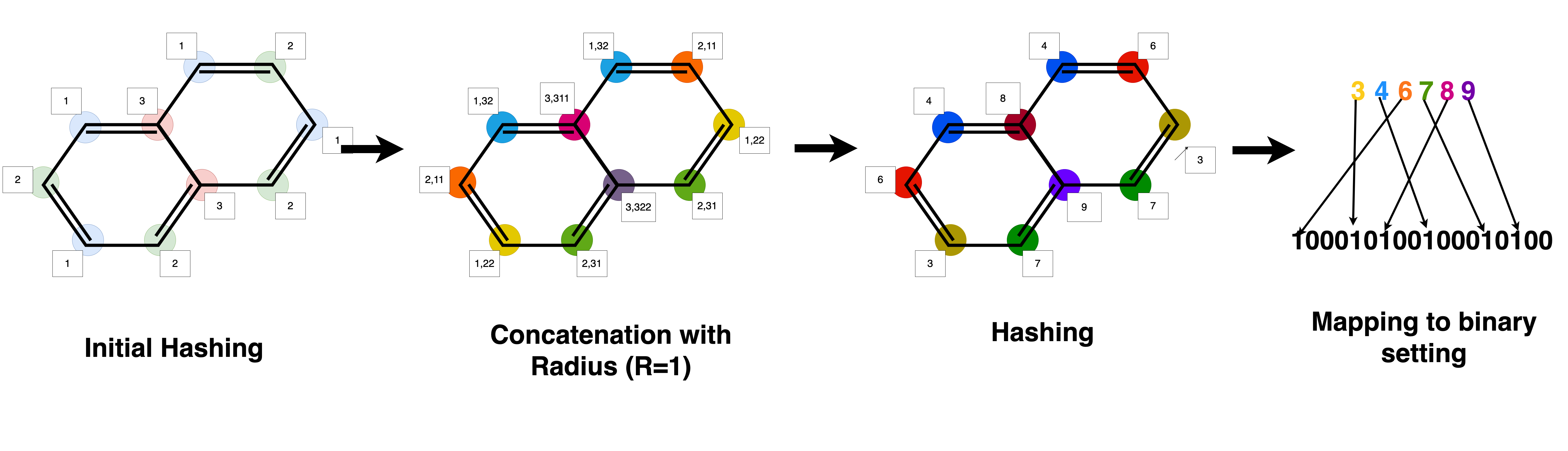}
    \caption[\textbf{Morgan fingerprinting process for a naphthalene molecule:}]{\textbf{Morgan fingerprinting process for a naphthalene molecule:}
    The process begins with initial hashing where each atom is assigned an integer based on its features (Step 1). These integers are then concatenated with the integers of their neighboring atoms (Step 2), followed by hashing the concatenated values to produce new integers (Step 3). Finally, the hashed integers are mapped to a binary string to generate the molecular fingerprint (Step 4). (\textit{\textbf{Adapted from Hernandez-Lobato, Jose M. "Machine Learning for Molecules.", 2018}})}
    \label{fig:1}
\end{figure}

An overview of the Morgan fingerprinting process is illustrated in Figure \ref{fig:1}. Each atom in the molecule is initially assigned a numeric identifier based on its atomic properties, such as the number of neighboring atoms, atomic number, and the number of attached hydrogens. These initial identifiers are collected into a set representing the atom-centered environment. Subsequently, in the next step, each atom's identifier is updated by incorporating information from the identifiers of its immediate neighbors. This process continues iteratively, with each iteration expanding the scope of the atomic neighborhood considered. The result is a series of increasingly complex identifiers that encapsulate larger sub-structural features of the molecule. Once all iterations are complete, the algorithm identifies and removes duplicate identifiers, ensuring that each substructure is represented only once in the final fingerprint vector. This step prevents redundancy and ensures the compactness of the fingerprint. The final step involves hashing the unique identifiers to generate a fixed-length binary string (or bit vector). Each bit in this vector indicates the presence or absence of a specific substructure within the molecule. This binary representation is what constitutes the Morgan fingerprint, which can be used in various cheminformatics applications \cite{landrum_2013_rdkit}\cite{adityaraymondthawani_2020_the}.

A crucial advantage of Morgan fingerprints is their ability to encode stereo-chemical information, which is particularly important where molecular chirality plays a critical role. The resulting identifiers from the iterative process are hashed into a fixed-length binary or integer vector creating a compact and information-rich representation of the molecule. Unlike structural key-based fingerprints, which rely on predefined patterns, Morgan fingerprints can represent an essentially infinite number of substructures, including those not explicitly coded in any database. 
\begin{figure}[H]
    \centering
    \includegraphics[width=0.8\linewidth]{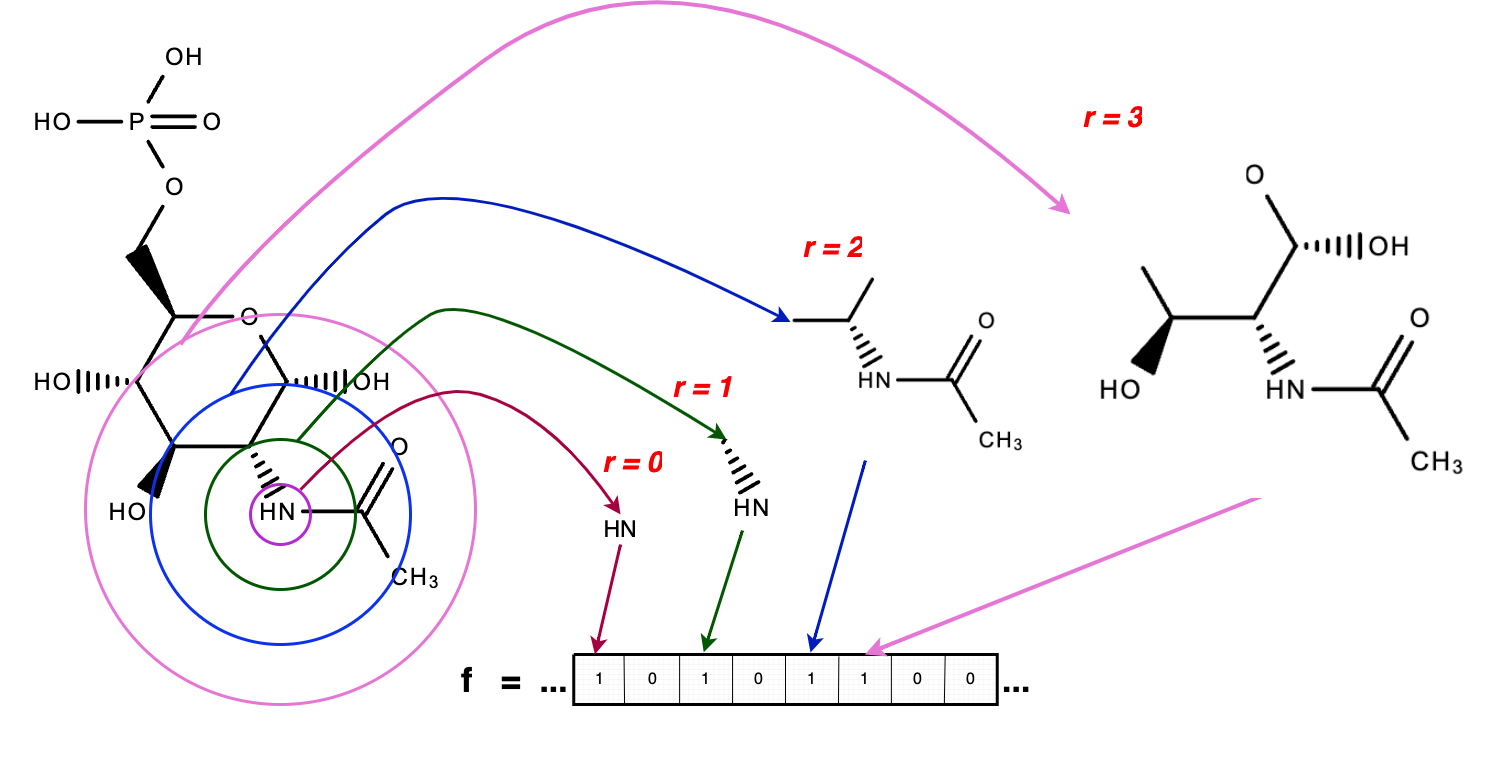}
    \caption[\textbf{Example of Morgan fingerprinting process applied to a molecule with a manually set radius parameter $R=3$.}] {\textbf{Example of Morgan fingerprinting process applied to a molecule with a manually set radius parameter $R=3$.}The central atom (NH) is considered at radius $r=0$. Atoms directly bonded to NH are at radius $r=1$, atoms bonded to those atoms are at radius $r=2$, and so on. The features of these atoms are concatenated and hashed at each step to generate the molecular fingerprints.(\textit{\textbf{Adapted from Hernandez-Lobato, Jose M. "Machine Learning for Molecules." MLSS, 2018}})}
    \label{fig:2}
\end{figure}

This allows them to capture novel or unexpected molecular features. This is useful for tasks such as similarity searching and clustering. Furthermore, their ability to generate unique, non-redundant representations ensures efficient and accurate comparisons between molecules, a critical requirement in modern cheminformatics\cite{landrum_2013_rdkit}. Acknowledged by Bradshaw et al(2020)\cite{bradshaw_2020_barking}, as these fingerprints are fixed, models cannot learn which characteristics of a molecule are important for our tasks.

\subsection{Bit Collisions Phenomenon}
Despite these advantages, a significant limitation arises in the context of dimensionality reduction. Most, if not all machine learning models (MIT's SynNET\cite{gao_2022_amortized}, SMILES GA\cite{yoshikawa_2018_populationbased}, Stanford's MolDQN\cite{zhou_2019_optimization}, GP BO \cite{tripp_2023_a} are some notable models) in cheminformatics, typically reduce the dimensionality of molecular fingerprints to a fixed size between 1024 to 4096 bits. This dimensionality reduction, while computationally efficient, introduces problem of bit collisions, where different molecular substructures might be hashed into the same bit position. This leads to loss of unique information, and a decrease in discriminative power of the fingerprint. Riniker and Landrum (2013)\cite{riniker_2013_similarity} highlight the impact of bit collisions in their study on dopamine receptor ligands, where reducing the bit size from 2048 to 1024 bits resulted in overlap of different chemical environments. This problem is addressed in our methodology in Section \ref{subsection:custompackage}.
\begin{figure}[H]
    \centering
    \includegraphics[width=0.8\linewidth]{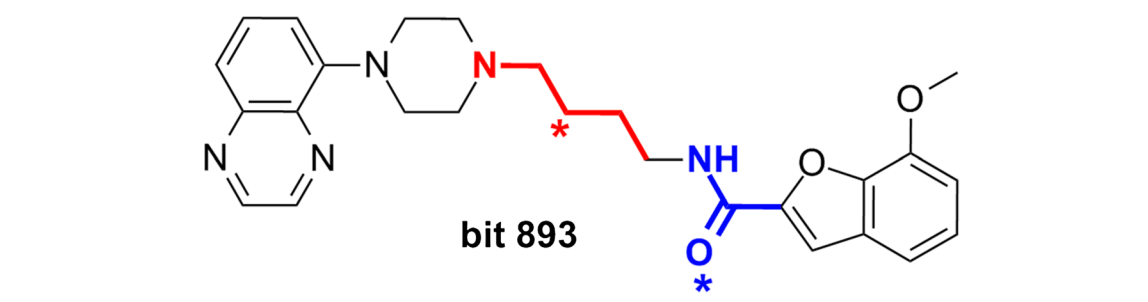}
    \caption[\textbf{Example of Bit Collision in Morgan2/CountMorgan2 Fingerprints:}]{\textbf{Example of Bit Collision in Morgan2/CountMorgan2 Fingerprints:} The molecular environments highlighted in red and blue are both hashed to the same bit position (bit 893) in the fingerprint vector. The central atom of each environment is marked with a star, indicating the points where the collision occurs, leading to a loss of unique molecular information in the fingerprint(Riniker and Landrum (2013)).}
    \label{fig:bit-collisions}
\end{figure}

\subsection{Binary vs Count Fingerprints}
\begin{figure}[H]
    \centering
    \includegraphics[width=\linewidth]{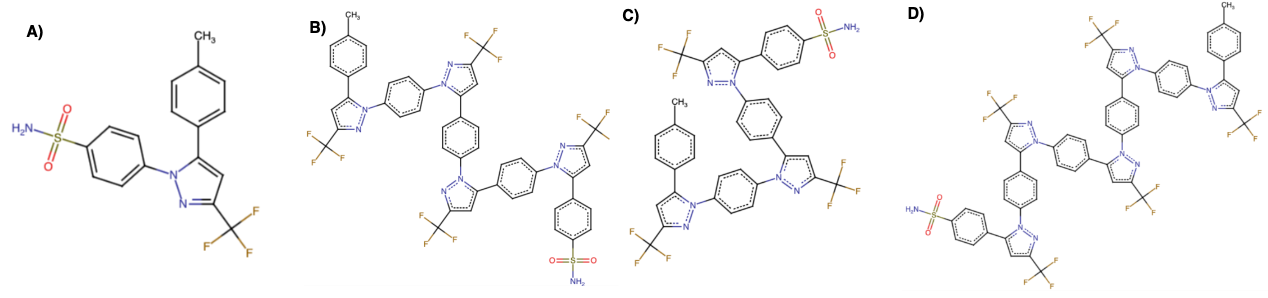}
    \caption[\textbf{Celecoxib (A) and its larger analogues (B,C,D) are represented by the same Binary Morgan Fingerprints:}]{\textbf{Celecoxib (A) and its larger analogues (B,C,D) are represented by the same Binary Morgan Fingerprints:} This issue arises due to the hash-based nature of Morgan fingerprints, which can lead to identical fingerprint representations when different molecules share certain substructures.}
    \label{fig:celecoxib}
\end{figure}
While binary fingerprints indicate the presence or absence of a substructure, count fingerprints provide additional granularity by representing the frequency of those substructures within a molecule\cite{tripp_2024_diagnosing}. Figure \ref{fig:celecoxib} highlights a limitation of binary fingerprints, where molecules with different structural sizes, are reduced to identical binary representations. This leads to a potential loss of molecular detail. In contrast, count fingerprints not only capture whether a molecular feature is present but how frequently it occurs. As noted in literature, count-based methods generally improve model performance in tasks requiring more detailed comparisons. Therefore, count fingerprints are often preferred over binary fingerprints where sub-structural frequency plays a role in molecular optimization. 

Contrary to popular belief, the computational efficiency of binary and count fingerprints is essentially the same. Both utilize efficient data structures such as hash tables and hash maps for message passing operations, with similar computational complexity. Therefore, the preference for count-based fingerprints stems from their ability to offer richer molecular representations, particularly in tasks that require more detailed sub-structural comparisons. As a result, count fingerprints are often favored in molecular optimization tasks where the frequency of molecular features plays a crucial role.

Having established the limitations of binary fingerprints and the advantages of count fingerprints, we now turn to the underlying mathematical models that guide our approach. Gaussian Processes (GPs) form the foundation of our model, with the Tanimoto kernel specifically tailored to handle binary fingerprints, capturing molecular similarity through presence-absence patterns. To extend this capability, we incorporate the MinMax kernel, a derivative of the Tanimoto kernel, to better represent count fingerprints, ensuring that the frequency of molecular features is accurately modeled. This theoretical framework, along with Multi-Objective Bayesian Optimization (MOBO), enables efficient exploration of the Pareto front, optimizing for multiple conflicting objectives simultaneously.

\newpage
\section{Gaussian Processes}
Gaussian Processes (GPs)\cite{rasmussen_2006_gaussian} are a powerful and versatile class of models used in machine learning for regression and classification tasks. Unlike traditional machine learning algorithms that focus on finding a single best-fit model, GPs offer a probabilistic framework, representing a distribution over possible functions that fit the data. This probabilistic nature allows GPs to naturally quantify uncertainty in predictions, which is valuable in tasks where understanding the confidence of predictions is crucial. 

At the core of GPs, they are defined as a collection of random variables, any finite subset of which has a joint Gaussian distribution. This means that instead of predicting a single output for the given input, GPs predict a distribution over possible outputs, characterized by mean function $m(x)$ and covariance function $k(x,x')$\cite{rasmussen_2006_gaussian}. The mean function represents the expected output, while the covariance function, which can be a kernel, determined the similarity between different inputs and governs the smoothness and generalization ability of predictions. Formally, a GP is defined as: 
\begin{equation*}
    f(x) \sim \mathcal{G}\mathcal{P}(m(x), k(x,x'))
\end{equation*}
where $m(x) = \mathbb{E} [f(x)]$, and $k(x,x') = \mathbb{E}[(f(x)-m(x))(f(x')-m(x'))]$ is the covariance function. The kernel function $k(x,x')$ plays a crucial role in GPs as it encodes the assumptions about the function we wish to learn, such as smoothness or periodicity. The choice of kernel, in which we will justify our case for using these Tanimoto kernels in Section , directly influences the GP's predictions and its ability to model the underlying data effectively. 

The flexibility of GPs comes from their non-parametric nature. Unlike parametric models, which assume a specific functional form for the data, GPs can model a wide range of functions by adjusting the kernel. This makes GPs a versatile and highly adaptable tool for various types of data, but also introduces the challenge of selecting an appropriate kernel, and optimizing its hyperparameters, which can be computationally demanding\cite{griffiths_2023_gauche}. 

\subsection{Predictive Inference with GPs}
Gaussian Process Regression (GPR)\cite{rasmussen_2006_gaussian} is the application of Gaussian Processes to regression problems. In GPR, the goal is to infer a distribution over possible functions that fit the observed data. Given a set of training data $\mathcal{D} = \{X,Y\}$, where X represents the input features and Y the corresponding outputs, GPR uses Bayes' theorem to update the prior distribution (the GP) with the observed data, resulting in a posterior distribution over functions. This joint distribution of observed outputs Y and the function values $f_*$ at the test points $X_*$ is given by:
\begin{equation*}
\begin{pmatrix}
Y \\
f_*
\end{pmatrix}
\sim \mathcal{N} \left(
\begin{pmatrix}
m(X) \\
m(X_*)
\end{pmatrix},
\begin{pmatrix}
K(X, X) + \sigma_n^2 I & K(X, X_*) \\
K(X_*, X) & K(X_*, X_*)
\end{pmatrix}
\right)
\end{equation*}
Here $m(X)$ and $m(X_*)$ are the mean functions, $K(X,X)$ is the covariance matrix between training points $K(X,X_*)$ is the covariance between the training points and test points, and $\sigma^2_n$ is the variance of the Gaussian noise added to observations. 

\begin{figure}[H]
    \centering
    \includegraphics[width=\linewidth]{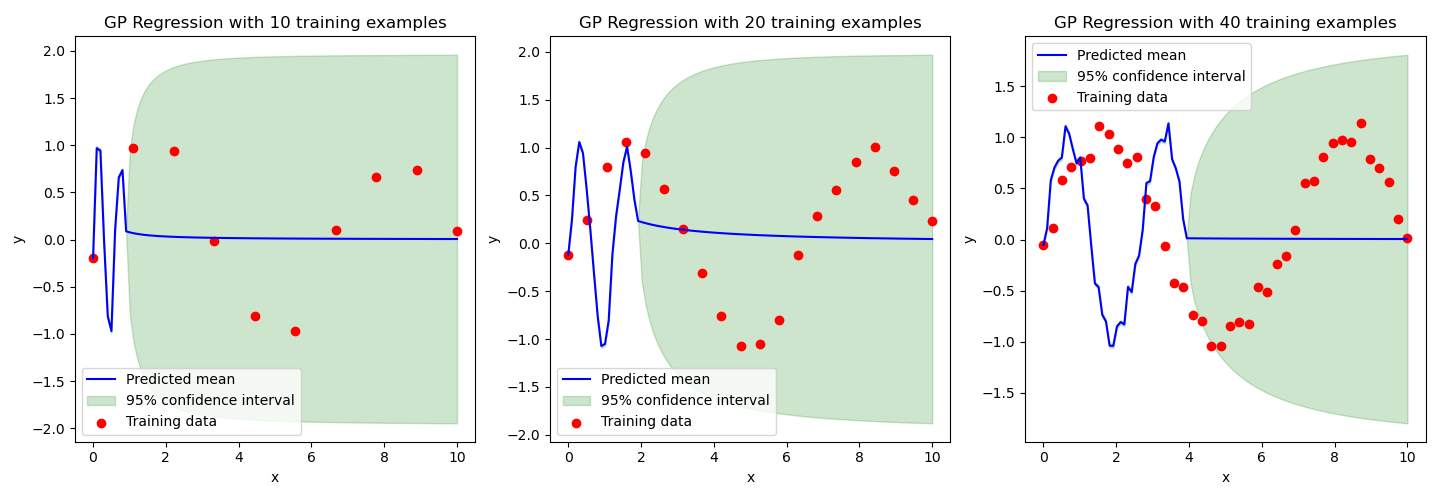}
    \caption[\textbf{Single-output Gaussian Process regression with a full-dimensional Tanimoto Kernel as used in \texttt{KERN\_GP} for exact molecular similarity calculations:}]{\textbf{Single-output Gaussian Process regression with a full-dimensional Tanimoto Kernel as used in \texttt{KERN\_GP} for exact molecular similarity calculations:} The blue line represents the mean of posterior predictive distribution and the green shaded region represents the 95\% confidence interval based on the model's variance estimates. As the number of training examples increases, the mean function begins to exhibit more complex behaviors to match the observed data, but the confidence region does not shrink in the same way as seen in RBF kernels.}
\label{fig:tanimotokernelgp}
\end{figure}

The posterior distribution, which is gives the predictive mean $\mu_*$ and variance $\sum_*$ for the test points, is derived as: 
\begin{align}
\mu_{*} &= m(X_{*}) + K(X_{*}, X)\left[K(X,X) + \sigma_n^2 I\right]^{-1}\left(Y - m(X)\right) \\
\Sigma_{*} &= K(X_{*},X_{*}) - K(X_{*}, X)\left[K(X,X) + \sigma_n^2 I\right]^{-1}K(X,X_{*})
\label{eq:muandsigma}
\end{align}
This result demonstrates the power of GPR: it not only predicts the mean values of the ouputs at new point but also provides a measure of the uncertainty of those predictions. This ability to model uncertainty is one of the key strengths of GPR, especially in scenarios where data is sparse or noisy\cite{rasmussen_2006_gaussian}. The predictive mean $\mu_*$ is a weighted sum of the observed outputs, where the weights are determined by the covariance between the test points and the training points, normalized by the covariance of the training points. The predictive variance $\sum_*$ on the other hand, decreases as more data points are observed reflecting the increasing certainty of the predictions. An example of this is seen in Figure \ref{fig:tanimotokernelgp} which shows the decrease in variance when increasing the number of training samples. 

\section{Multi-Output Gaussian Processes}
While GPRs are a powerful approach in regression tasks, where the goal is to predict a continuous output given an input, many real-world applications, especially in fields of cheminformatics and drug discovery, require the simultaneous optimization of multiple objectives. In such cases, a traditional single-output GP model might fall short. The concept of Multi-Output Gaussian Processes (MOGPs)\cite{alvarez_2024_kernels}\cite{beckers_2021_an}\cite{wilson_2011_gaussian}, extend this single-output GP framework to model multiple outputs jointly. However, in some cases, as assumed in our model, tasks are treated as independent, meaning there is no correlation between the outputs. Each task is modeled by an independent GP, which can be beneficial for computational simplicity and ease of interpretation.  An illustration of the MOGP framework assuming independence is provided in Figure \ref{fig:multioutput-gp-overview}. Now in these later sections, we delve into the theoretical details of what an MOGP entails. 

\subsection{Covariance Matrix and Multi-Output GP}
In the independent task scenario, given a set of $D$ independent objectives, the prior/joint distribution of the functions $f_1, f_2,...,f_D$ can be represented by the following multivariate normal distribution: 
\begin{equation}
\begin{bmatrix}
f_1(x) \\
f_2(x) \\
\vdots \\
f_D(x)
\end{bmatrix}
\sim \mathcal{N}\left(
\begin{bmatrix}
m_1(x) \\
m_2(x) \\
\vdots \\
m_D(x)
\end{bmatrix}, 
\begin{bmatrix}
K_1 & 0 & \dots & 0 \\
0 & K_2 & \dots & 0 \\
\vdots & \vdots & \ddots & \vdots \\
0 & 0 & \dots & K_D
\end{bmatrix} + \Sigma \right)
\label{equation-structure}
\end{equation}
where: 
\begin{enumerate}
    \item $K_i$ is the covariance matrix is the covariance matrix corresponding to the i-th output, determined by an arbitrary kernel. 
    \item $\sum = \text{diag}(\sigma_1^2I, \sigma_2^2I,...,\sigma_D^2I)$ represents the noise in each objective function. 
\end{enumerate}
Generally, without any prior knowledge about the trends of the data, the prior mean function $[m_1(x), m_2(x),...,m_D(x)]^T$ are usually set to 0 (see Rasmussen et al(2005)\cite{rasmussen_2006_gaussian}. Hence, we set the mean functions to - for the remainder of this research, unless stated otherwise. 
\subsection{Covariance/Kernel Structure for Multi-Output GP}
In more detail from Equation \ref{equation-structure} above, Since we assume independence between tasks, the covariance function $K(f_j, f_{j'})$ for a multi-output GP model is block diagonal:
\[
\mathbf{K}_{f,f} = 
\begin{bmatrix}
K_1(x, x') & 0 & \dots & 0 \\
0 & K_2(x, x') & \dots & 0 \\
\vdots & \vdots & \ddots & \vdots \\
0 & 0 & \dots & K_D(x, x')
\end{bmatrix}
\]
This block diagonal structure of the covariance matrix indicates that there is no direct correlation between the different output functions $f_j(x)$ and $f_{j'}(x)$ for $j \neq j'$. This is consistent with the assumption that the outputs are conditionally independent given the latent functions. Each $K_i(x,x')$ represents the covariance for the i-th objective and is computed using a kernel. 

\begin{figure}[H]
    \centering
    \includegraphics[width=\linewidth]{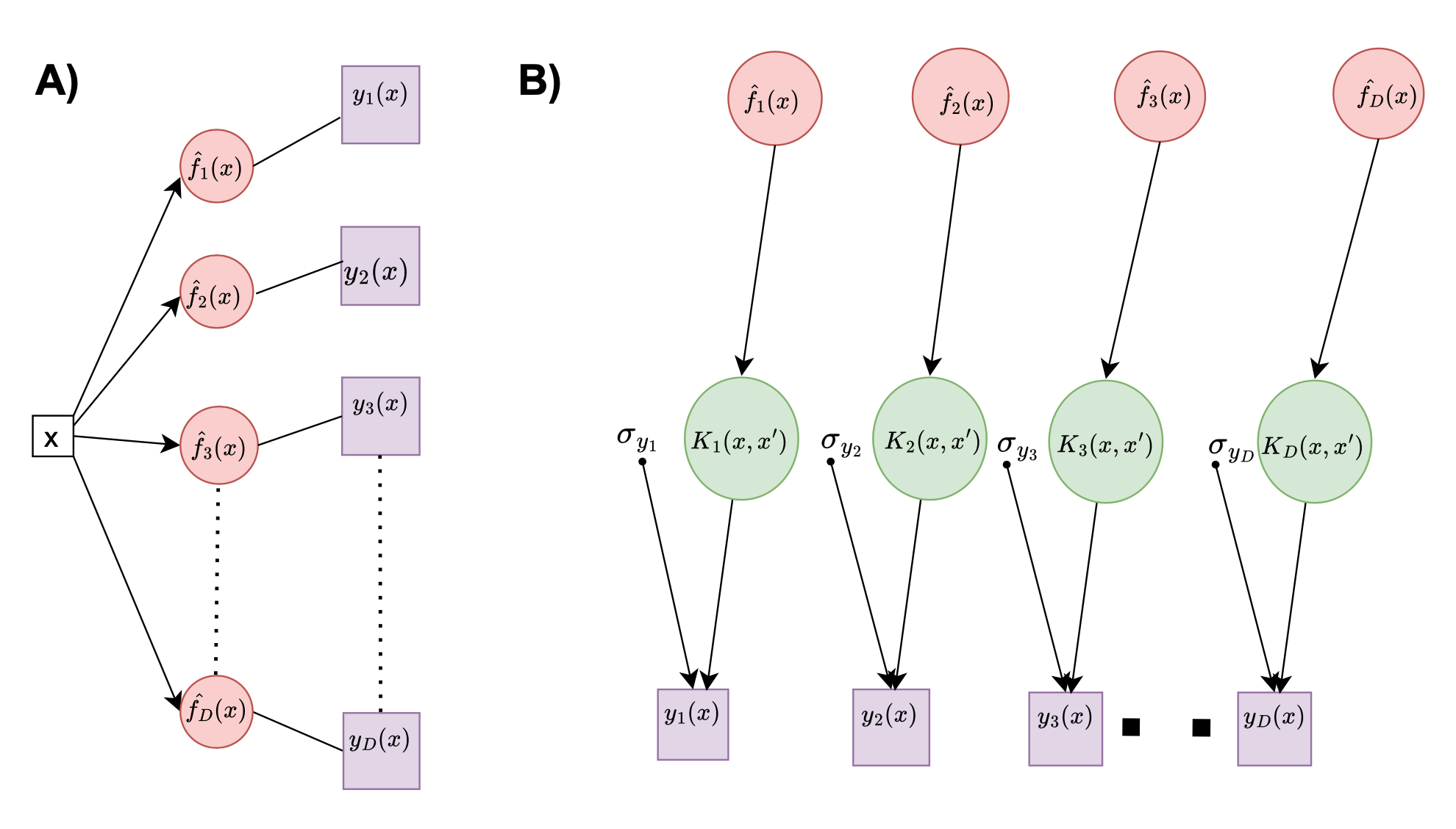}
    \caption[\textbf{Multi-Output Gaussian Processes (MOGP): }]{\textbf{Multi-Output Gaussian Processes (MOGP): }Panel (A) illustrates independent modeling of multiple outputs $y_1(x), y_2(x)...,y_D(x)$ using individual GPs $\hat{f_1}(x), \hat{f_2}(x),...,\hat{f_D}(x)$ for each output. Input x is mapped to each latent function $\hat{f_i}(x)$ independently which predicts outputs. There is no cross-correlation between outputs, allowing each GP to operate independently. Panel (B) shows each latent function $\hat{f_i}(x)$ is associated with its kernel function $K_i(x,x')$, denoting covariance structure for the $i$-th output. Noise term $\sigma_{y_i}$ added to each output to account for observation noise. This model depicted is suited for scenarios where outputs are conditionally independent given latent functions.}
\label{fig:multioutput-gp-overview}
\end{figure}

\subsection{Reproducing Kernel for Vector-valued Functions}
The kernel functions $k_i(x,x')$, denoted in Figure \ref{fig:multioutput-gp-overview}, used in a multi-output GP model are reproducing kernels in the Hilbert space of vector-valued functions \cite{alvarez_2024_kernels}. Specifically, for each (molecular) objective $f_i(x)$, the kernel $k_i(x,x')$ is a matrix-valued function that maps from $\mathcal{X} \times \mathcal{X} \rightarrow \mathbb{R}^{D \times D}$, where D denotes the dimensionality of the output space, where for us, is dependent on the number of objectives we are trying to optimize for\cite{alvarez_2024_kernels}. 

Given a vector-valued function $f(x) = [f_1(x),....,f_D(x)]^T$ belonging to a Hilbert space $\mathcal{H}$, the reproducing property of the kernel $K(x,x')$ ensures that the inner product in $\mathcal{H}$ corresponds to the evaluation of the function $f(x)$. This is formally expressed as: 
\begin{equation*}
    \langle f(x), f(x')\rangle_{\mathcal{H}} = f(x)^TK(x,x')f(x')
\end{equation*}
This kernel function $K(x,x')$ can be expressed as a matrix acting on a vector $c_j \in \mathbb{R}^D$, allowing us to represent the function $f(x)$ as a sum of kernel evaluations over the training data points: 
\begin{equation*}
    f(x) = \sum_{i=1}^N K(x_i, x)c_i
\end{equation*}
Here, the matrix $K(x,x')$ is positive semi-definite and encodes the covariance structure between the different outputs and $c_i$ are the coefficients. We detail this further in our kernels implemented in Section \ref{subsec:TanimotoKernels} and \ref{subsec:MinMax}. 

\subsection{Gaussian Processes for Vector Valued Functions}
For vector-valued functions in a multi-output GP model, the GP is defined as: 
\begin{equation*}
    f(X) \sim \mathcal{G}\mathcal{P}(m(X), K(X,X))
\end{equation*}
where $m(X)$ is the vector that concatenates the mean vectors associated with the outputs and the covariance matrix $K(X,X)$ is block-diagonal, with each block corresponding to the covariance matrix $K_i(X,X)$ for the i-th output. The prior distribution is over the outputs is given by: 
\begin{equation*}
    f(X) \sim \mathcal{N}(m(X), K(X,X))
\end{equation*}
As these outputs $f_i(x)$ are independent, and given a set of input points X and corresponding outputs Y, the posterior distribution can be written as: 
\begin{equation*}
    p(f|Y, X, \sum) = \mathcal{N}(f(X), K(X,X))
\end{equation*}
where $\sum$ accounts for noise in each task. For our independent objectives, the predictions for a new test point $x_*$ have the same $\mu_*(x_*)$ and $\sigma_*(x_*)$ calculations to as seen in Equation \ref{eq:muandsigma} of the Single-output GPs. The predictive mean $\mu_*(x_*)$ and variance $\sigma_*(x_*)$ for each objective are returned as a tuple of vectors:
\[
\mu_*(\mathbf{x}_*) = [\mu_1(\mathbf{x}_*), \mu_2(\mathbf{x}_*), \dots, \mu_D(\mathbf{x}_*)],
\]
\[
\sigma^2_*(\mathbf{x}_*) = [\sigma^2_1(\mathbf{x}_*), \sigma^2_2(\mathbf{x}_*), \dots, \sigma^2_D(\mathbf{x}_*)].
\]

\subsection{Gaussian Process Training}
Now that we have introduced Gaussian Process Regression (GPR), Multi-Output Gaussian Processes (MOGPs) and their predictive mean and variance, it is crucial to discuss the training of Gaussian Processes, particularly focusing on how the hyperparameters of the model are determined. These hyperparameters play a critical role in the performance and flexibility of the GP model. In this section, we explore the concept of GP training by optimizing the \textbf{Negative Log Marginal Likelihood (NLML)}\cite{rasmussen_2006_gaussian}, which balances the model's fit to the data and its complexity, thereby avoiding overfitting. 

In a GP model, the choice of kernel (or covariance function) is important. The kernel defines the relationship between points in the input space and governs the smoothness, periodicity and other properties of the functions drawn from the GP. Each kernel is parameterized by a set of hyperparameters, denoted by $\theta$. For example, in the commonly used Radial Basis Function (RBF) kernel, we have: 
\begin{equation*}
    k(x,x') = \sigma^2_f \exp \left(-\frac{(x-x')^2}{2l^2}\right)
\end{equation*}
Here, $\sigma_f$ controls variance of the function, and l(the length scale) determines how quickly the correlation between points decreases as they move apart in the input space. The noise term $\sigma_n^2$ is treated as a hyperparameter, controlling the variance of the noise assumed in the observations. We will discuss the Tanimoto and MinMax kernel functions in more detail in Section \ref{subsec:TanimotoKernels} and \ref{subsec:MinMax}.

\subsubsection*{Negative Log Marginal Likelihood (NLML)}
\label{section:NLML}
The hyperparameters $\theta$ are optimized by maximizing the marginal likelihood of the observed data y, given the inputs X and the hyperparameters. The marginal likelihood is obtained by integrating out the functions values from the joint probability distribution, resulting in a Gaussian distribution for the data $y$: 
\begin{equation*}
    \log p(y|X, \theta) = \textcolor{blue}{-\frac{1}{2}y^\top (K_{\theta}(X,X) + \sigma^2_y I)^{-1}y} \textcolor{red}{-\frac{1}{2}\log |K_{\theta}(X,X) + \sigma^2_y I|} -\frac{n}{2}\log(2\pi)
\end{equation*}
This expression combines the \textcolor{blue}{$-\frac{1}{2}y^\top (K_{\theta}(X,X) + \sigma^2_y I)^{-1}y$, the data fit term} which encourages the model to fit the observed data closely and \textcolor{red}{$-\frac{1}{2}\log |K_{\theta}(X,X) + \sigma^2_y I|$, the complexity penalty term}, which penalizes overly complex models to prevent over-fitting. The final term is a normalization constant that does not depend on model parameters. Together, these terms represent the NLML which is minimized to find optimal hyperparameters usually. 

The NLML embodies the principle of Occam's Razor\cite{rasmussen_2000_occams}, which favours simpler models that sufficiently explain the data without unnecessary complexity. Rasmussen et al(2006)\cite{rasmussen_2006_gaussian}, states that models that are too simple underfit the data, while those that are too complex may overfit, capturing noise rather than the underlying function. The marginal likelihood naturally penalizes models that are too complex by incorporating the determinant of the covariance matrix, which grows with model complexity. 

The hyperparameters $\theta$ are typically optimized using gradient-based methods, given that the marginal likelihood is differentiable with respect to these parameters. This optimization process is a form of Bayesian model selection, where the model automatically balances fit and complexity to avoid over-fitting. For the zero-mean GP with the kernel $\alpha \cdot k(x,x')+ s\cdot I$, the optimization of $\alpha$ and s involves computing the gradient of the NLML with respect to these hyperparameters and iteratively updating them to minimize the NLML. However, in this research, we are not optimizing by minimizing the NLML but an integral part of composing our Exact GP framework (see Section \ref{subsection:custompackage}) and as a reminder for future work. In our setup as described in our Experimental Design section, we manually set these GP hyperparameters to the recommended values such as seen in Tripp et al (2021)\cite{tripp_2023_a}\cite{tripp_2024_diagnosing}.

\newpage
\section{Fingerprint-Based Kernels: Tanimoto \& MinMax Kernels}
Now, that we have discussed multi-output GPs, their theoretical underpinnings and hyperparameters, readers will recognise now that kernels are an integral part within the GP framework. In this section here, we discuss the type of kernel that is implemented within our algorithm and discuss why this is the recommended kernel in cheminformatics. Kernels are a powerful and flexible framework for measuring the similarity between data representations, particularly in fields of cheminformatics and bioinformatics \cite{ralaivola_2005_graph}. While traditional graph kernels are designed to work directly with graph structures, in this work, we focus on a specific subset of kernels that operate on molecular fingerprints - \textbf{a vectorized representation derived from molecular graphs} as described in Section \ref{section:fingerprints}. 

The Tanimoto and MinMax kernels (later introduced in Section \ref{subsec:TanimotoKernels} and \ref{subsec:MinMax}), have been traditionally employed as graph kernels in various computational chemistry and cheminformatics studies\cite{mah_2006_the}\cite{hisashi_2004_kernels}. These kernels were originally designed to measure the similarity between graphs by comparing specific substructures or features within the graphs, such as paths, walks, or other graph sub-components\cite{metzig_2012_graph}\cite{nikolentzos_2021_graph}. The work by Ralaivola et al(2005)\cite{ralaivola_2005_graph} introduced the Tanimoto kernel as a normalized variant that evaluates the similarity between two graphs based on the overlap of common features, and the MinMax kernel \cite{ralaivola_2005_graph}\cite{li_2015_minmax} modifies this approach to better handle variations in the feature distributions between different graphs. Both of these kernels have traditionally been used in the context of graph-based representations \cite{mah_2006_the}\cite{nikolentzos_2021_graph}\cite{sjoshuaswamidass_2005_kernels}\cite{young_2022_literature}. 

However, one of the fundamental challenges in working with graph data, is the problem of graph isomorphism, where two graphs $\mathcal{G}_1$ and $\mathcal{G}_2$ are considered identical if there exists a bijective mapping $f$ from the vertices of $\mathcal{G}_1$ to the vertices of $\mathcal{G}_2$ such that the edges are preserved\cite{babai_2016_graph}\cite{vert_2007_graph}. Determining whether two graphs are isomorphic is a computationally challenging problem, and no polynomial-time algorithm is known for general graph isomorphism, making it an NP-complete problem. In other words, the process of checking if two graphs are the same (graph isomorphism), is tricky as there is no fast straightforward way to do it as it requires a lot of computation\cite{babai_2016_graph}\cite{vert_2007_graph}. 

To address these challenges, molecular fingerprints offer a practical alternative by providing a vectorized summary of the graph's structure. This allows the application of fingerprint-based kernels, which compute a similarity measure between molecules based on these fingerprints rather than requiring direct comparison of the entire graph structures. This has recently been attempted by Tripp et al(2021)\cite{tripp_2023_a}, Tripp and Lobato(2024)\cite{tripp_2024_diagnosing} and Griffiths et al(2022)\cite{griffiths_2023_gauche}.

Fingerprint-based kernels, such as Tanimoto and MinMax kernels, map molecular fingerprints into a high-dimensional feature space where a similarity measure, generally the inner product, can be computed efficiently. The key idea is to define a kernel function $k(x,x')$ that captures the similarity between two molecules represented by their fingerprints $x$ and $x'$. These kernels are particularly effective because they can leverage structural information encoded in the fingerprints while avoiding the computational complexity associated with direct graph comparisons. 

To further understand how these fingerprint-based kernels function in practice, we now delve into the mathematical foundation of kernel methods using the concept of Reproducing Kernel Hilbert Spaces (RKHS)\cite{gretton_2019_introduction}\cite{xing_22}. The RKHS framework provides an understanding how kernel functions operate by mapping input data into high-dimensional spaces, enabling computation of similarity measures without explicitly performing the mapping. This approach is known as the \textit{kernel trick}\cite{gretton_2019_introduction}.

\subsection{Defining a Reproducing Kernel Hilbert Space by implicit mapping}
Reproducing Kernel Hilbert Spaces (RKHS) provide a powerful theoretical framework for kernel methods, which are central to many machine learning algorithms\cite{gretton_2019_introduction}\cite{xing_22}, including those in cheminformatics. In this section, we delve into the fundamentals of RKHS and illustrate how they are applied in fingerprint-based kernels like Tanimoto and MinMax kernels. 

A Hilbert space $\mathcal{H}$ is a complete inner product space, meaning that it is a vector space over a field of scalars (typically real or complex) equipped with an inner product $\langle \cdot, \cdot \rangle_{\mathcal{H}}$. The inner product induces a norm $||\cdot||_{\mathcal{H}} = \sqrt{\langle \cdot, \cdot \rangle_{\mathcal{H}}}$, which can be interpreted as a measure of length or distance between vectors. Completeness in this context means that every Cauchy sequence (a sequence where the distance between successive terms can be made arbitrarily small) converges to a point within the space. This ensures that the space has no \textit{\textbf{"gaps"}} and can support application of limit processes, which are critical in the analysis of functions and operators in infinite-dimensional spaces. 

\subsubsection*{Reproducing Kernel and Feature Maps}
A concept in RKHS is the reproducing kernel\cite{gretton_2019_introduction}, annotated generally as $k(x,y)$, which defines the inner product between elements in the space. 

\begin{definition}{\textbf{Hilbert spaces: }}
Given a set $\mathcal{X}$ and a Hilbert space $\mathcal{H}$ of real-valued functions on $\mathcal{X}$, the function $k: \mathcal{H} \cdot \mathcal{H} \rightarrow \mathbb{R}$ is a reproducing kernel if it satisfies the following properties for all $x,y \in \mathcal{X}$:
\begin{itemize}
    \item $k(x,\cdot)$ belongs to $\mathcal{H}$ for every $x \in \mathcal{H}$
    \item For every function $f \in \mathcal{H}$ and every $x \in \mathcal{X}$, the reproducing property holds: 
    \begin{equation*}
        f(x) = \langle f, k(x,\cdot) \rangle_{\mathcal{H}}
    \end{equation*}
\end{itemize}
\end{definition}
This reproducing property allows the evaluation of any function in $\mathcal{H}$ through an inner product with the kernel function. It implies that kernel function $k(x,y) = \langle k(x,\cdot), k(y, \cdot)\rangle$ not only acts as a measure of similarity between points $x$ and $y$ in the input space, but also represents the action of evaluating $f(x)$ any any x through the inner product in the Hilbert space $\mathcal{H}$. The RKHS framework is important in kernel methods as it allows the implicit mapping of input data into a high-dimensional feature space without explicitly computing the mapping. This is achieved through the kernel trick, which we will discuss in the next section. An example of this is molecular graphs are mapped into the Hilbert space here using their fingerprint vectors in the Figure \ref{fig:3} below.  

\begin{figure}[H]
    \centering
    \includegraphics[width=0.7\linewidth]{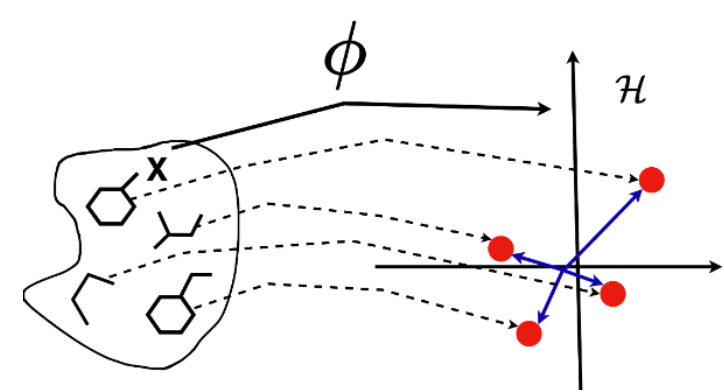}
    \caption[\textbf{Molecular structures mapped into the Hilbert Space $\mathcal{H}$ using fingerprint vectors (e.g. a binary digit string):}]{\textbf{Molecular structures mapped into the Hilbert Space $\mathcal{H}$ using fingerprint vectors (e.g. a binary digit string):}Molecular structures are represented as points in a high-dimensional feature space, where fingerprint-based kernels, such as Tanimoto and MinMax, compute the similarity between these molecular fingerprints. These kernels operate on the vector representation of molecules, capturing structural similarities implicitly in the Hilbert space $\mathcal{H}$, without requiring direct comparison of the original molecular graph structures}
    \label{fig:3}
\end{figure}

\subsection{Kernel Trick in Fingerprint-based Kernels}
The kernel trick is an essential concept in machine learning, particularly in the context of algorithms that involve similarity measures, such as Support Vector Machines (SVMs)\cite{mah_2005_graph}. The kernel trick revolves around the idea that many linear algorithms for regression or pattern recognition can be expressed solely in terms of inner products between feature vectors. Specifically let $\Phi(x)$ be a mapping that represents a molecular fingerprint $x$ in a high-dimensional feature space $\mathcal{H}$. The kernel function $k(x,x')$ can be defined as the inner product of the mapped vectors in this space:
\begin{equation*}
    k(x,x') = \langle \Phi(x), \Phi(x') \rangle_{\mathcal{H}}
\end{equation*}
The inner product computes the similarity  between two molecular fingerprints $x$ and $x'$ by comparing them in high-dimensional representations. However, directly computing the mapping $\Phi(x)$ for molecular fingerprints can be computationally inefficient, particularly when the feature space $\mathcal{H}$ is of very high or infinite dimensionality. From the Kernel Trick definition below, this allows us to bypass explicit computation of the mapping $\Phi$ by computing the kernel $k(x,x')$ directly, which is often significantly more efficient. 
\begin{definition}{\textbf{Kernel Trick:}}
Consider the computation of distances in the feature space $\mathcal{H}$ here, the distance between the feature representations of two fingerprints $x$ and $x'$ in $\mathcal{H}$ can be expressed as: 
\begin{equation*}
    d_{\mathcal{K}} (x,x')^2 = ||\Phi(x) - \Phi(x')||^2_{\mathcal{H}}
\end{equation*}
We have: 
\begin{equation*}
    d_{\mathcal{K}} (x, x')^2 = \langle \Phi(x), \Phi(x) \rangle_{\mathcal{H}} + \langle \Phi(x'), \Phi(x') \rangle_{\mathcal{H}} - 2\langle \Phi(x), \Phi(x' \rangle_{\mathcal{H}}
\end{equation*}
With kernel trick, this distance is computed as:
\begin{equation*}
    d_{\mathcal{K}} (x, x')^2 = k(x, x) + k(x', x') - 2k(x, x')
\end{equation*}
\label{def:kerneltrick}
\end{definition}
From this formulation, it is advantageous as it avoids to explicitly compute or store the high-dimensional feature vectors, allowing kernel methods to scale to large datasets and complex graph structures. For the kernel trick to be applicable, it is crucial that the kernel function $k(x, x')$ be a positive definite kernel (see Appendix \ref{section:PDK} for details on Positive Definite Kernels). 

Having laid the theoretical foundation of the kernel trick and reproducing kernels, we now turn our focus to specific kernels used in molecular similarity analysis: the Tanimoto and MinMax kernels. These fingerprint-based kernels are relevant to cheminformatics, where measuring the similarity between molecular fingerprints is essential for tasks in drug discovery and molecular optimization. Both the Tanimoto and MinMax kernels implement the kernel trick to compute similarity efficiently without explicitly mapping molecular structures into high dimensional spaces. As proven by Ralaivola et al (2005)\cite{ralaivola_2005_graph}, the Tanimoto kernel is a Mercer kernel, which means it satisfies Mercer's theorem and can be used within different frameworks of kernel-based learning methods like SVMs or Gaussian Processes (GPs).

\subsection{Tanimoto Kernels: Binary Morgan Fingerprints}
\label{subsec:TanimotoKernels}
The Tanimoto Kernel is particularly well-suited for binary Morgan fingerprints (explained in Section \ref{section:fingerprints}), which are designed to represent the presence or absence of specific molecular features within a molecule. These binary fingerprints are vectors of 0s and 1s, where each element signifies whether a particular molecular substructure or feature is present (1) or absent (0). The key advantage of using binary fingerprints lies in their ability to encode molecular structures compactly, allowing for efficient similarity calculations using kernel methods.

In cheminformatics, the Tanimoto kernel is a fundamental tool for quantifying the similarity between two molecular fingerprints by comparing shared features. This kernel focuses on the structural overlap between molecules, making it ideal for applications where shared substructures are of primary importance.

\subsubsection*{Defining the Tanimoto Coefficient}
At its core, the Tanimoto coefficient (sometimes referred to as the Jaccard index)\cite{ralaivola_2005_graph}\cite{jaccard_2015_jaccard}\cite{tanimoto_2019_t} provides a measure of similarity between two sets, in this case, the sets of features present in two molecules. Given two binary fingerprints $f_1$ and $f_2$, the Tanimoto coefficient is defined as: 
\begin{equation}
    T(f_1, f_2) = \frac{|f_1 \cap f_2|}{|f_1 \cup f_2|}
\label{eq:tanimotocoefficients}
\end{equation}
This formulation here expresses the ratio of the number of shared features (the intersection) to the total number of features present in either molecule. The Tanimoto coefficient takes values between 0 and 1: 
\begin{itemize}
    \item $T(f_1, f_2)= 1$ when $f_1$ and $f_2$ are identical (i.e. all features are shared).
    \item $T(f_1, f_2)= 0$ when $f_1$ and $f_2$ have no shared features. 
\end{itemize}
This measure is widely used in cheminformatics, particularly for comparing molecular structures encoded as binary fingerprints. When extended to kernels, the Tanimoto coefficient naturally forms the bases for the Tanimoto kernel, formally defined by Ralaivola et al(2005)\cite{ralaivola_2005_graph}, as:
\begin{equation}
    k_T(f_1, f_2) = \frac{k_{\varphi_d} (f_1, f_2)}{k_{\varphi_d} (f_1, f_1)+k_{\varphi_d} (f_2, f_2)- k_{\varphi_d} (f_1, f_2)}
\label{eq:tanimotokernel}
\end{equation}
where $k_{\varphi_d} (f_1, f_2)$ represents a dot product kernel between the feature maps of the two fingerprints, as explained below. 

\subsubsection*{Dot Product Kernels on Molecular Fingerprints}
Once the molecular fingerprints are transformed into vectors via the binary features map, the next step is to define a \textbf{dot product kernel} that quantifies the similarity between two fingerprints based on their binary representations. For two fingerprints $f_1$ and $f_2$, the dot product kernel is: 
\begin{equation*}
    k_{\varphi_d^{\text{bin}}} (f_1, f_2) = \sum_{p \in \mathcal{P}(d)} \mathbb{I} \{p \subseteq f_1\} \cdot \mathbb{I}\{p \subseteq f_2\}
\end{equation*}
This dot product kernel measures the number of shared features between two fingerprints, where the binary feature map $\varphi_d^{\text{bin}}$ is used to represent each features presence. Building on the dot product kernel, the Tanimoto kernel introduces normalization to account for the size of molecular fingerprints as defined in Equation \ref{eq:tanimotokernel}. The kernel's ability to normalize for size of fingerprints makes it a robust similarity measure in cheminformatics. It ensures that the fingerprints with a large number of features do not unduly bias the similarity score, providing a balanced comparison between different-sized molecules. Additionally, it has been proven by Ralaivola et al(2005)\cite{ralaivola_2005_graph}, that this kernel is positive semi-definite and satisfies the Mercer's Theorem (definition provided in Appendix \ref{mercertheorem}).

\subsection{MinMax Kernels: Count-based Fingerprints}
\label{subsec:MinMax}
The MinMax kernel \cite{li_2015_minmax}\cite{ralaivola_2005_graph} builds upon the Tanimoto kernel, focusing on count-based Morgan fingerprints rather than binary representations. Whereas the Tanimoto kernel compares molecular structures based on the presence or absence of substructures, the MinMax kernel quantifies the similarity between fingerprints by considering both the presence and multiplicity of substructures. This makes it particularly well-suited for cases where features may repeat across different substructures within a molecule. MinMax kernels are often employed when comparing more complex molecules that have varying degrees of feature repetition, providing a more nuanced similarity measure compared to the Tanimoto kernel.

\subsubsection*{Count-based Morgan Fingerprints}
Count-based Morgan fingerprints (as generally defined in Section \ref{section:fingerprints} extend the binary representations discussed in the Tanimoto kernel by counting the number of occurrences of a given substructure or feature in the molecule. Each fingerprint is transformed into a vector where each element represents the number of times a particular substructure (or feature) occurs. For a molecule fingerprint $f$, the count-based feature map is defined as: 
\begin{equation*}
    \varphi_{d}^{\text{count}} (f) = (\#\{p \subseteq f\})_{p \in \mathcal{P}(d)}
\end{equation*}
where $\#\{p \subseteq f\}$ represents the number of occurrences of a particular substructure p within the fingerprint $f$, and $\mathcal{P}(d)$ represents the set of possible features of length $d$. This vector captures not only the presence of features but also their multiplicity, which is essential for molecules where substructures may occur more than once. 

\subsubsection*{MinMax Kernel Definition}
The MinMax kernel $k_{\text{MinMax}}(f_1, f_2)$ is designed to compare two molecular fingerprints $f_1$ and $f_2$ by computing the ratio of the sum of the minimum values to the sum of the maximum values of the feature counts across all paths $p \in \mathcal{P}(d)$. This formulation takes into account the multiplicity of features in each fingerprint, providing a detailed similarity measure. 

For two fingerprints $f_1$ and $f_2$, the MinMax kernel is defined as: 
\begin{equation}
    k_{\text{MinMax}}(f_1, f_2) = \frac{\sum_{p \in \mathcal{P}(d)} \min (\#\{p \subseteq f_1\}, \#\{p \subseteq f_2\})}{\sum_{p \in \mathcal{P}(d)} \max (\#\{p \subseteq f_1\}, \#\{p \subseteq f_2\}}
\label{eq:MinMax}
\end{equation}
where: 
\begin{itemize}
    \item $\min (\#\{p \subseteq f_1\}, \#\{p \subseteq f_2\})$ represents the minimum number of occurrences of a substructure $p$ across two fingerprints. 
    \item $\max (\#\{p \subseteq f_1\}, \#\{p \subseteq f_2\})$ represents the maximum number of occurrences of the same substructure. 
\end{itemize}

This ratio ensures that the kernel reflects the proportional similarity between the two fingerprints, accounting for both shared substructures and their multiplicities. In scenarios where features are repeated within the molecules, such as larger analogues of repeated substructures, as shown in Figure \ref{fig:celecoxib}, the MinMax kernel offers a more accurate measure of molecular similarity than the Tanimoto kernel, which only considers binary presence or absence. 

\subsection*{Connection between MinMax and Tanimoto Kernel}
\begin{figure}[H]
    \centering
    \includegraphics[width=1.1\linewidth]{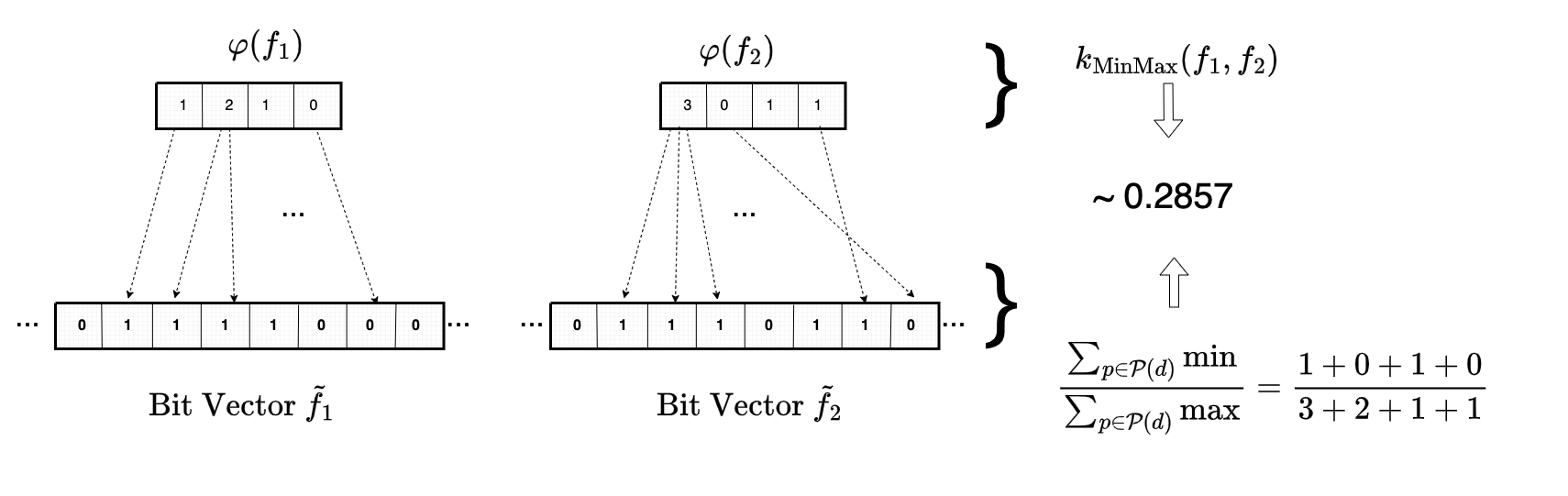}
    \caption[\textbf{The connection between MinMax and Tanimoto Kernels: }]{\textbf{The connection between MinMax and Tanimoto Kernels: }The feature vectors $\varphi{f_1}$ and $\varphi{f_2}$, representing two molecules, are transformed into their corresponding bit vectors $\tilde{f_1}$ and $\tilde{f_2}$, derived from the Morgan fingerprinting process (as depicted in Figure \ref{fig:1} and \ref{fig:2}. The MinMax kernel is calculated by taking ratio fo sum of minmum values to sum of maximum values across all paths $p \in \mathcal{P}(d)$. This is an extension of the Tanimoto kernel, where feature counts are considered.}
\label{fig:minmax}
\end{figure}
Formally, from Figure \ref{fig:minmax}, the feature vectors $\varphi{f_1}$ and $\varphi{f_2}$ represent count-based Morgan fingerprints, where the values indicate how many times a substructure appears in a molecule. The corresponding bit vectors $\tilde{f_1}$ and $\tilde{f_2}$ represent the binary Morgan fingerprinting interpretation (depicted in Figure \ref{fig:1} for the two molecules. The MinMax kernel uses the feature counts (top row) whereas the Tanimoto kernel operates on the binary information in the bottom row. 

The MinMax kernel can thus be considered an extension of the Tanimoto kernel, where binary presence or absence is replaced with feature counts, allowing for a finer-grained similarity measurement. In fact, the MinMax kernel reduces to the Tanimoto kernel when the fingerprints are strictly binary, where each features appears either once or not at all. The connection between the two lies in their shared structure - both compute a ratio of intersection to union, but differ in how they handle feature representation.

\newpage
\section{Multi-objective Bayesian Optimization}
In multi-objective optimization, we often aim to find a set of solutions rather than a single optimal solution due to the inherent conflict between objectives. These solutions form the Pareto front, representing the best trade-offs across the objectives. The core goal of Multi-Objective Bayesian Optimization (MOBO) is to efficiently approximate this Pareto front by leveraging Gaussian Process (GP) surrogate models that estimate the underlying objective functions.

Building on the foundation laid in the previous sections on Tanimoto kernels and GPs, this section delves into MOBO by utilizing the predictive uncertainty provided by GPs to guide the search for Pareto-optimal solutions.

However, unlike Single-Objective Bayesian Optimization (SOBO), where only one objective is optimized, MOBO tackles the complexity of real-world problems involving conflicting objectives. The search for a Pareto front necessitates more sophisticated acquisition functions that account for the trade-offs between objectives. This leads us to the Hypervolume Indicator (HVI)\cite{guerreiro_2021_the}(Section \ref{hypervolume-indicator}), which is critical for assessing the quality of Pareto approximations.

While Expected Improvement (EI) and other acquisition functions work well in SOBO, in the multi-objective setting, we require acquisition functions like Expected Hypervolume Improvement (EHVI)\cite{emmerich_2016_a}, which can identify new points that improve the Pareto front. EHVI operates by measuring the increase in the hypervolume dominated by the Pareto front. Hence, understanding hypervolume computations becomes essential in multi-objective settings.

In the following sections, we will dive deep into the Hypervolume Indicator (HVI) and its role in guiding MOBO towards better Pareto approximations. This detailed discussion is necessary because hypervolume-based methods, such as EHVI, are computationally intensive, especially as the number of objectives increases. Therefore, efficient algorithms for hypervolume calculation, like the Hypervolume by Slicing Objectives (HSO)\cite{while_2006_a}(Section \ref{subsection:HSO}) and Improved Dimension-Sweep (IDSA)\cite{fonseca_2006_an}(Section \ref{subsection:IDSA}) methods, are integral to ensuring the scalability of MOBO.

\subsection{Acquisition Functions}
\label{acquisitionfuncs}
In Bayesian Global Optimization (BGO), acquisition functions play a pivotal role in balancing exploration and exploitation by utilizing the uncertainty quantification provided by the Gaussian Process (GP) model. Common acquisition functions include Expected Improvement (EI), Probability of Improvement (PI), and Upper Confidence Bound (UCB), which guide the selection of the next points to evaluate by optimizing a criterion that considers both the predicted mean and uncertainty.

For single-objective optimization, Expected Improvement (EI) is widely used. It is computed as
\begin{equation}
    EI(x_*) = \mathbb{E}[\text{max}(0, f(x_*)- f_{\text{best}})]
\end{equation}
where $f_{\text{best}}$ is the best function value observed so far. The EI acquisition function encourages sampling in regions where the GP predicts high mean values and/or high uncertainty, thereby efficiently guiding the search towards the global optimum.

However, in the context of multi-objective optimization, where the goal is to find a Pareto-optimal front rather than a single optimal point, the EI approach becomes insufficient. In multi-objective settings, we require an acquisition function that can simultaneously handle multiple objectives and conflicting trade-offs. This is where the concept of Expected Hypervolume Improvement (EHVI) comes into play.

Proposed by Emmerich et al(2006)\cite{emmerich_2006_single}, EHVI extends the idea of EI by incorporating the Hypervolume Indicator (HV), which measures the region in the objective space dominated by the Pareto front. EHVI leverages both a Pareto-front approximation and the predictive uncertainty provided by the GP model to identify points that improve the Pareto front. While EI works well for single objectives, EHVI is designed to handle the computational complexities of multi-objective optimization, providing an effective balance between convergence and diversity across objectives.

In the following sections, we will discuss hypervolume computations in more detail, as these are critical for multi-objective acquisition functions like EHVI, which require efficient handling of high-dimensional Pareto fronts.

\subsection{Hypervolume Indicator}
\label{hypervolume-indicator}
In Multi-Objective Bayesian Optimization (MOBO), the Hypervolume Indicator (HV) is a vital metric for evaluating the performance of Pareto-front approximations. The HV indicator measures the size of the region in the objective space that is dominated by the Pareto front and bounded by a predefined reference point $r$\cite{guerreiro_2021_the}. It plays a key role in acquisition functions like Expected Hypervolume Improvement (EHVI), which we will introduce in due time.

\begin{definition}{\textbf{Hypervolume Indicator: }}
Given a set of points $P=\{y^{(1)},y^{(2)}...,y^{(n)}\} \subset \mathbb{R}^d$, the Hypervolume Indicator HV(P) is formally defined as the $d$-dimensional Lebesgue measure of the region dominated by P and bounded above by a predefined reference point r:
\begin{equation}
    HV(P) = \lambda_d \left(\bigcup_{y \in P} [y,r]\right)
\label{eq:hvcalculation}
\end{equation}
where $\lambda_d$ denotes the Lebesgue measure on $\mathbb{R}^d$, and $[y,r] = \{z \in \mathbb{R}^d |y \leq z \leq r\}$ represents the axis aligned hyperrectangle with diagonal corners at y and r. 
\end{definition}
Computing the HV indicator, specifically in higher dimensions, is computationally expensive. For dimensions $d \geq 2$, it can computed in $\mathcal{O}(n \log n)$ time, where n is the number of points in the set P. However, as the number of dimensions increases, complexity times grows exponentially. Despite the computational complexity, the HV indicator is often used in evolutionary multi-objective optimization algorithms (EMOAs)\cite{guerreiro_2021_the}\cite{yang_2019_multiobjective} as it is one of the few indicators that capture both convergence and diversity of solutions effectively. Additionally, the properties and relevance of the Lebesgue Measure \cite{meisters_1997_lebesgue} for the HV indicator are shown in Appendix \ref{Lebesgue-Measure}. 

\subsection{Pareto Points and Pareto Optimality}
\label{section:paretopoints}
In MOBO, Pareto optimality \cite{fleischer_2003_the} is crucial in navigating the trade-offs between conflicting objectives. No single solution can typically be considered "best" across all objectives, so we instead focus on finding solutions that improve some objectives without deteriorating others. These solutions form the Pareto-optimal front, representing the optimal trade-offs in the objective space.

\begin{definition}{\textbf{Non-dominated Solution Set: }}
This is the set of all solutions that are not dominated by any other solution in the decision space. Formally a solution $x^*$ is considered non-dominated (or Pareto-optimal) if there is no other solution x such that: 
\begin{equation*}
    f_i(x) \leq f_i(x^*)  \text{ for all } i \in \{1,2,...,m\}
\end{equation*}
and
\begin{equation*}
    \exists_j \in \{1,2,..,m\} \text{ such that } f_j(x) < f_j(x^*)
\end{equation*}
\end{definition}
Here, $f_1, f_2,...,f_m$ are the objective functions, and $x^*$ represents a decision vector in the feasible decision space. The pareto-optimal front consists of all such non-dominated solutions, providing a set of optimal trade-offs in the decision space. 

In the context of MOBO, finding the Pareto-optimal set is essential for constructing effective acquisition functions that balance multiple objectives. As we will see, acquisition functions like EHVI rely on this concept here to guide exploration and exploitation across conflicting objectives. 

\subsection{Hypervolume by Slicing (HSO) Algorithm}
\label{subsection:HSO}
Hypervolume is one of the most critical metrics for evaluating Pareto-front solutions in MOBO, as it captures both the diversity and convergence of solutions. However, computing hypervolumes, particularly in higher-dimensional objective spaces, presents significant computational challenges. This section introduces the Hypervolume by Slicing Objectives (HSO) algorithm by While et al(2006)\cite{while_2006_a}, a powerful method that addresses these challenges by efficiently calculating hypervolumes through lower-dimensional slices. Understanding this algorithm is crucial as it forms the backbone of hypervolume-based acquisition functions like Expected Hypervolume Improvement (EHVI), which are integral to MOBO strategies. The HSO algorithm here improves upon previous methods by focusing on processing objectives rather than individual points, which allows for significant reductions in computational complexity, especially in optimization problems with 3 or more objectives. 

HSO operates by slicing the objective space into hypervolumes of lower dimensionality, processing these slices individually, and then summing their contributions to compute the total hypervolume. This approach significantly reduces redundant calculations, particularly in higher-dimensional spaces, making HSO faster than other HV indicator methods, such as the LebMeasure algorithm. 

\begin{definition}{\textbf{Hypervolume by Slicing (HSO): }}
Let $S = \{x_1, x_2,...,x_m\}$ be a set of m mutually non-dominating points in n objectives, where each $x_i$ is a vector from $(x_{i1}, x_{i2},...,x_{in})$. The hypervolume $HV(S)$ is the measure of the union of the hyperrectangles defined by these points and a reference point $r = \{r_1, r_2,...,r_n\}$. The hypervolume is expressed as: 
\begin{equation*}
    HV(S) = \int_{\mathbb{R}^n} \mathbbm{1}_{\bigcup_{x \in S}R(x)}(z) \delta z
\end{equation*}
where $R(x)$ is the hyperrectangle dominated by x and bounded by r.
\label{def:HSO_slicing}
\end{definition}
HSO simplifies this by slicing the space along each objective, reducing the problem to a series of lower-dimensional hypervolume calculations. Specifically, after sorting the points by the first objective, HSO slices the hypervolume into sections, each corresponding to a distinct value of the first objective. Each section is then a hypervolume calculation in $n-1$ objectives, and the process is repeated recursively. For convenience, as the original authors have not provided a high-level pseudoalgorithm for HSO, this is presented here below in Algorithm \ref{alg:HSO}. A visual interpretation of the HSO algorithm is additionally provided below in Figure \ref{fig:HSO}. 

\begin{figure}[H]
    \centering
    \includegraphics[width=0.8\linewidth]{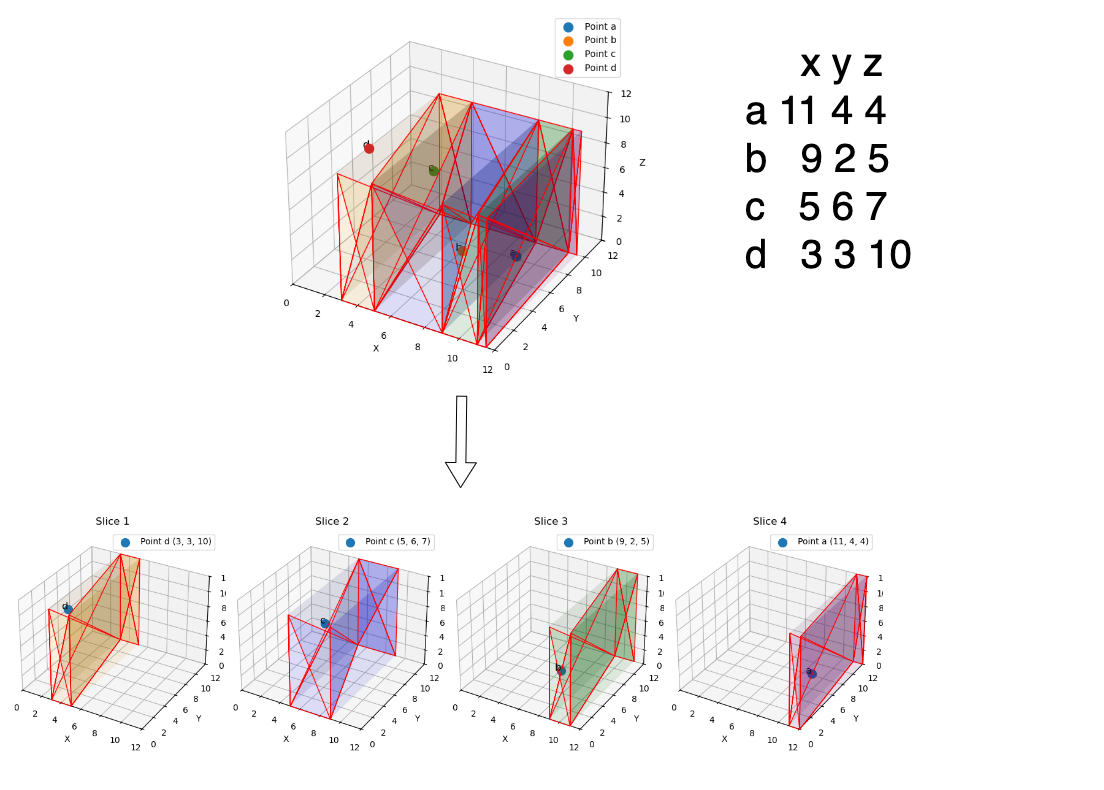}
    \caption[\textbf{Hypervolume by Slicing Objectives (HSO) applied to 4 three-objective points: }]{\textbf{Hypervolume by Slicing Objectives (HSO) applied to 4 three-objective points: }The 3D-space is sliced along the X-axis, generating two-objective shapes in the Y-Z plane for each slice. Points are labeled with their respective coordinates. The red lines represent the boundaries of each 3D slice, showing how the objective space is partitioned at each step in the HSO process.}
\label{fig:HSO}
\end{figure}

They key operation here, is the slicing of the objective space. We will attempt to provide a definition here: 
\begin{definition}{\textbf{Objective Space Decomposition in HV calculation: }}
At each step k, the list of points is sliced based on their values in the k-th objective. Each slice represents a section of the objective space where all points have a fixed value in the k-th objective, reducing the dimensionality by one. For a given slice at dimension k, the hypervolume contribution $V_k$ can be expressed as: 
\begin{equation*}
    V_k = (x_{(i+1)k} - x_{ik}) \times HV_{n-1}(S')
\end{equation*}
where $x_{(i+1)k}$ and $x_{ik}$ are the boundaries of the slice in the k-th objective, and $HV_{n-1}(S')$ is the hypervolume of the slice calculated in the $(n-1)-$dimensional space.
\end{definition}
This process is recursively applied, with each step reducing the problem by one dimension, until the final one-dimensional slices are summed to give the total hypervolume. 

In Algorithm \ref{alg:HSO} below, it begins by sorting the set S according to the values of the first objective $x_1$ in descending order. It then initializes a list L with a single entry containing the entire set of S and a multiplier of 1. The algorithm proceeds by iteratively slicing S across each objective k from 1 to $n-1$, where each slice generates sublists for the next dimension. The contribution of each sublist is computed by multiplying the width of that slice in the current dimension by the sublist's corresponding multiplier, and the results are accumulated in a new list $L'$. This process is repeated until only the final dimension remains, at which point the hypervolume is computed by summing the products of the multipliers and the widths of the slices in this last dimension.

\begin{algorithm}[H]
\caption{Hypervolume by Slicing Objectives (HSO) Algorithm (While et al(2006))}
\begin{algorithmic}
\State \textbf{Input:} Set of points $S$ in $n$ objectives, Reference point $r$
\State \textbf{Output:} Hypervolume $HV(S)$
\Procedure{HSO}{$S, n$}
    \State $S \gets \text{sort } S \text{ by Objective 1 descending}$
    \State Initialize $L \gets \{(1, S)\}$ \Comment{Each entry: (multiplier, point list)}
    \For{$k \gets 1$ to $n-1$}
        \State $L' \gets \{\}$
        \For{$(\text{m}, \text{pl})$ in $L$}
            \State $L' \gets L' \cup \text{slice}(\text{pl}, k, m)$
        \EndFor
        \State $L \gets L'$
    \EndFor
    \State $HV \gets \sum_{(\text{m}, \text{pl}) \in L} \text{m} \times (\text{head}(\text{pl})[n] - r[n])$
    \State \Return $HV$
\EndProcedure

\Procedure{slice}{$\text{pl}, k, \text{m}$}
    \State Initialize $S \gets \{\}$
    \State $p \gets \text{head(pl)}$, $pl \gets \text{tail(pl)}$
    \While{$pl \neq \emptyset$}
        \State $q \gets \text{head(pl)}$
        \State $S \gets S \cup \{(m \times |p[k] - q[k]|, \text{current\_slice})\}$
        \State $p \gets q$, $pl \gets \text{tail(pl)}$
    \EndWhile
    \State \Return $S \cup \{(m \times |p[k] - r[k]|, \text{current\_slice})\}$
\EndProcedure

\end{algorithmic}
\label{alg:HSO}
\end{algorithm}

\subsubsection*{Complexity Analysis}
While HSO offers significant computational efficiency improvements, particularly for problems with 3 or more objectives, it still faces scalability issues as the number of objectives grows. To address these, the Improved Dimension-Sweep Algorithm (IDSA)\cite{fonseca_2006_an} further refines the process by introducing advanced pruning techniques and the reuse of previous calculations. This not only reduces redundant computations but also ensures faster convergence for higher-dimensional problems.

\subsection{Improved Dimension-Sweep Algorithm}
\label{subsection:IDSA}
The Improved Dimension-Sweep algorithm (IDSA) by Fonseca et al(2006)\cite{fonseca_2006_an}, builds directly on the HSO framework. It incorporates key innovations in pruning and computational reuse that allow it to scale more effectively in higher-dimensional spaces. By reducing the number of redundant calculations and ensuring that previous hypervolume computations are reused where possible, IDSA offers a significant improvement in performance over traditional hypervolume calculation methods. The primary objective of this algorithm is to efficiently compute the HV indicator defined by Definition \ref{def:HSO_slicing} here for a set of $n$-non-dominated points in $d$ dimensions. These improvements are discussed in the Sections A, B, and C below. 

\subsubsection*{A) Recursive Dimension-Sweeping HV Calculation}
Based on Paquete et al (2006)\cite{paquete_2024_an}, the computation of the HV indicator can be acknowledged as a specialized instance of Klee's Measure Problem (detailed in Appendix \ref{section:KMP}). In this context of the HV indicator, a special case of the Klee's Measure problem is considered, where all hyperrectangles share the same lower vertex, typically by a reference point $r$.

\begin{definition}{\textbf{Recursive Dimension-Sweep Algorithm: }}
Let P be a set of points n points in $\mathbb{R}^d$ and $HV_d(P)$ denote the hypervolume of the region dominated by P with respect to a reference point r. The algorithm works by decomposing $HV_d(P)$ into lower-dimensional hypervolumes:
\begin{equation*}
    HV_d(P) = \sum_{i=1}^n(p_{i,d}- p_{(i+1),d}) \cdot HV_{d-1}(P_i)
\end{equation*}
where $p_{i,d}$ is the d-th coordinate of the i-th point, and $HV_{d-1}(P_i)$ is the hypervolume of the (d-1)-dimensional region dominated by the points $P_i$ in the slice corresponding to $p_{i,d}$.
\end{definition}
The computational complexity of this algorithm is $\mathcal{O}(n^{d-2} \log n)$ for general $d$-dimensional cases, which is a significant improvement over $\mathcal{O}(n^{d-1})$ complexity of non-recursive approaches such as HSO. For $d>3$, the algorithm first processes the highest dimension, decomposing the problem into a series of $(d-1)$-dimensional hypervolume calculations. Each of these calculations, in turn is further decomposed until the algorithm reaches the base case of 3 dimensions, where an optimized algorithm with $\mathcal{O}(n \log n)$ complexity is applied. 

\subsubsection*{B) Pruning the Recursion Tree}
The central idea behind pruning is to recognize when certain recursive branches can be safely ignored without affecting the final hypervolume calculation. Specifically, if a point $p$ in a higher-dimensional space dominates another point $q$ in all lower dimensions, the contribution of $q$ to the hypervolume in those dimensions becomes redundant. As a result, the algorithm can skip the recursive call associated with $q$. 
\begin{figure}[H]
    \centering
    \includegraphics[width=\linewidth]{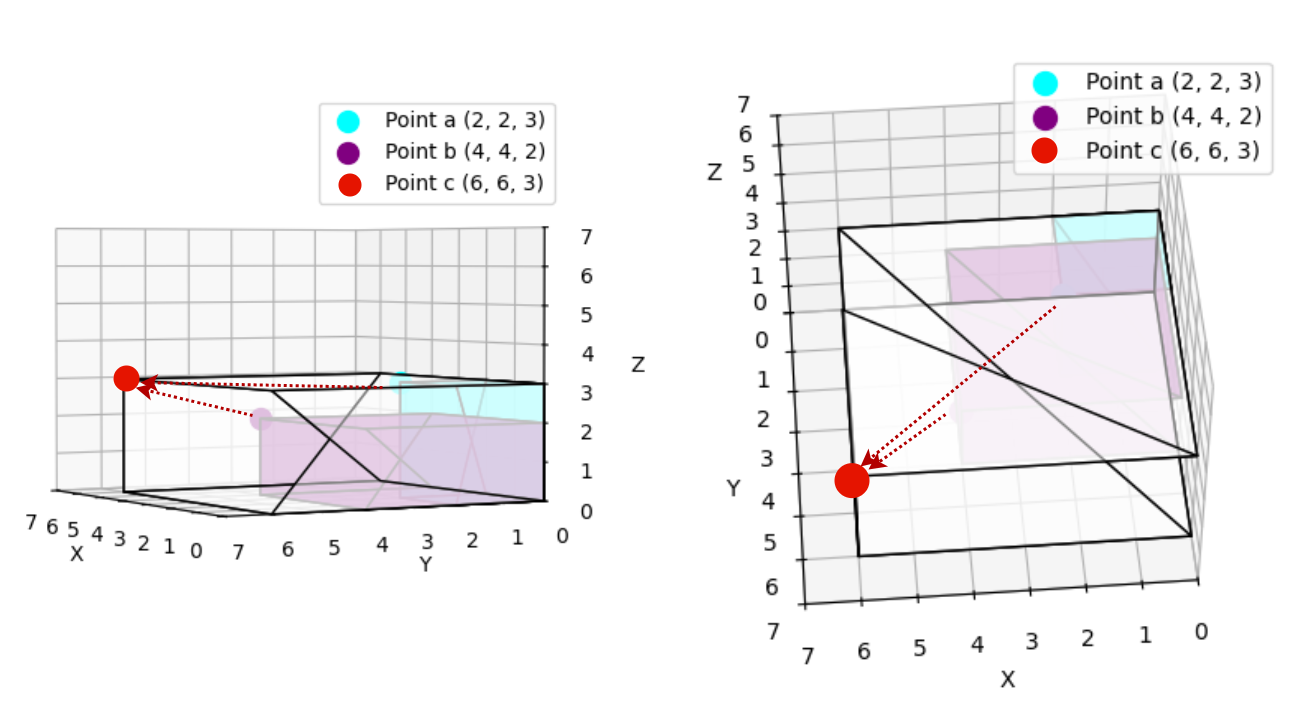}
    \caption[\textbf{Pruning the Recursion Tree in Recursive Dimension-Sweep Algorithm: }]{\textbf{Pruning the Recursion Tree in Recursive Dimension-Sweep Algorithm: }The hyperrectangles corresponding each respective point are shown, with the hypervolume in 3D space depicted by the rectangles' colors. The red arrow indicates the pruning process, where the hyperrectangles for Points A and B can be \textbf{ignored} in further calculations as it is dominated by Point C in the subsequent 2D slices.}
\end{figure}
\begin{definition}{\textbf{Pruning Recursion Tree Condition: }}
Let $P = \{p_1, p_2,...,p_n\}$ be a set of n non-dominated points in $\mathbb{R}^d$. The hypervolume contribution of a point $p_i$ is denoted by $HV_d(p_i)$. Specifically, for each point $p_i$, if there exists another point $p_j$, such that $p_j$ dominates $p_i$ in the subspaces corresponding to the last $d-1$ dimensions, the contribution $HV_d(p_i)$ becomes redundant and can be omitted. 

For $p_j$ to dominate $p_i$ in the last d-1 dimensions, it must hold that: 
\begin{equation*}
    p_{jk} \geq p_{ik} \forall k \in \{2,3,...,d\}
\end{equation*}
where $p_{jk}$ and $p_{ik}$ represent the coordinates of $p_j$ and $p_i$ in dimension k, respectively. If this condition is satisfied, hypervolume contribution of $p_i$ within subspace defined by dimensions $\{2,...,d\}$ is fully dominated by $p_j$, implies: 
\begin{equation*}
    HV_d(p_i) \subseteq HV_d(p_j)
\end{equation*}
\end{definition}
Recursive calculation for $p_i$ can be pruned/skipped, as it does not contribute to any additional volume to the total HV. 

\subsubsection*{C) Reusing Previous Calculations:}
Another further improvement is the reuse of previous hypervolume calculations. When processing the hypervolume of a $(d-1)-$polytope defined by the remaining points below a certain level in T, the computed hypervolume can be stored and reused when possible. This reduces the number of calculations needed, as the algorithm does not need to recompute the hypervolume for every recursive call. 

Let $V[p_i, j]$ be the stored hypervolume of the polytope defined by points below $p_i$ in dimension j. Then, the hypervolume at each level can be updated efficiently as: 
\begin{equation*}
    HV_d(p_i) = HV_d(p_{i+1}) + (p_{i,j}-p_{(i+1),j}) \times V[p_i, j]
\end{equation*}
This approach maintains a vector of bound values, b, which stores intermediate hypervolume calculations that can be quickly accessed and updated. 
 
Now, with these improved properties for the HSO algorithm discussed above, these are summarised in this Pseudoalgorithm \ref{alg:dimension-sweep} below, which is an adaptation and concise version of the improved dimension-sweep algorithm shown below. Note that the pseudo-algorithm described below is Version 4 from Fonseca et al (2006)\cite{fonseca_2006_an} which has not been directly described in detail in any of his existing papers. 

\begin{algorithm}
\caption{Dimension-Sweep Algorithm (Version 4)(adapted from Version 3 of Fonseca et al(2006))}
\begin{algorithmic}
\State \textbf{Input:} $d$ (dimensions), $P$ (non-dominated points), $r$ (reference point), $L_i$ (sorted list by dimension $i$), $len$ ($|L_{i-1}|$)
\State \textbf{Output:} $hvol$ (Hypervolume)

\Procedure{H}{$i, L_i, r, len$}
    \If{$i = 3$}
        \State Call the specialized 3D hypervolume function (see Fonseca et al (2006)\cite{fonseca_2006_an}); \Return
    \EndIf
    
    \State Reset flags for all $p$ in $L_i$
    \State $hvol \gets 0$, $p \gets nil(L_i)$
    
    \While{$prev_i(p) > b_i$ and $len > 1$}
        \State $p \gets prev_i(p)$, $b_j \gets \min\{b_j, p_j\}$ for $j < i$
        \State Delete $p$ from $L_i$, $len \gets len - 1$, $q \gets prev_i(p)$
    \EndWhile
    
    \If{$len > 1$}
        \State $hvol \gets V[prev_i(q), i] + H[prev_i(q), i] \cdot (q_i - prev_i(q))$
        \State $V[q, i] \gets hvol$
    \EndIf
    
    \State Call \textsc{SkipDom}$(q, i, L_i, r, len)$
    
    \While{$p \neq nil(L_i)$}
        \State $hvol \gets hvol + H[q, i] \cdot (p_i - q_i)$
        \State $b_i \gets p_i$, $b_j \gets \min\{b_j, p_j\}$ for $j < i$
        \State Reinsert $p$ into $L_i$, $len \gets len + 1$
        \State $q \gets p$, $p \gets next_i(p)$
        \State $V[q, i] \gets hvol$
        \State Call \textsc{SkipDom}$(q, i, L_i, r, len)$
    \EndWhile
    
    \State $hvol \gets hvol + H[q, i] \cdot (r_i - q_i)$
    \State \Return $hvol$
\EndProcedure

\Procedure{SkipDom}{$q, i, L_i, r, len$}
    \If{flag$[q] \geq i$}
        \State $H[q, i] \gets H[prev_i(q), i]$
    \Else
        \State $H[q, i] \gets H(i-1, L_{i-1}, r, len)$
        \If{$H[q, i] \leq H[prev_i(q), i]$}
            \State flag$[q] \gets i$
        \EndIf
    \EndIf
\EndProcedure
\end{algorithmic}
\label{alg:dimension-sweep}
\end{algorithm}

\subsubsection*{Complexity Analysis}
In Version 3, given in the Fonseca's paper\cite{fonseca_2006_an}, the time complexity is $\mathcal{O}(n^{d-1})$ as this version recursively handles each dimension down to 3 dimensions. For each dimension $i$, it performs operations involving iterations over the set of points, resulting in an $\mathcal{O}(n)$ complexity for each level. The lack of the pruning and reuse of previous calculations means that algorithm often recalculates intermediate hypervolumes. Further, as it sorts and handles points for each dimension, this implies each recursive call is $\mathcal{O}(n)$ and since there are $d-1$ recursive levels, the total complexity is $\mathcal{O}(n^{d-1})$. Version 4 above, there is an explicit step where the algorithm checks if the hypervolume for a given subset has already been calculated and stored in \textbf{$V[q,i]$}. The algorithm additionally  uses conditions \textbf{'flag[q]'} and \textbf{skipdom} (described in detail in Section 3C of Fonseca et al (2006)\cite{fonseca_2006_an}) to decide which parts can be skipped or reused, leading to fewer recursive calls and decrease in time complexity with $\mathcal{O}(n^{d-2})$.

\subsection{Expected Hypervolume Improvement (EHVI)}
\label{subsection:EHVI}
The algorithms discussed above HSO and ISA, offer efficient methods to compute the hypervolume. The significance of these algorithms extends beyond hypervolume calculation itself - they form the backbone of the Expected Hypervolume Improvement (EHVI)\cite{emmerich_2006_single} acquisition function in MOBO.

EHVI uses the hypervolume as a key indicator of progress in the optimization process. Specifically, EHVI measures the potential improvement in the hypervolume when a new solution is proposed, guiding the optimization process toward regions in the objective space that provide the most significant Pareto-front expansion. The computational efficiencies offered by HSO and IDSA are critical here, as EHVI relies on frequent and accurate hypervolume calculations to inform decision-making.

By efficiently computing the hypervolume, these algorithms enable EHVI to function effectively in high-dimensional objective spaces, ensuring that the optimization process is both computationally feasible and accurate. The next section delves deeper into the concept of EHVI and how it leverages hypervolume to drive multi-objective optimization toward the best possible trade-offs.

\begin{definition}{\textbf{Hypervolume Improvement (HVI): }}
Given a reference point \( r \in \mathbb{R}^d \) and a set \( P = \{y_1, y_2, \ldots, y_n\} \subset \mathbb{R}^d \), the hypervolume improvement when adding a point \( y \) to \( P \) is defined as: 
\begin{equation}
    HVI(P,y) = HV(P \cup \{y\}) - HV(P)
\end{equation}
where \( HV(P) \) is the hypervolume of the region dominated by the set \( P \) with respect to the reference point \( r \). The quantity \( HVI(P,y) \) measures the contribution of point \( y \) to expanding the hypervolume dominated by the Pareto front. 
\end{definition}

When the reference point \( r \) needs to be emphasized, the hypervolume improvement can be denoted as \( \Delta(P,y,r) \). This quantity is positive if \( y \) improves the Pareto front and contributes to the hypervolume. The improvement region is given by: 
\begin{equation*}
    \Delta(y,P,r) = \lambda_d \{z \in \mathbb{R}^d \mid y \prec z, z \prec r \text{ and } \nexists q \in P: q \prec z \}
\end{equation*}
where \( \lambda_d \) denotes the Lebesgue measure in \( \mathbb{R}^d \). 

\subsubsection*{EHVI Definition}
The EHVI extends the concept of HVI by accounting for uncertainty in the location of \( y \). This extension is particularly relevant in the context of Bayesian optimization, where predictions are modeled using Gaussian random fields (GRFs) or Gaussian Processes (GPs)(1-dimensional GRF)\cite{emmerich_2006_single}.

\begin{definition}{\textbf{EHVI: }}
Consider a vector of objective functions \( y \in \mathbb{R}^d \) whose distribution is governed by a multivariate Gaussian process \( \mathcal{N}(\mu(x), \Sigma(x, x')) \), where \( \mu(x) \) is the mean vector and \( \Sigma(x, x') \) is the covariance matrix determined by the covariance function \( k(x, x') \). The EHVI for adding \( y \) to the Pareto front \( \mathcal{P} \) is defined as: 
\begin{equation}
    EHVI(\mu(x), \Sigma(x, x'), P, r) = \int_{\mathbb{R}^d} HVI(P,y) \cdot PDF_{\mu, \Sigma} (y)\,dy
\label{eq:ehvi}
\end{equation}
where \( PDF_{\mu, \Sigma} (y) \) is the probability density function of the multivariate Gaussian distribution defined by \( \mu(x) \) and \( \Sigma(x, x') \). This integral represents the expected increase in hypervolume over all possible realizations of \( y \), weighted by their probability under the GRF model (described in Appendix \ref{def:GRFs}).
\end{definition}

Given the properties of a GRF (detailed in Appendix \ref{def:GRFs}), each individual point \( Y(x) \) follows a Gaussian distribution, and any subset of points also follows a multivariate Gaussian distribution. Specifically, the probability density function (PDF) for \( n \) points \( Y(x_1), Y(x_2), \ldots, Y(x_n) \) is given by: 
\begin{equation}
    P(Y) = \frac{1}{(2\pi)^{n/2}|C|^{1/2}} \exp \left(-\frac{1}{2}(Y-\mu)^T C^{-1}(Y-\mu)\right)
\label{eq:multivariate}
\end{equation}
where \( |C| \) is the determinant of the covariance matrix and \( C^{-1} \) is its inverse. 

In the context of EHVI, GRFs are relevant when considering uncertainty in predictions from surrogate models such as Gaussian Process Regression. The EHVI calculation integrates over the joint distribution of the objective functions, which under the assumption of Gaussian processes, is multivariate Gaussian. The integration in \ref{eq:ehvi}, given parameters \( \mu(x) \) and \( \Sigma(x, x') \) from Gaussian Process models, and Pareto-front approximation set \( \mathcal{P} \), accounts for the expected improvement in hypervolume considering uncertainty in the predictions.

\subsection{Numerical Integration of EHVI: Challenges and Complexity}
The computation of EHVI requires integrating over the possible realizations of $y$, given the uncertainty in the predictions from Gaussian Process models. Given the dimensional complexity of the problem, the integration can be decomposed for clarity as follows: 
\begin{equation}
    EHVI(\mu, \sum, \mathcal{P}, r)= \int_{\mathbb{R}^{d}}\left(\int_{\mathbb{R}^{d-1}} \cdot \cdot \cdot \left(\int_{\mathbb{R}^{1}} HVI(\mathcal{P}, y) \cdot PDF_{\mu, \sum}(y)\delta y_d \right) \delta y_{d-1} \cdot \cdot \cdot \delta y_2\right) \delta y_1
\label{EHVI-equation}
\end{equation}
The primary analytical challenges in computing the EHVI arises from the complexity of the integral itself, due to 3 reasons:
\begin{enumerate}
    \item \textbf{Non-linearity of the HVI function: } The hypervolume $HV(\mathcal{P})$ improvement in Equation \ref{eq:hvcalculation} depends non-linearly on $y$ because of the union of hyperrectangles defined by $\mathcal{P} \bigcup \{y\}$ which cannot be easily expressed in closed form, especially in high dimensions. 
    \item \textbf{Multivariate Gaussian Distribution: } The distribution involves the quadratic form $(y-\mu)^T\sum^{-1}(y-\mu)$ which represents the Mahalanobis distance from $y$ to the mean $\mu$ in the space defined by $\sum$. This is generally represented in Equation \ref{eq:multivariate} This creates a highly nontrivial integral that is difficult to solve analytically. 
    \item \textbf{High Dimensionality: } The number of Pareto-dominating and dominated regions grows exponentially with $d$. This increases number of regions over which integration must be performed. The covariance matrix $\sum$ grows in size with $d$ and the matrix inversion $\sum^{-1}$ becomes computationally expensive, scaling with $\mathcal{O}(d^3)$. The evaluation of the multivariate normal distribution involves calculating the determinant and inversion of $\sum$ which becomes increasingly difficult as $d$ increases. 
\end{enumerate}

\subsection{Monte Carlo Integration Method}
\label{subsection:MCintegration}
To address these challenges, Monte Carlo integration was proposed by Emmerich et al(2006)\cite{emmerich_2006_single} for a feasible approximation technique. Monte Carlo methods rely on generating a large number of random samples drawn from the Gaussian distribution and averaging the hypervolume improvement across these samples. This approach approximates the integral in \ref{EHVI-equation}. The Monte Carlo method breaks down the complexity by averaging the results over N samples $y_i$ drawn from the multivariate Gaussian distribution $\mathcal{N}(\mu, \sum)$.
\begin{equation*}
    EHVI_{MC} = \frac{1}{N}\sum_{i=1}^N HVI(P,y_i)
\end{equation*}
where $y_i \sim \mathcal{N}(\mu, \sum)$ are independent samples from the Gaussian process models. 

Monte Carlo methods offer a scalable solution for EHVI in high-dimensional spaces and are particularly suited for multi-objective optimization tasks where hypervolume computation becomes increasingly difficult. The key benefit of Monte Carlo integration is its ability to handle the non-linearity and high dimensionality in a flexible manner, although at the cost of requiring a potentially large number of samples for high precision.

In this chapter, we have covered the essential concepts of acquisition functions in Bayesian optimization, focusing on both single- and multi-objective optimization scenarios. We explored the transition from Expected Improvement (EI) to more sophisticated multi-objective acquisition functions, such as Expected Hypervolume Improvement (EHVI), which are necessary for balancing the exploration-exploitation trade-off across multiple objectives. We delved into computational methods such as the Hypervolume by Slicing Objectives (HSO) algorithm and its improvements, followed by the challenges posed by integrating EHVI analytically. To address these challenges, we discussed the Monte Carlo Integration method as a practical and scalable solution. This comprehensive review sets the stage for applying these advanced acquisition functions and optimization techniques in practical multi-objective optimization problems, ensuring efficient and robust search processes in high-dimensional spaces.

%%%%%%%%%%%%%%%%%%%%%%%%%%%%%%%%%%%%%%%%%%%%%%%%%%%%%%%%%%%%%%%%%%%%%%%%%%%%%%%%
\chapter{Methodology} \label{Chap4}
\section{KERN-GP: Achieving Full Dimensionality of Molecular Fingerprints}
\label{subsection:custompackage}
Molecular fingerprints are often represented as high-dimensional and extremely sparse vectors, with only 1-2\% non-zero elements. These vectors are commonly reduced in dimensionality to maintain computational efficiency. However, such reductions may lead to less accurate similarity measures. In contrast, \texttt{KERN\_GP} enables the exact calculation of Tanimoto coefficients (see Equation \ref{eq:tanimotocoefficients} of Section \ref{subsec:TanimotoKernels})
) across the full dimensionality of molecular fingerprints, ensuring greater precision in molecular similarity calculations.

Traditional implementations, particularly in frameworks like PyTorch, typically reduce the dimensionality to ranges between 1024-4096\cite{tripp_2023_a}\cite{karlova_2021_molecular}\cite{tazhigulov_2022_molecular}\cite{zhou_2019_optimization}], primarily because PyTorch expects dense matrices, which would result in excessive memory usage and computational overhead when handling sparse, high-dimensional vectors.

By leveraging the full dimensionality and using exact Tanimoto similarity calculations, \texttt{KERN\_GP} allows us to overcome the inefficiencies associated with dense matrix operations in PyTorch. This is particularly beneficial when working with large-scale molecular datasets, where dimensionality reduction could compromise the accuracy of downstream predictions.

The memory savings from using sparse representations instead of dense matrices are significant. As demonstrated in Table \ref{tab:fingerprint-sparsity-memory}, dense arrays can consume up to 2029 MB for Klekota-Roth fingerprints\cite{klekota_2008_chemical}, while their sparse representation reduces this to 23 MB - a 88.2x memory saving (Adamczyk \& Ludynia, 2024)\cite{adamczyk_2024_scikitfingerprints}. These savings are particularly impactful during tasks such as virtual screening, where full-scale fingerprints are required.

\begin{table}[H]
\centering
\resizebox{\textwidth}{!}{
\begin{tabular}{|l|c|c|c|}
\hline
\textbf{Fingerprint name}    & \textbf{Dense array size (MB)} & \textbf{Sparse array size (MB)} & \textbf{Memory savings} \\ \hline
Klekota-Roth                 & 2029                          & 23                              & 88.2x                 \\ \hline
FCFP                         & 855                           & 15                              & 57x                   \\ \hline
Physiochemical Properties     & 855                           & 17                              & 50.3x                 \\ \hline
ECFP                         & 855                           & 19                              & 45x                   \\ \hline
Topological Torsion           & 855                           & 19                              & 45x                   \\ \hline
\end{tabular}
}
\caption{Memory usage of fingerprints in dense and sparse versions (Adamczyk \& Ludynia (2024)}
\label{tab:fingerprint-sparsity-memory}
\end{table}

Given that \texttt{KERN\_GP} allows for exact Tanimoto coefficient calculations without dimension reduction, it ensures that all available information in the fingerprint is utilized. This provides a more faithful similarity measure compared to approaches that reduce dimensionality, potentially improving the accuracy of downstream predictions in molecular property estimation.

That said, it is important to note that while this implementation detail improves efficiency and accuracy, it does not fundamentally alter the results when compared to reduced-dimensional approaches. The added precision is incremental rather than transformative. Therefore, while \texttt{KERN\_GP} is a more exact method, the overall outcomes remain in line with conventional approaches, with the main advantage being the ability to scale up without sacrificing fidelity.

In summary, the key advantage of \texttt{KERN\_GP} lies in its ability to handle full-dimensional molecular fingerprints in an exact manner, without succumbing to the inefficiencies associated with dense matrix operations. For large-scale molecular datasets, this allows for more precise similarity calculations, which is particularly useful in tasks like virtual screening and similarity searching. However, it is important to recognize that this improvement primarily addresses the technical challenge of full-dimensionality rather than dramatically altering the predictive performance. What was developed is as demonstrated in Algorithm \ref{alg:minimalkernelonlyGP} is provided below here. 

\subsubsection*{Minimal Kernel-only GP Package for Fingerprints - Full Dimensionality}
\begin{algorithm}[H]
\caption{Kernel-Only Gaussian Process Inference for Fingerprints Full Dimensionality}
\label{alg:minimalkernelonlyGP}
\textbf{Input:} Training data $\mathbf{x}_i$, $\mathbf{y}_i$, GP hyperparameters $a$, $s$, Query SMILES $\mathbf{x}_q$ \\
\textbf{Output:} Predictive means and variances for query molecules
\begin{algorithmic}
    \State \textbf{Compute fingerprints} for $\mathbf{x}_i$ and $\mathbf{x}_q$
    \State \textbf{Compute kernel matrices} $K_{ii}$, $K_{iq}$, $K_{qq}$
    \State Perform \textbf{Cholesky decomposition} of $K_{ii} + \frac{s}{a} I$ \textcolor{blue}{using $\mathbf{L} = \text{Cholesky}(K_{ii} + \frac{s}{a} I)$ (see Appendix \ref{section:cholesky})}
    \State \textbf{Compute Negative Log Marginal Likelihood (NLML) \textcolor{blue}{(see Section \ref{section:NLML})}}:
    \State \hspace{1em} \textbf{a)} \textbf{Data fit term:} Calculate $-\frac{1}{2a} \mathbf{y}_i^\top \mathbf{L}^{-1} \mathbf{y}_i$
    \State \hspace{1em} \textbf{b)} \textbf{Complexity penalty term:} Calculate $-\frac{1}{2} \log \det (\mathbf{L}) - \frac{\log a}{2} |\mathbf{L}|$
    \State \hspace{1em} \textbf{c)} \textbf{Combine to get NLML:} $\text{NLML} = \text{data fit} + \text{complexity penalty} + \text{constant term}$
    \State \textbf{Compute predictive mean:} $\mu_*(\mathbf{x}_q) = K_{iq}^\top (K_{ii} + sI)^{-1} \mathbf{y}_i$
    \State \textbf{Compute predictive variance:} 
    \State \hspace{1em} \textbf{a)} \textbf{Covariance adjustment:} $\mathbf{V} = \text{Triangular\_Solve}(\mathbf{L}, K_{iq}^\top)$
    \State \hspace{1em} \textbf{b)} \textbf{Variance:} $\sigma^2(\mathbf{x}_q) = K_{qq} - \mathbf{V}^\top \cdot \mathbf{V}$
    \State \textbf{Return:} Predictive means $\mu_*(\mathbf{x}_q)$ and variances $\sigma^2(\mathbf{x}_q)$
\end{algorithmic}
\end{algorithm}

Additionally, it is good to note that the base package of this \texttt{KERN\_GP} can be modified to any base kernel we want. The GP model can also be modified to other GP variations, such as sparse GPs, and in our work, we coined the multi-output GP under this framework of \texttt{KERN\_GP}, as Multi-output Tanimoto Kernel GPs (MOTKGP).

\newpage
\section{State-of-the-art: \textbf{GP-MOBO}}
\label{section:gp-mobo}
In this work, we present a novel approach to multi-objective Bayesian optimization (MOBO), specifically tailored for molecular optimization using count fingerprints. The key innovation lies in leveraging Tanimoto Kernel Gaussian Processes (GPs), which were previously defined in Section \ref{subsection:custompackage} (\texttt{KERN-GP}). This GP-based MOBO framework models each molecular objective independently, unlike previous multi-objective works, which often propose complex models that assume correlations between outputs. Our approach provides a more scalable and computationally efficient solution, especially for large-scale molecular optimization tasks.

Unlike scalarization methods that combine multiple objectives into a single number, multi-objective problems more realistically capture the trade-offs between objectives without implicitly enforcing a specific preference. Scalarization methods, while commonly used, assume that trade-offs (such as preferring a=1, b=2 over a=2, b=1) are predetermined. However, real-world problems often involve uncertainties in these trade-offs, leading to a desire to explore the Pareto frontier directly, which captures all possible non-dominated solutions.

Despite the simplicity of this framework, it has not been investigated thoroughly in molecular optimization. In this thesis, we propose a straightforward algorithm that models each objective independently with Gaussian Processes (GPs), while utilizing a standard multi-objective acquisition function (EHVI) for efficient exploration and optimization. As a result, our \textbf{\textbf{GP-MOBO}} approach ensures scalability and computational efficiency, making it particularly well-suited for tasks that require optimizing multiple molecular properties simultaneously, such as virtual screening and molecular design. We will explain thoroughly and concisely what this algorithm entails in the subsequent sections below. 

\subsection*{Our Novel Algorithm for Multi-Objective Bayesian Optimization (\textbf{GP-MOBO})}
\begin{figure}[H]
    \centering
    \includegraphics[width=\linewidth]{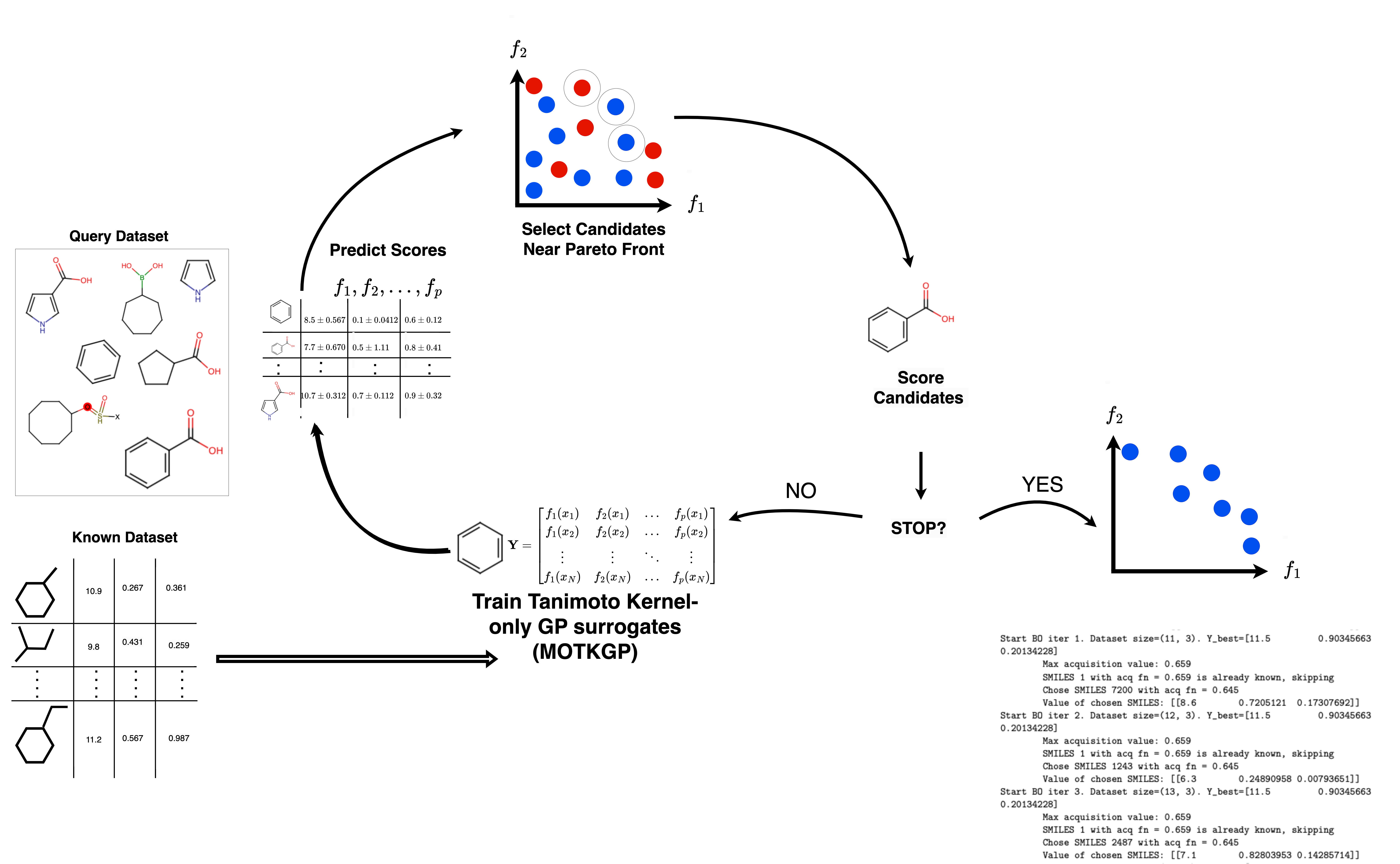}
    \caption[\textbf{Overview of our \textbf{GP-MOBO} algorithm: Combining independent Tanimoto Kernel GP surrogates with EHVI to guide molecular optimization, identifying optimal candidates near the Pareto frontier.}]{\textbf{Overview of our \textbf{GP-MOBO} algorithm: Combining independent Tanimoto Kernel GP surrogates with EHVI to guide molecular optimization, identifying optimal candidates near the Pareto frontier.} The process begins with an initial dataset of molecular structures (SMILES), where independent Gaussian Processes (GPs) for each objective are trained using the Tanimoto kernel. The figure illustrates the iterative optimization process. In each iteration, candidate molecules are selected based on their proximity to the Pareto front, scored for their objective values, and appended to the dataset. The GP is retrained after each iteration, refining the Pareto front until a stopping criterion is met. This iterative loop continues until the trade-offs between objectives are optimized, yielding non-dominated solutions.}
\label{nice-overview-gp-mobo}
\end{figure}

The core of our methodology is illustrated in Figure \ref{nice-overview-gp-mobo}, which provides an overview of the \textbf{\textbf{GP-MOBO}} process. This approach begins with an existing dataset of molecular SMILES/structures, and independent GP surrogate models are trained for each molecular objective. These GPs leverage the Tanimoto kernel, allowing us to capture molecular similarities across the full dimensionality of the fingerprint space, as implemented in the \texttt{KERN\_GP} framework. This is consequently extended to MOTKGP, which is the multi-output version defined below here in Definition \ref{MOTKGP}.

\begin{definition}[\textbf{Multi-Output Tanimoto Kernel Gaussian Processes (MOTKGP)}]
\label{MOTKGP}
For each molecular property \( f_1, f_2, \dots, f_k \), we define independent Gaussian Process (GP) models. The joint predictive distribution is multivariate Gaussian with a diagonal covariance matrix:
\[
\begin{bmatrix}
f_1(m) \\
f_2(m) \\
\vdots \\
f_k(m)
\end{bmatrix}
\sim \mathcal{N}
\left( 
\begin{bmatrix}
\mu_1(m) \\
\mu_2(m) \\
\vdots \\
\mu_k(m)
\end{bmatrix}
, 
\begin{bmatrix}
\sigma_1^2(m) & 0 & \dots & 0 \\
0 & \sigma_2^2(m) & \dots & 0 \\
\vdots & \vdots & \ddots & \vdots \\
0 & 0 & \dots & \sigma_k^2(m)
\end{bmatrix}
\right)
\]

Each property \( f_j \) is modeled with an independent GP:

\[
f_j \sim \mathcal{GP}(\mu_j, K_j(x_i, x_q))
\]

where \( \mu_j \) is the mean function and \( K_j(x_i, x_q) \) is the Tanimoto kernel, representing molecular similarity between fingerprints \( x_i \) and \( x_q \). The kernel is given by:

\[
k(x, x') = a_j T(x, x')
\]

with \( a_j \) as the kernel amplitude. Predictions for each property are returned as:

\[
\vec{\mu}(m) = [\mu_1(m), \dots, \mu_k(m)]
\quad \text{and} \quad
\vec{\sigma}^2(m) = [\sigma_1^2(m), \dots, \sigma_k^2(m)]
\]
\end{definition}

Following from the MOTKGP Definition \ref{MOTKGP}, we now transition to the core methodology of the Bayesian Optimization method of our proposed \textbf{\textbf{GP-MOBO}} algorithm. This algorithm (defined in Algorithm \ref{gpmoboalgorithm}) leverages Bayesian optimization (BO) within the context of multi-objective optimization, where each molecular property is modeled independently. In Bayesian Optimization, the idea is to balance exploration (finding diverse regions of the chemical space) with exploitation (focusing on regions that are known to yield high-quality results). In multi-objective optimization (MO), the objective is not to find a single optimal solution but to approximate the Pareto front, which consists of non-dominated solutions, representing optimal trade-offs between multiple objectives.

\subsubsection*{Problem Setup for \textbf{GP-MOBO}}
Given a dataset $\mathcal{D}_0 = \{(x_i, y_i)\}_{i=1}^n$ where $x_i$ are the SMILES strings and $y_i$ are the corresponding objective function values for each property $f_1, f_2,...f_D$, we aim to sequentially add new data points to $\mathcal{D}_0$ such that we refine our estimate of the Pareto front. 

We define each objective function as shown in MOTKGP framework above, and the kernel function which is the Tanimoto kernel (otherwise known as MinMax for count-based fingerprints), evaluates molecular similarity between fingerprints $x_i$ and $x_q$. 

\begin{algorithm}[H]
\caption{Our Proposed Novel Algorithm: \textbf{GP-MOBO}}
\label{gpmoboalgorithm}
\begin{algorithmic}

\State \textbf{Input:} Dataset $\mathcal{D}_0 = \{(\mathbf{x}_1, \mathbf{y}_1), \dots, (\mathbf{x}_n, \mathbf{y}_n)\}$, GP hyperparameters $\{\mu_j, a_j, s_j\}$, EHVI acquisition function $\alpha$, max reference point $R_{\max}$, scale $\lambda$, number of iterations $n_{\text{iter}}$
\State \textbf{Output:} Pareto-front approximation $\mathcal{P}$ and optimized SMILES

\State Initialize $\mu$ points $\{\mathbf{x}^{(1)}, \dots, \mathbf{x}^{(\mu)}\}$ from $\mathcal{D}_0$
\State Train independent MOTKGP model $p(\hat{f})$ on $\mathcal{D}_0$
\State Evaluate the initial set of $\mu$ points $\mathbf{y}(\{\mathbf{x}^{(1)}, \dots, \mathbf{x}^{(\mu)}\})$
\State Store evaluated points in $\mathcal{D}_0$
\State Compute initial Pareto front $\mathcal{P}_0$ from non-dominated subset of $\mathcal{D}_0$

\For{$t = 1$ to $n_{\text{iter}}$}
    \State Compute reference point $R \gets \text{infer\_reference\_point}(\mathcal{P}_{t-1}, R_{\max}, \lambda)$
    \State Calculate hypervolume $HV_{\text{current}} \gets \text{compute\_hypervolume}(\mathcal{P}_{t-1}, R)$

    \For{each $\mathbf{x_q} \in \text{Query SMILES}$}
        \State Predict mean $\mu_{\mathbf{x_q}}$ and variance $\sigma^2_{\mathbf{x_q}}$ using MOTKGP models
        \State Compute EHVI $\alpha(\mathbf{x_q})$
    \EndFor
    
    \State Select next candidate $\mathbf{x_*} \gets \arg \max_{\mathbf{x_q}} EHVI_{\mathbf{x_q}}$
    \State Acquire new objective $\mathbf{y_*} \gets \mathbf{y}(\mathbf{x_*})$
    \State Update dataset $\mathcal{D}_t \gets \mathcal{D}_{t-1} \cup \{(\mathbf{x_*}, \mathbf{y_*})\}$
    \State Update Pareto front $\mathcal{P}_t$ as non-dominated subset of $\mathcal{D}_t$
    \State Recalculate hypervolume $HV_{\text{new}} \gets \text{compute\_hypervolume}(\mathcal{P}_t, R)$

    \If{budget exhausted}
        \State \Return $\mathcal{P}_t$ \Comment{Terminate if budget is exhausted}
    \EndIf
\EndFor

\State \Return Pareto-front approximation $\mathcal{P}_t$ and optimized SMILES

\end{algorithmic}
\end{algorithm}

\subsubsection*{Bayesian Optimization in \textbf{GP-MOBO}}
The Bayesian Optimization loop starts by predicting objective values using the trained GP surrogates. The Expected Hypervolume Improvement (EHVI) acquisition function guides the selection of new query molecules, balancing exploration and exploitation.

The EHVI function here (discussed in Section \ref{subsection:EHVI}) is designed to compute the expected gain in the hypervolume (HV) of the Pareto front. This improvement select candidates that will best expand the Pareto front in subsequent iterations. Formally, for each query point (SMILES) $x_q$, EHVI is computed as: 
\begin{equation*}
    EHVI(x_q) = \mathbb{E}[\max(HV_{x_q} - HV_{\text{current}}, 0)]
\end{equation*}
where $HV_{x_q}$ is the hypervolume of the Pareto front where new query SMILES $x_q$ is added, and $HV_{\text{current}}$ is current hypervolume of Pareto front. 

As illustrated in Algorithm \ref{gpmoboalgorithm}, the \textbf{GP-MOBO} process starts by initializing with a dataset $\mathcal{D}_0$ with known SMILES and their corresponding objective values $y_i$. The non-dominated solutions are computed from $\mathcal{D}_0$ forming the initial approximation of the Pareto front $\mathcal{P}_0$. At each iteration t, the GP models predict the mean and variance $\mu_{x_q}, \sigma^2_{x_q}$ for each query molecule $x_q$. These predictions are fed into the EHVI acquisition function to score each query molecule for its potential to improve the Pareto front. The molecule with the highest EHVI score is selected and its true objectives are evaluated and the dataset $\mathcal{D}_t$ is updated. The Pareto front is recalculated and hypervolume is updated accordingly. The loop continues until the computational budget is exhausted or the desired Pareto front is sufficiently refined. The final Pareto front is returned, providing the optimal trade-offs across all molecular objectives. 

The core mathematical operations - training of independent GPs, hypervolume computation, and acquisition function evaluation are all defined within the Background Chapter in Sections \ref{eq:hvcalculation}, \ref{hypervolume-indicator}, \ref{section:paretopoints}, \ref{subsection:HSO}, and \ref{subsection:EHVI}, where we discuss how this entire optimization pipeline is computationally efficient. It is also good to acknowledge that our implementation of EHVI, has been tested with the readily available test cases that have been provided by BoTorch and have passed with the same numerical accuracy. The test case results are as shown in the Appendix \ref{ehvi-test-cases-section}.

Now that we have thoroughly explained the design and technicalities of the \textbf{GP-MOBO} algorithm, including how it efficiently handles multi-objective optimization using independent GP surrogates and EHVI as the acquisition function, we can proceed to experimental validation. In the following chapter, we will benchmark our \textbf{GP-MOBO} model against single-objective approaches, highlighting the benefits and trade-offs of modeling objectives independently. Through a series of experiments, we aim to demonstrate the scalability, computational efficiency, and accuracy of our method when applied to real-world molecular optimization tasks.

%%%%%%%%%%%%%%%%%%%%%%%%%%%%%%%%%%%%%%%%%%%%%%%%%%%%%%%%%%%%%%%%%%%%%%%%%%%%%%%%
\chapter{Experimental Design} \label{Chap5}
\section{Datasets}
For benchmarking GP-MOBO, we use two widely referenced datasets: DockSTRING and GUACAMOL. Both are frequently used in molecular optimization tasks and are considered standard benchmarks in the literature \cite{gao_2024_sample}\cite{tripp_2023_a}.
\begin{itemize}
    \item DockSTRING\cite{garcaortegn_2021_dockstring}: This dataset offers a robust framework for docking and binding affinity prediction tasks. We use a subset of first 10000 SMILES, where for the toy MPO setup, we begin with a set of 10 initial \texttt{known\_SMILES}, and GP-MOBO will sample \texttt{query\_SMILES} from the remaining dataset over 20 Bayesian Optimization iterations.
    \item GUACAMOL\cite{brown_2019_guacamol}: This dataset focuses on de novo molecular design, targeting drug-likeness, novelty, and synthetic accessibility. For our setup, we also use a subset of 10000 SMILES from the \texttt{guacamol\_v1\_train.smiles} file. We trained all the benchmarking models and \textbf{GP-MOBO} with 10 initial \texttt{known\_SMILES}, and the models will sample from the $\sim 9980$ \texttt{query\_SMILES} for the next 20 Bayesian Optimization iterations. Further, for assessing the GP's prediction, we use \texttt{guacamol\_v1\_valid.smiles}.
\end{itemize}

\section{Oracles}
In our experimental design, oracles serve as evaluation functions that mimic real-world drug discovery tasks. Each oracle computes specific molecular properties or docking scores for a given SMILES string $x$ and returns the corresponding objective values $y$. Our utility functions handle these evaluations, ensuring that only valid values are processed. These evaluations are implemented in our utility functions (see example in Appendix \ref{oracle}) , which handle multiple objectives for each dataset, filtering out any $NaN$ values and ensuring consistency in the evaluation. For the toy MPO setup (see Section \ref{toy-setup} below), we defined 3 distinct objectives to be optimized concurrently. For more complex, real-world drug discovery tasks, we specifically chose 3 GUACAMOL's MPO tasks, Fexofenadine MPO, Amlodipine MPO, and Perindopril MPO. These GUACAMOL MPO setup is clearly demonstrated after our Toy MPO Setup. These MPO definitions are provided in Appendix \ref{molecularobjectivedefinition}. 

\section{Toy Multi-Property Objective (MPO) Setup}
\label{toy-setup}
To validate our GP-MOBO model, we devised a toy experiment using a set of toy multi-property objectives from the DockSTRING dataset. The chosen objectives were selected to challenge our model with diverse molecular properties, aiming to balance and optimize conflicting criteria simultaneously. The selected objectives are:
\begin{equation}
    f_1(m) = -\text{DockingScore}(\text{PPARD}, m)
\end{equation}
\begin{equation}
    f_2(m) = \text{QED}(m)
\end{equation}
\begin{equation}
    f_3(m) = -\text{sim}(m, \text{celecoxib})
\end{equation}
The definitions of these objectives are as described in Appendix \ref{molecularobjectivedefinition}. These objectives target molecules that bind to the PPARD protein, are drug-like, and are structurally similar to the reference molecule, celecoxib. Though this setup does not represent a realistic drug discovery task, it serves as a demonstrative example for evaluating our methodology.

For each optimization step, we begin with a small set of molecules $10$ where all objective values have been observed (with some Gaussian noise). These molecules (represented as SMILES) form our \texttt{known\_SMILES} list. The corresponding objective values are stored in the array \texttt{known\_Y}, which has the shape $(N,3)$ where N represents the number of molecules and 3 corresponds to the objective values of $f_1$, $f_2$ and $f_3$. To compute these objective values, we have the \texttt{evaluate\_objectives()} function which ensures the handling of \texttt{NaN} values and provides consistency in input-output mapping. As the GP model is trained independently for each objective (see Definition \ref{MOTKGP}), we have the specific hyperparameters used for this setup here, for 3 independent objectives in Table \ref{gphyperparametersdockstring}.
\begin{table}[h!]
\centering
\begin{tabular}{|c|c|c|c|}
\hline
\textbf{GP Hyperparameter} & \textbf{$f_1$} & \textbf{$f_2$} & \textbf{$f_3$}\\ \hline
GP Mean ($\mu$)                 & 0.0     &    0.0      &     0.0   \\ \hline
GP Noise ($s$)                  & $1e^{-4}$  & $1e^{-4}$ & $1e^{-4}$ \\ \hline
GP Amplitude ($\alpha$)         & 1.0 & 1.0 & 1.0  \\ \hline
\end{tabular}
\caption{\textbf{GP Hyperparameters for Seed Prototype Model of GP-MOBO implemented on DockSTRING dataset: } These hyperparameters were specifically chosen for comparison with the original GP-BO model which have these hyperparameters (Tripp \& Hernandez-Lobato(2024))}
\label{gphyperparametersdockstring}
\end{table}
For the Bayesian Optimization (BO) process, we utilize Expected Hypervolume Improvement (EHVI) as the acquisition function to guide the selection of new SMILES strings that will improve the current objectives $f_1(m)$,$f_2(m)$ and $f_3(m)$. As computing EHVI in closed form can be computationally intractable for multi-objective optimization, we approximate it using Monte Carlo (MC) integration. In our setup, we employ 1000 Monte Carlo samples to estimate the EHVI at each step of the optimization. These MC samples are drawn from the posterior distribution of the objectives, providing an efficient and scalable way to explore the objective space and identify promising candidates. By doing so, we ensure that the BO process can effectively balance the trade-offs between objectives while progressively improving the molecular properties of the generated SMILES strings over 20 iterations.

As a result, 20 chosen SMILES will then be evaluated using \texttt{evaluate\_objectives()} function which will return their $f_1$, $f_2$ and $f_3$ values respectively. 

\section{GP-MOBO on GUACAMOL's MPO Setup}
For the GUACAMOL MPO Setup, we extend our GP-MOBO evaluation to real-world drug discovery tasks by focusing on three distinct multi-property objectives (MPOs) from the GUACAMOL dataset (Table \ref{guacamol:mpo}).
\begin{table}[H]
\centering
\begin{tabular}{|l|l|l|l|}
\hline
\textbf{MPO Task} & \textbf{Mean} & \textbf{Scoring Function(s)} & \textbf{Modifier} \\ \hline
\multirow{3}{*}{Fexofenadine MPO} & \multirow{3}{*}{geom} & sim(fexofenadine, AP) & Thresholded(0.8) \\ \cline{3-4} 
 & & TPSA & MaxGaussian(90, 2) \\ \cline{3-4} 
 & & logP & MinGaussian(4, 2) \\ \hline
\multirow{2}{*}{Amlodipine MPO} & \multirow{2}{*}{geom} & sim(amlodipine, ECFP4) & none \\ \cline{3-4} 
 & & number rings & Gaussian(3, 0.5) \\ \hline
\multirow{2}{*}{Perindopril MPO} & \multirow{3}{*}{geom} & sim(perindopril, ECFP4) & none \\ \cline{3-4} 
 & & number aromatic rings & Gaussian(2, 0.5) \\ \hline
\end{tabular}
\caption{Selected Guacamol's MPO Tasks for Benchmarking GP-MOBO performance with GP BO}
\label{guacamol:mpo}
\end{table}
In this section here, we illustrate how GP-MOBO tackles this MPO task. As an example, we focus on Fexofenadine MPO (Table \ref{guacamol:mpo}), the selected objectives are: 
\begin{equation}
    f_1(m) = -\text{sim}(m, \text{Fexofenadine}, AP)
\end{equation}
\begin{equation}
    f_2(m) = \text{TPSA}(m)
\end{equation}
\begin{equation}
    f_3(m) = \log P(m)
\end{equation}
It's important to note that we modified the original Fexofenadine MPO oracle, which originally combined the objectives above here by scalarizing into a single multi-property score with the geometric mean. We split it into \textbf{3 separate objectives} for these  experiments. This modification allows us to evaluate and optimize each molecular property individually, increasing interpretability of model's performance on specific attributes. For this task, the GP hyperparameters are as standardized from the Toy MPO setup (Table \ref{gphyperparametersdockstring}), and the number of Monte Carlo (MC) samples to approximate the EHVI at each step maintains the same at $N=1000$. Now, with this setup, we expect, for this Fexofenadine task, for 20 BO iterations, 20 new SMILES will be selected that maximize these properties independently. These 20 chosen SMILES will then be evaluated using \texttt{evaluate\_objectives()} function which will return their $f_1$, $f_2$ and $f_3$ values respectively. 

This approach is similarly applied to the other two tasks, Amlodipine MPO and Perindopril MPO, each with their respective scoring functions and modifiers, as outlined in Table 4.2. For Amlodipine MPO and Perindopril MPO, both MPOs will return $f_1$ and $f_2$. This section sets the stage for the evaluation and comparison of GP-MOBO's performance across these tasks with 3 other derivatives of the single-objective GP-BO model by Tripp et al(2021). 

\section{Benchmarking GP-MOBO}
The purpose of our experimental design is to systematically compare the performance of the GP-MOBO algorithm against existing single-objective Bayesian optimization models in both low and full fingerprint dimensionalities. This design evaluates how single- and multi-objective optimization techniques perform on a variety of molecular optimization tasks, using acquisition functions such as UCB, EI, and EHVI. The workflow is divided into three main stages: initial SMILES selection, fingerprint dimensionality preprocessing, and Bayesian optimization.

\begin{figure}[H]
    \centering
    \includegraphics[height=0.81\textheight]{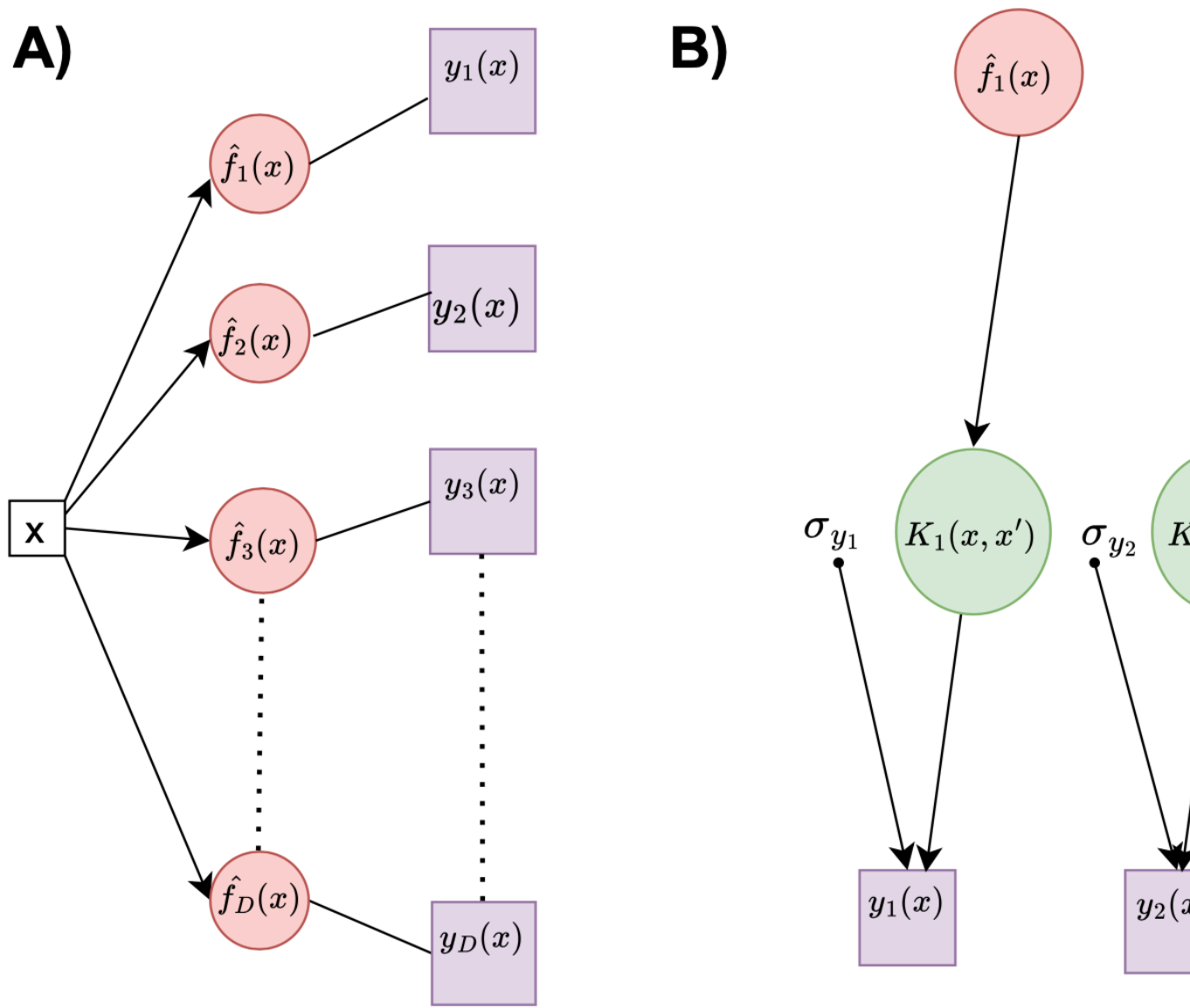}  % Scale by height
    \caption[\textbf{Experimental Design Workflow:}]{\textbf{Experimental Design Workflow:} This design compares SMILES values using both single- and multi-objective optimization. We define objectives $f_1, f_2, f_3$, followed by preprocessing SMILES data with either full fingerprint dimensionality (\texttt{FP\_DIM = FULL}) or reduced default dimensionality(\texttt{FP\_DIM = 2048}). Whether scalarization (geometric mean) is applied, either a single-output GP (TKGP) or multi-output GP (MOTKGP) is trained. Acquisition functions EI, UCB, \& EHVI are then used to identify and evaluate new SMILES until termination. Final SMILES are selected via UCB-PT (GP-BO), EI-PT, KERN-GP-EI, or KERN-GP-EHVI (GP-MOBO).}
    \label{fig:summary-experimental-design}
\end{figure}

\subsection*{Initial SMILES selection}
For both the Toy MPO setup and the GUACAMOL MPO Tasks, we randomly selected 10 initial \texttt{known\_SMILES}. These initial molecules serve as a starting point for the optimization process. For each optimization task, we evaluate the selected SMILES based on their scalarized MPOs (for single-objective cases) or independent objective functions (for multi-objective cases), represented by $f_1$, $f_2$ and $f_3$. This ensures that the optimization process begins from the same baseline across all experiments. The initial objective values are consistent in both setups, reducing any bias and allowing for fair comparisons.

\subsection*{Count Fingerprint Dimensionality Processing}
The selected SMILES are converted into molecular fingerprints using two approaches:
\begin{itemize}
    \item Full Dimensionality (\texttt{FP\_DIM = FULL}): We employ our \texttt{KERN\_GP} model to preserve the full dimensionality of count molecular fingerprints. This configuration allows the model to leverage the complete information encoded in the molecular structures.
    \item Reduced Dimensionality (\texttt{FP\_DIM = 2048}): We apply the default reduced fingerprint size (2048 bits), a common configuration used in most PyTorch (PT)-based model implementations.
\end{itemize}
This step enables us to investigate how variations in fingerprint dimensionality affect the performance of Bayesian optimization across different molecular complexities.

\subsection*{Bayesian Optimization Setup}
The optimization process is divided into two branches based on the task's objectives:

\textbf{Single-Objective Optimization:} In these cases, the objectives are scalarized using the geometric mean of $f_1$, $f_2$, and $f_3$, reducing the optimization task to a single scalar value. We employ two acquisition functions:
\begin{itemize}
    \item \textbf{Upper Confidence Bound (UCB):} Optimizes for the upper bound of confidence intervals around expected values.
    \item \textbf{Expected Improvement (EI):} Focuses on maximizing the expected improvement over the current best-known SMILES.
\end{itemize}
These configurations are referred to as \textbf{UCB-PT} and \textbf{EI-PT} respectively when processed with \texttt{FP\_DIM = 2048}, and as \textbf{KERN-GP-EI} when processed with full-dimensional fingerprints.

\textbf{Multi-Objective Optimization (GP-MOBO):} In contrast to scalarization, the GP-MOBO framework optimizes multiple objectives simultaneously, leveraging the EHVI acquisition function to balance trade-offs between objectives $f_1$, $f_2$, and $f_3$. This approach allows the model to efficiently explore the Pareto front of the optimization task, ensuring a diverse set of optimal molecules.

\subsection{Model Training}
The GP-MOBO model is trained using the Tanimoto kernel on full-dimensional \textbf{count} fingerprints (MinMax Kernel (Section \ref{subsec:MinMax}), allowing the model to directly optimize for multiple objectives in parallel. After each Bayesian Optimization iteration, new SMILES strings are selected, and their objective values are updated in the \texttt{known\_SMILES} list. The same process is applied to the single-objective models, with UCB, EI, and scalarized objectives. Training proceeds for \textbf{20 BO iterations} across all setups, ensuring a comprehensive and consistent evaluation. We additionally evaluated the GP's predictions with negative log predictive density (NLPD) (see Appendix \ref{section:NLPD}).

\subsection{Comparison}
To fairly benchmark the performance of the GP-MOBO model, we compare it against the baseline GP-BO models (Tripp et al., 2024)\cite{tripp_2024_diagnosing}, which employ UCB-based acquisition functions and reduced fingerprint dimensionality (FP\_DIM = 2048). Additionally, we extend the original GP-BO to include the KERN-GP-EI method, where we maintain full fingerprint dimensionality for comparison. This setup allows us to isolate the effects of dimensionality and acquisition function on model performance. Furthermore, as the original GP BO that optimizes the GP acquisition function with Graph GA methods in an inner loop \cite{tripp_2023_a}, this was modified to just sample from the query dataset (\texttt{query\_SMILES}) for both setups to make a fair comparison. 

\subsection{Evaluation Procedure}
These methods are evaluated in terms of how effectively they balance multiple objectives or optimize scalarized objectives in the BO process. This setup ultimately allows us to systematically benchmark the GP-MOBO model and compare its performance in terms of 20 chosen BEST SMILES from \texttt{query\_SMILES}.

To conclude our experimental design, we outline the evaluation procedure across all methods. For the single-objective methods (UCB-PT, EI-PT, KERN-GP-EI), the initial SMILES were scalarized using the geometric mean of the three objectives ($f_1$, $f_2$, $f_3$) to guide their selection process. To evaluate how these single-objective acquisition functions perform in terms of approaching the Pareto front, the 20 SMILES chosen by these methods were re-evaluated independently, as performed with the multi-objective EHVI method. This allowed us to investigate how well the single-objective approaches balance the conflicting objectives when compared to the EHVI method.

Additionally, we sought to determine if the GP-MOBO method could select better SMILES than the single-objective methods. For this, the 20 SMILES chosen by each method were scalarized again, calculating the geometric mean of their respective $f_1$, $f_2$, and $f_3$ values to provide a holistic measure of performance. The formula for calculating the geometric mean across the three objectives is given by:
\begin{equation*}
    \text{Geometric Mean} = \left(\Pi_{i=1}^{n}x_i\right)^{\frac{1}{n}} = (f_1(m) \times f_2(m) \times f_3(m))^{\frac{1}{3}}
\end{equation*}

In the results that follow, we present a detailed comparison of the performance of these methods. Specifically, we show how the geometric mean of $f_1$, $f_2$, and $f_3$ was used to assess the selected SMILES for the single-objective methods and how the independent evaluations of these objectives in the multi-objective EHVI method led to more diverse and balanced selections. This comparative analysis provides critical insights into the strengths and trade-offs between the different optimization strategies, facilitating a thorough evaluation of the GP-MOBO model.

%%%%%%%%%%%%%%%%%%%%%%%%%%%%%%%%%%%%%%%%%%%%%%%%%%%%%%%%%%%%%%%%%%%%%%%%%%%%%%%%
\chapter{Results and Analysis} \label{Chap6}
\section{Performance of our GP-MOBO Over Current GP BO Methods in Toy MPO Setup}
\begin{figure}[H]
    \centering
    \includegraphics[width=\linewidth]{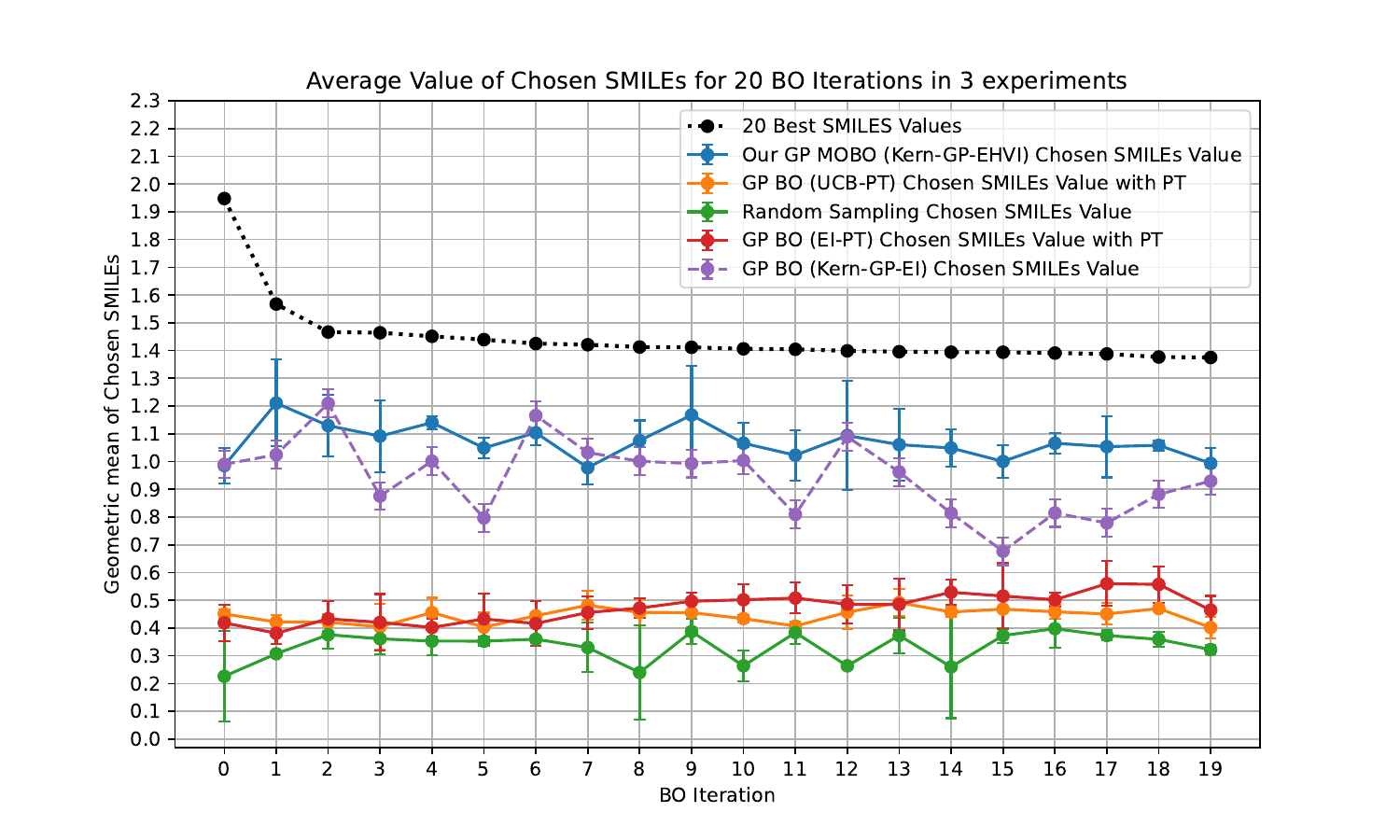}
    \caption[\textbf{Comparison of our GP-MOBO (KERN-GP-EHVI) and GP BO (UCB and EI with PyTorch(PT)) Models on Chosen SMILES Values across 20 BO Iterations on Toy-MPO DockSTRING dataset:}]{\textbf{Comparison of our GP-MOBO (KERN-GP-EHVI) and GP BO (UCB and EI with PyTorch(PT)) Models on Chosen SMILES Values across 20 BO Iterations on Toy-MPO DockSTRING dataset:}The error bars represent the standard deviation across three independent experimental runs for each model, indicating the variability in the performance of chosen SMILES values during the optimization process.}
\label{fig:summary-results}
\end{figure}

Our GP-MOBO model, implemented with the KERN-GP Package (detailed in \textcolor{blue}{Section \ref{subsection:custompackage}}), demonstrates a significant advantage over the GP BO models implemented by Tripp et al (2024)\cite{tripp_2024_diagnosing}. The KERN-GP package enables superior optimization performance, particularly when employing Expected Hypervolume Improvement (EHVI) acquisition function. This approach not only outperforms traditional PyTorch-based GP BO implementations but also consistently provides higher values of chosen SMILES across 20 BO iterations, showcasing robustness and efficacy of our methodology. 
\begin{figure}[H]
    \centering
    \includegraphics[width=\linewidth]{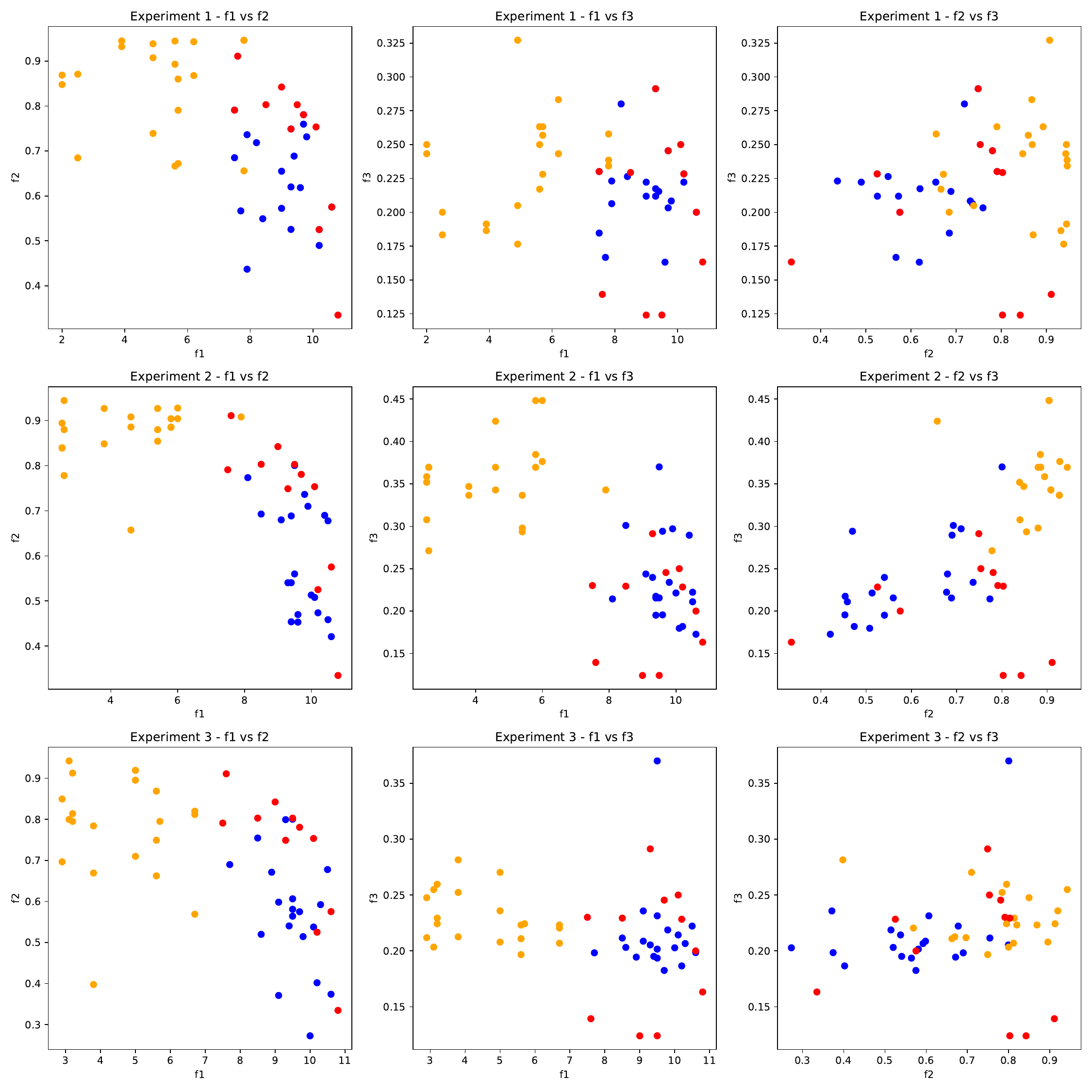}
    \caption[\textbf{Comparison of Pareto Front Clustering Between Our GP-MOBO (Kern-GP-EHVI) Model and GP BO (UCB-PT) Across Three Experiments: }]{\textbf{Comparison of Pareto Front Clustering Between Our GP-MOBO (Kern-GP-EHVI) Model and GP BO (UCB-PT) Across Three Experiments: } The performance of our GP-MOBO model (KERN-GP-EHVI) is contrasted with the GP BO (UCB-PT)(Tripp et al (2024)) approach in terms of how closely the selected points cluster around the Pareto optimal points. Each subplot represents a pairwise comparison of the objectives $f_1, f_2, f_3$. The blue points represent the results from our GP-MOBO model, the orange points represent the GP BO (UCB-PT) results, and the red points denote the Pareto optimal points.}
\end{figure}
Across all three experiments, it is evident that the points selected by our GP-MOBO model consistently cluster closer to the Pareto front compared to the GP BO (UCB-PT) method. This clustering indicates that our model is more effective at identifying solutions that achieve a balanced trade-off between the multiple objectives. The enhanced proximity to the Pareto front illustrates the superior exploration-exploitation balance achieved by our model, driven by maximizing the EHVI. This leads to a more nuanced and accurate optimization process, especially in complex multi-objective landscapes.

\begin{figure}[htbp]
    \centering
    \includegraphics[width=\linewidth]{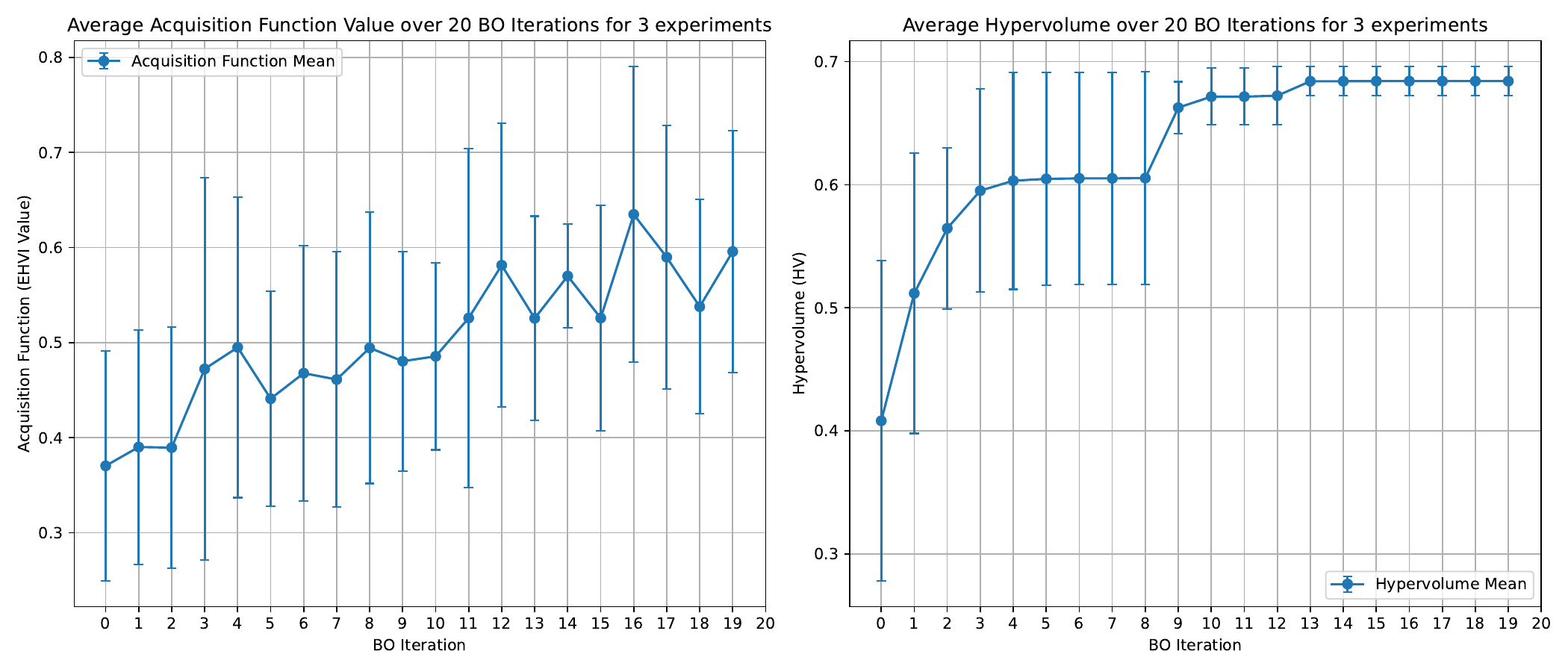}
    \caption[\textbf{Average EHVI Acquisition Function and Hypervolume Values for GP-MOBO across 20 BO iterations:}]{\textbf{Average EHVI Acquisition Function and Hypervolume Values for GP-MOBO across 20 BO iterations:} The left plot illustrates the average acquisition function value (EHVI) over 20 BO iterations, an indication that GP-MOBO's optimization process is working. The right plot shows average hypervolume values demonstrating that GP-MOBO is converging towards the Pareto front, as iterations increases. Error bars represent the standard deviation across three independent experimental runs for GP-MOBO with 10 random \texttt{known\_SMILES}, indicating the variability in the performance of chosen SMILES values during the optimization process.}
\label{acquisionfunchypervolumeplot}
\end{figure}
In 3 experiments, GP-MOBO's EHVI acquisition function, as illustrated in Figure \ref{acquisionfunchypervolumeplot}, demonstrates a clear convergence pattern over 20 BO iterations. EHVI increases steadily indicating the optimization process is effectively identifying better candidate SMILES. This behaviour aligns with the behaviour observed in Figure \ref{fig:summary-results},  where the geometric mean values of the chosen SMILES of GP-MOBO levels off close to the dataset best compared to the other methods, which are the 20 best SMILES evaluated in the dataset(see Appendix \ref{subsec:dataset_best_toy}). This suggests that GP-MOBO method, particularly with EHVI acquisition function, is refining its search near the dataset's best possible values early on in the process.

\section{Guacamol MPO Tasks}
\label{guacamolsection}
\subsection{Guacamol's Fexofenadine MPO Task (3 Objectives)}
\label{subsec:fexofenadine-results}
\begin{figure}[H]
    \centering
    \resizebox{0.75\linewidth}{!}{%
        \includegraphics{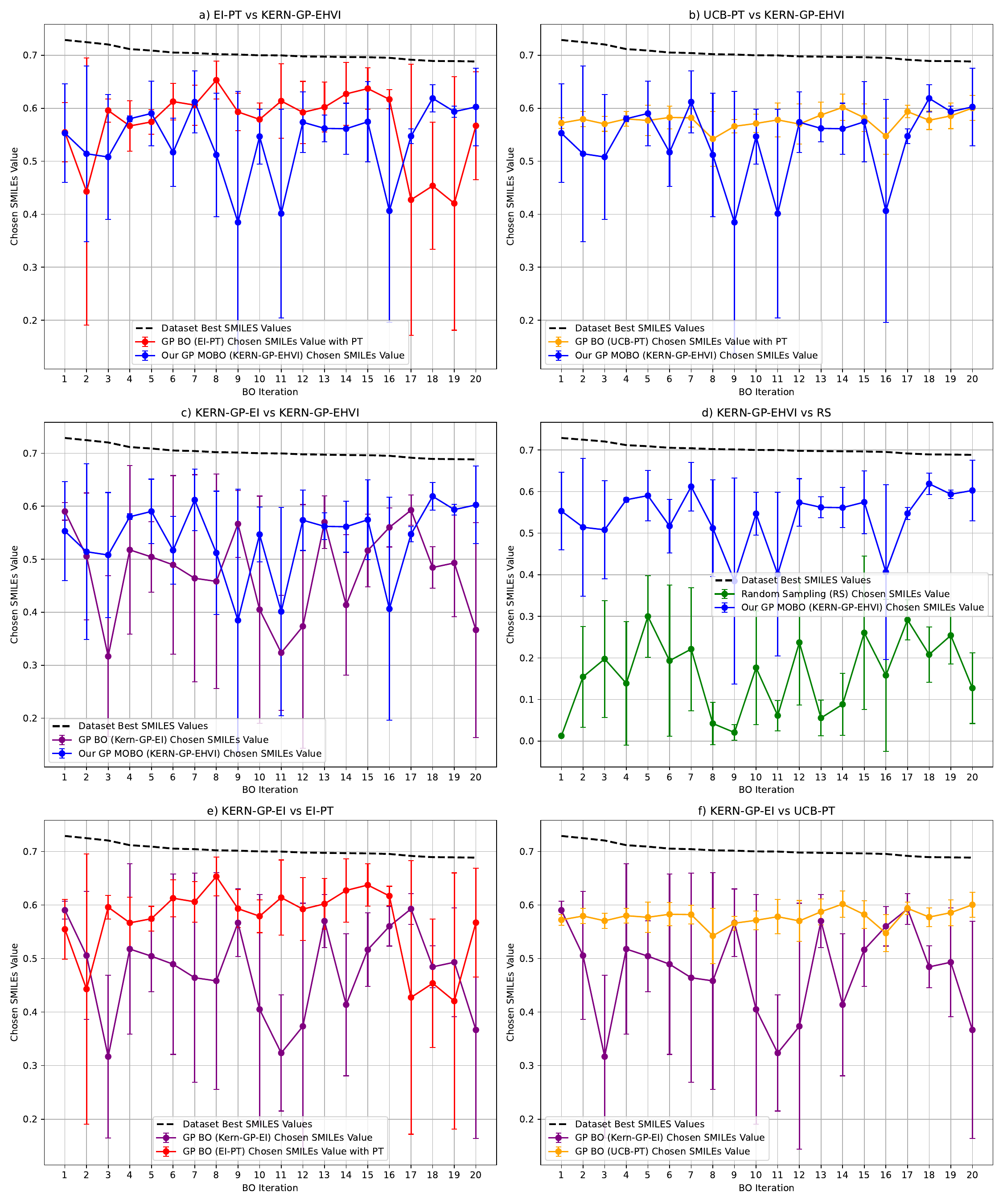}
    }
    \caption[\textbf{Fexofenadine MPO Task: Comparison of the average value of chosen SMILES across 20 Bayesian Optimization (BO) iterations for different methods: KERN-GP-EHVI, KERN-GP-EI, EI-PT, UCB-PT, and Random Sampling.}]{\textbf{Fexofenadine MPO Task: Comparison of the average value of chosen SMILES across 20 Bayesian Optimization (BO) iterations for different methods: KERN-GP-EHVI, KERN-GP-EI, EI-PT, UCB-PT, and Random Sampling.} The error bars represent the standard deviation across three independent experimental runs with 10 random initial \texttt{known\_SMILES} for each model, indicating the variability in the performance of chosen SMILES values during the optimization process.}
    \label{fig:comparison}
\end{figure}

\begin{figure}[H]
    \centering
    \includegraphics[width=\linewidth]{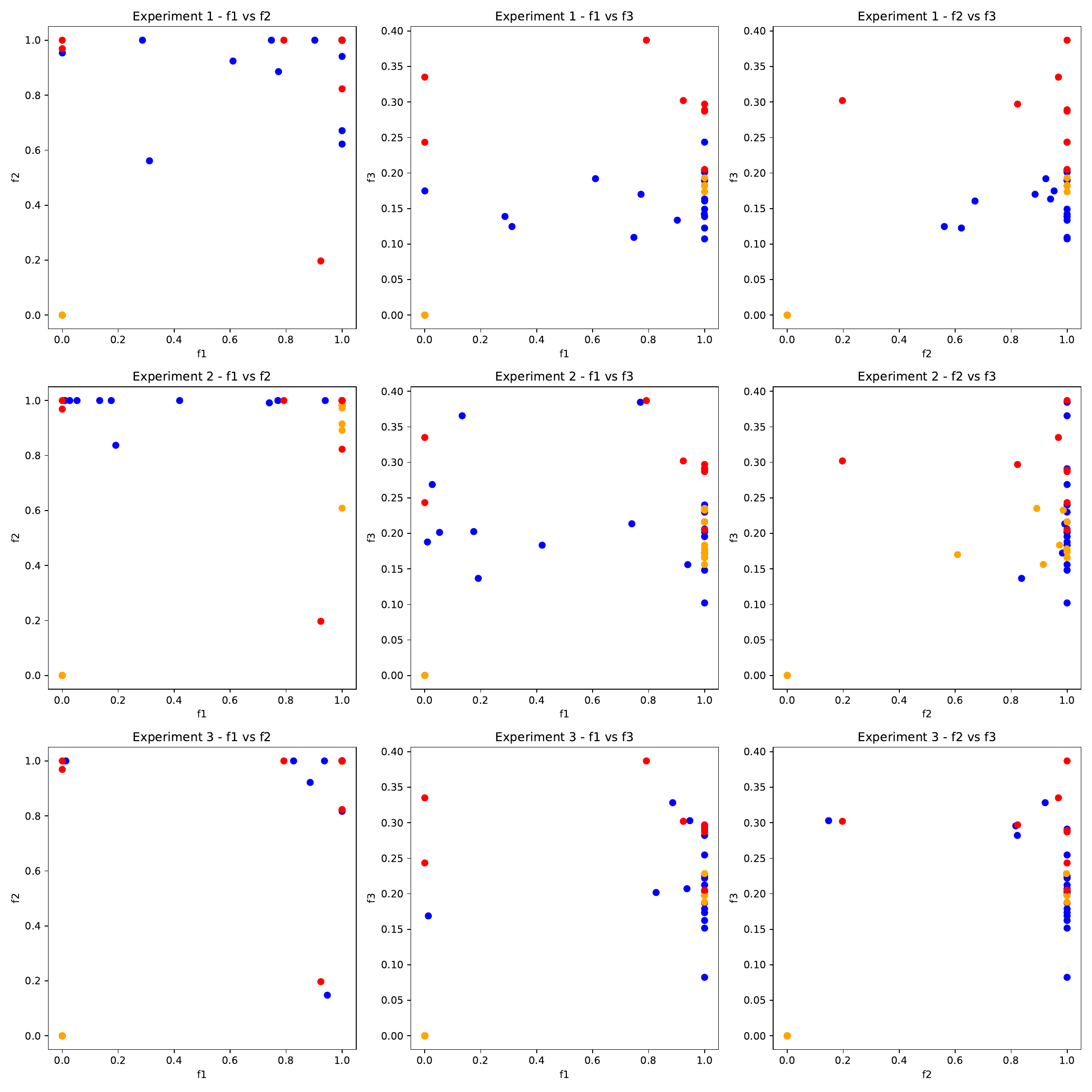}
    \caption[\textbf{Fexofenadine MPO Pareto Plots: Pareto plots for the Fexofenadine MPO optimization problem showing the objective values ($f_1, f_2, f_3$) for three different optimization experiments. The blue points represent the SMILES strings selected using the KERN-GP-EHVI approach, while the yellow points are those selected using the UCB-PT approach. The red points indicate the Pareto optimal solutions.}]{\textbf{Fexofenadine MPO Pareto Plots: Pareto plots for the Fexofenadine MPO optimization problem showing the objective values ($f_1, f_2, f_3$) for three different optimization experiments. The blue points represent the SMILES strings selected using the KERN-GP-EHVI approach, while the yellow points are those selected using the UCB-PT approach. The red points indicate the Pareto optimal solutions.} The distribution of blue points closer to the Pareto frontier across multiple plots indicates that KERN-GP-EHVI achieves a more diverse and effective exploration of the objective space, leading to better coverage of the Pareto front compared to UCB-PT.}
    \label{pareto-points-fexofenadine}
\end{figure}

\subsection{Guacamol's Amlodipine MPO Task (2 Objectives)}
\begin{figure}[H]
    \centering
    \resizebox{0.8\linewidth}{!}{%
        \includegraphics{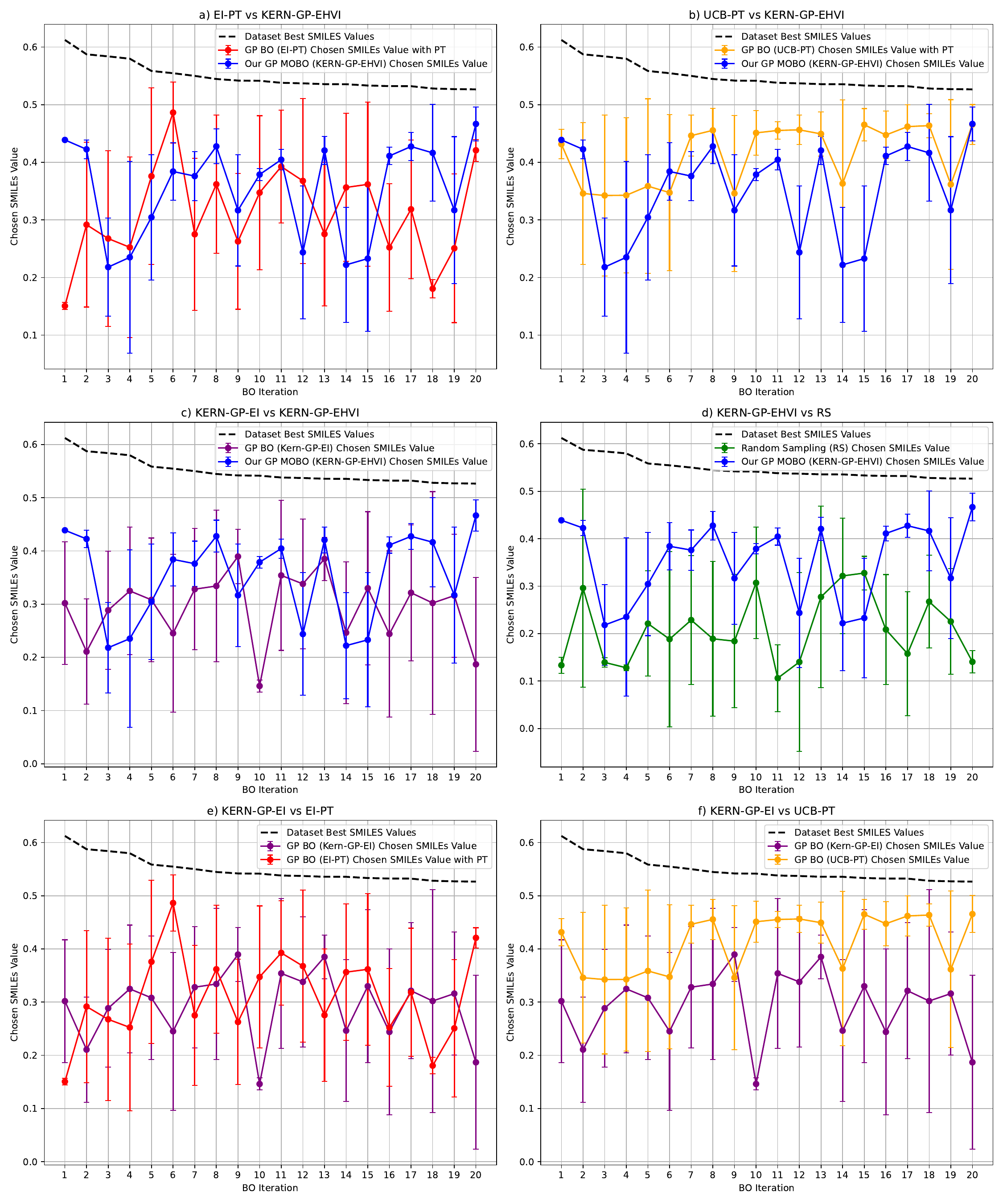}
    }
    \caption[\textbf{Amlodipine MPO Task: Comparison of the average value of chosen SMILES across 20 Bayesian Optimization (BO) iterations for different methods: KERN-GP-EHVI, KERN-GP-EI, EI-PT, UCB-PT, and Random Sampling.}]{\textbf{Amlodipine MPO Task: Comparison of the average value of chosen SMILES across 20 Bayesian Optimization (BO) iterations for different methods: KERN-GP-EHVI, KERN-GP-EI, EI-PT, UCB-PT, and Random Sampling.} The error bars represent the standard deviation across three independent experimental runs with 10 random initial \texttt{known\_SMILES} for each model, indicating the variability in the performance of chosen SMILES values during the optimization process.}
    \label{amlodipine-comparison}
\end{figure}

\begin{figure}[H]
    \centering
    \includegraphics[width=0.7\linewidth]{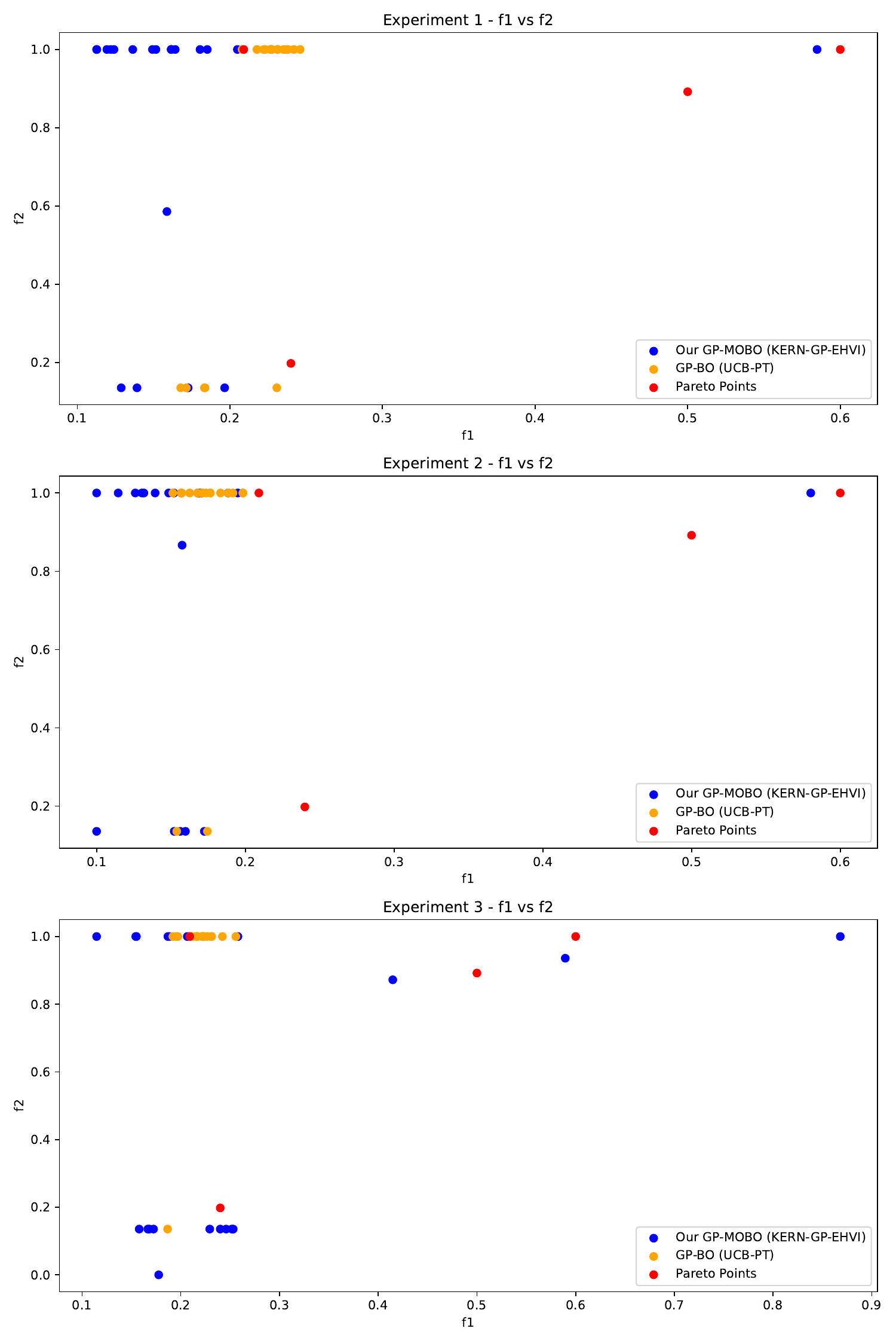}
    \caption[\textbf{Amlodipine MPO Pareto Plots: This figure presents pairwise plots of the objectives $f_1$ and $f_2$ across 3 experiments for  Amlodipine MPO:}]{ \textbf{Amlodipine MPO Pareto Plots: This figure presents pairwise plots of the objectives $f_1$ and $f_2$ across 3 experiments for  Amlodipine MPO:} The blue points correspond to the SMILES strings selected using the KERN-GP-EHVI approach, while the yellow points are selected using the UCB-PT approach. The red points represent the Pareto-optimal solutions, which are non-dominated with respect to the other points.}
    \label{pareto-points-amlodipine}
\end{figure}

\subsection{Guacamol's Perindopril MPO Task (2 Objectives)}
\begin{figure}[H]
    \centering
    \resizebox{0.8\linewidth}{!}{%
        \includegraphics{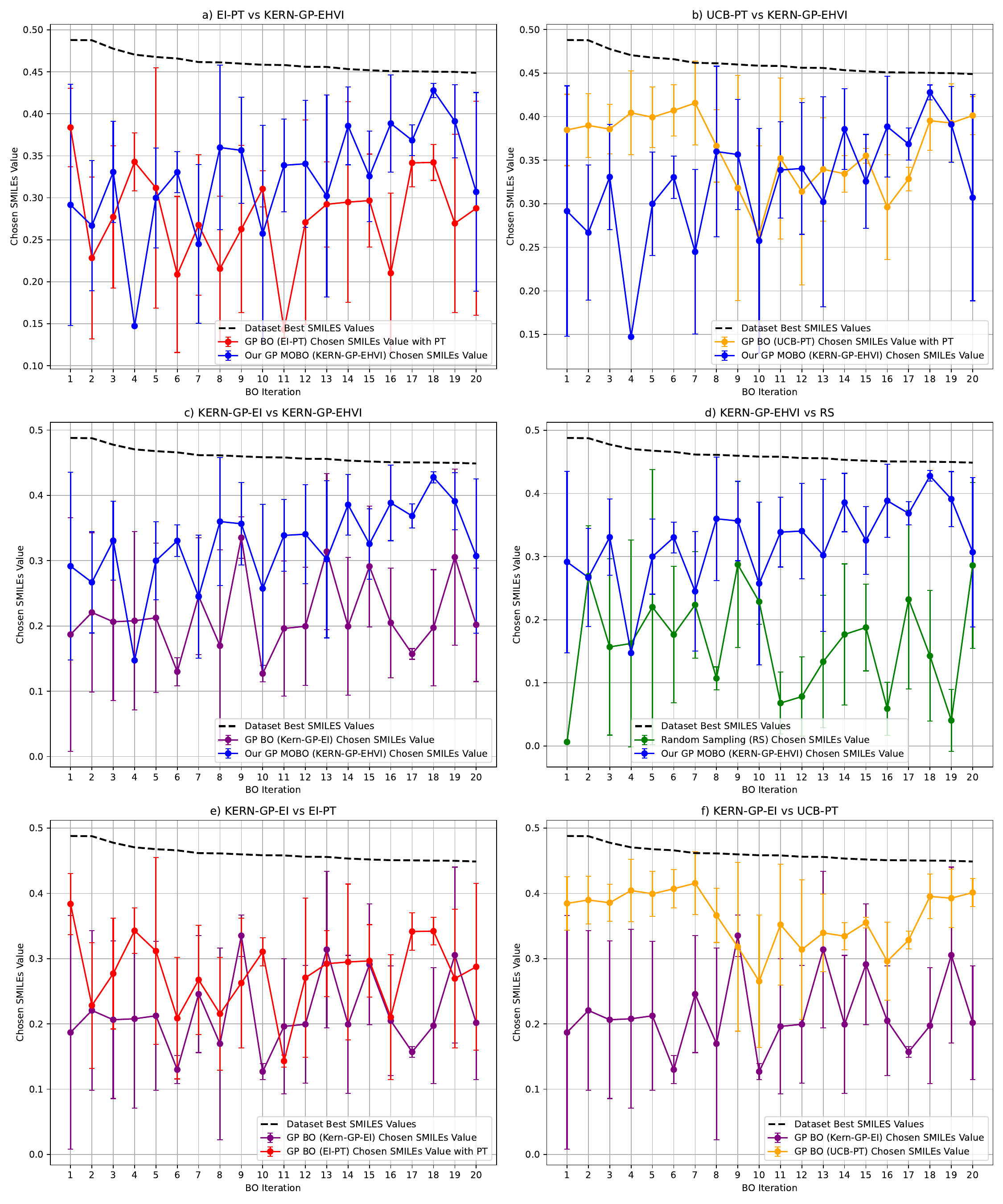}
    }
    \caption[\textbf{Perindopril MPO Task: Comparison of the average value of chosen SMILES across 20 Bayesian Optimization (BO) iterations for different methods: KERN-GP-EHVI, KERN-GP-EI, EI-PT, UCB-PT, and Random Sampling.}]{ \textbf{Perindopril MPO Task: Comparison of the average value of chosen SMILES across 20 Bayesian Optimization (BO) iterations for different methods: KERN-GP-EHVI, KERN-GP-EI, EI-PT, UCB-PT, and Random Sampling.}The error bars represent the standard deviation across three independent experimental runs with 10 random initial \texttt{known\_SMILES} for each model, indicating the variability in the performance of chosen SMILES values during the optimization process}
    \label{perindopril-comparison}
\end{figure}

\begin{figure}[H]
    \centering
    \includegraphics[width=0.7\linewidth]{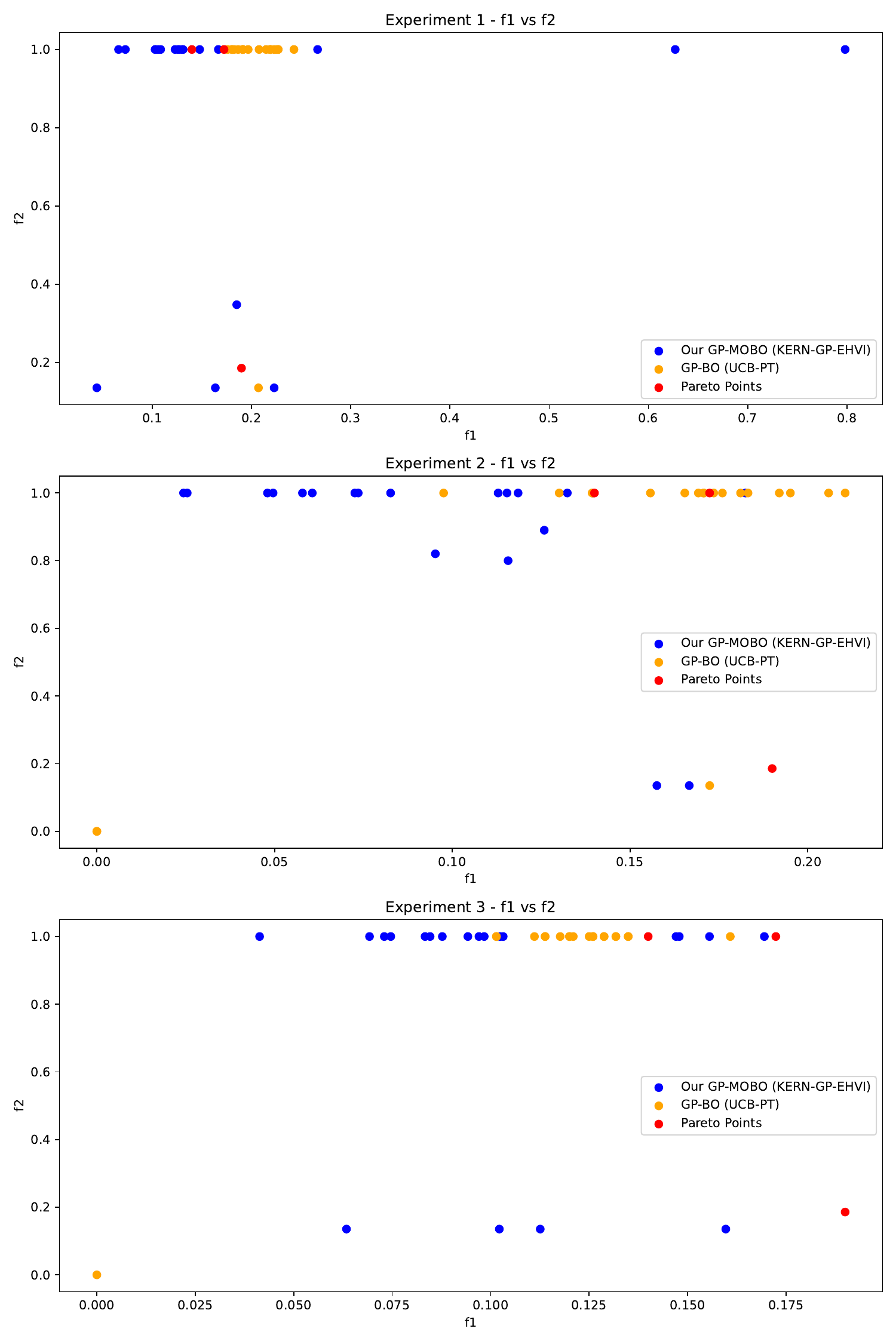}
    \caption[\textbf{Perindopril MPO Pareto Plots: This figure presents pairwise plots of the objectives $f_1$ and $f_2$ across 3 experiments for  Perindopril MPO.}]{\textbf{Perindopril MPO Pareto Plots: This figure presents pairwise plots of the objectives $f_1$ and $f_2$ across 3 experiments for  Perindopril MPO.} The blue points represent the SMILES strings selected using the KERN-GP-EHVI approach, while the yellow points are selected using the UCB-PT approach. The red points indicate the Pareto-optimal solutions.}
    \label{pareto-points-perindopril}
\end{figure}

\begin{figure}[H]
    \centering
    \includegraphics[width=1.1\linewidth]{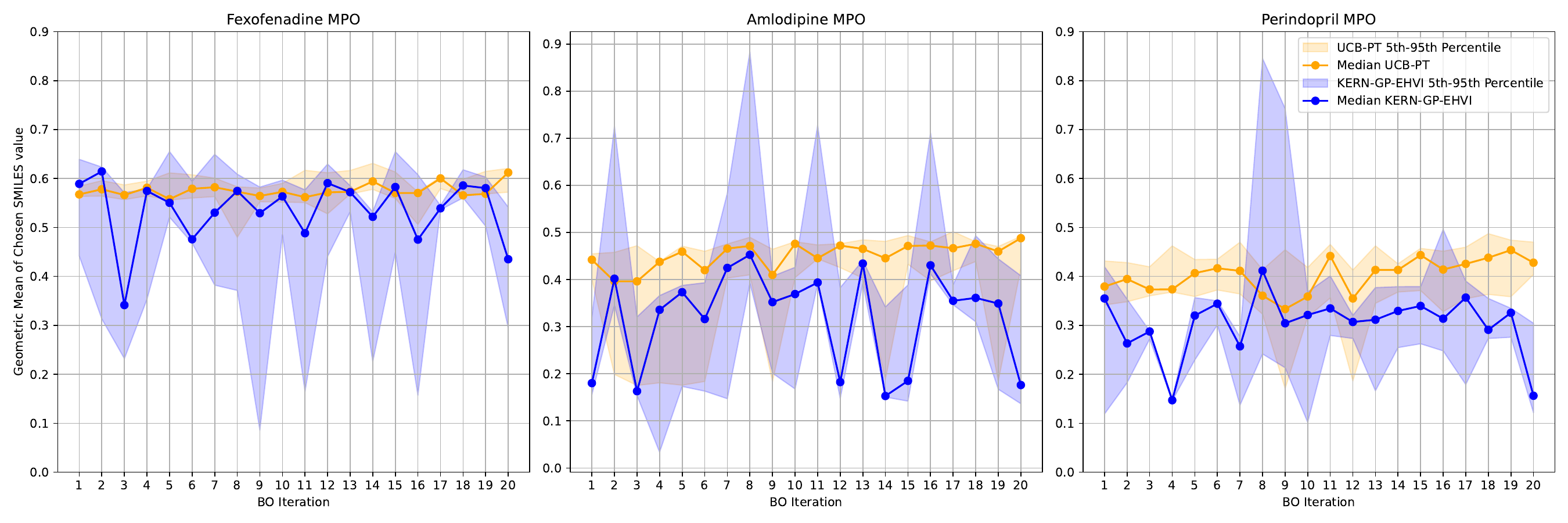}
    \caption[\textbf{Comparison of Geometric Mean of Chosen SMILES Values Across Three MPO Tasks for UCB-PT and KERN-GP-EHVI Methods Over 20 BO Iterations: The performance comparison between the UCB-PT (orange) and KERN-GP-EHVI (blue) methods across three multi-objective optimization (MPO) tasks: Fexofenadine, Amlodipine, and Perindopril.}]{\textbf{Comparison of Geometric Mean of Chosen SMILES Values Across Three MPO Tasks for UCB-PT and KERN-GP-EHVI Methods Over 20 BO Iterations: The performance comparison between the UCB-PT (orange) and KERN-GP-EHVI (blue) methods across three multi-objective optimization (MPO) tasks: Fexofenadine, Amlodipine, and Perindopril.} The geometric mean of the chosen SMILES values is plotted across 20 Bayesian Optimization (BO) iterations. The shaded regions represent the 5th to 95th percentile range, providing insight into the variability of each method, while the solid lines denote the median values.}
    \label{quantile-mpo}
\end{figure}

The KERN-GP-EHVI (GP-MOBO) method demonstrates broader exploration of the chemical space for all 3 MPO tasks, as evidenced by the wider shaded areas compared to UCB-PT (GP-BO). This indicates that KERN-GP-EHVI captures a more diverse range of SMILES, leading to better coverage of the Pareto front. However, the median values for KERN-GP-EHVI are generally lower than those for UCB-PT, suggesting that while KERN-GP-EHVI explores more, UCB-PT consistently identifies higher-value SMILES on average. This trade-off highlights KERN-GP-EHVI's strength in exploring the chemical space, which may be advantageous when discovering diverse molecules is crucial, while UCB-PT excels when maximizing SMILES value is prioritized.

To assess the performance of GP-MOBO in comparison to UCB-PT, we visualized the distribution of selected SMILES in Figures \ref{pareto-points-fexofenadine}, \ref{pareto-points-amlodipine} and \ref{pareto-points-perindopril}. A key observation is that GP-MOBO tends to select SMILES closer to the Pareto front across all plots, especially for Fexofenadine MPO, indicating higher diversity and exploration. GP-MOBO shows better coverage of the Pareto front for Fexofenadine, results in 2D cases indicate similar trends for both methods. While UCB-PT excels in identifying higher objective values, GP-MOBO provides superior diversity.

%%%%%%%%%%%%%%%%%%%%%%%%%%%%%%%%%%%%%%%%%%%%%%%%%%%%%%%%%%%%%%%%%%%%%%%%%%%%%%%%
\chapter{Discussion} \label{Chap7}
\section{Why our GP-MOBO over GP BO?}
The performance across the 2-dimensional problems, such as the final two MPO tasks, showed limited differences between GP-MOBO and GP-BO. However, in the case of Fexofenadine MPO, Figures \ref{pareto-points-fexofenadine} and \ref{quantile-mpo} indicate a clear advantage for GP-MOBO. While GP-MOBO did not consistently select better SMILES than GP-BO in every instance, it demonstrated a broader exploration of the chemical search space, identifying more diverse molecules. This characteristic allows GP-MOBO to uncover molecules with higher objective scores, which GP-BO often misses. The potential to explore a wider search space with higher-scoring molecules makes GP-MOBO a strong candidate for optimizing molecular properties where traditional models like GP-BO fall short.

In the following sections, we delve into the specific reasons why GP-MOBO outperforms GP-BO in certain setups, and why it should be considered as a preferred model in generative chemistry tasks.

\subsection{Why Our GP-MOBO Outperforms GP BO in Toy MPO, but Not in Real-World Drug Discovery GUACAMOL MPO Tasks?}
In the Toy MPO Setup in Figure \ref{fig:summary-results}, the performance of GP-MOBO consistently shows advantages over GP BO, even from the first Bayesian Optimization (BO) iteration. We investigated the first 10 initial \texttt{known\_SMILES} to determine whether there was a difference in the training dataset, indicating a less fair outcome. Prompting further, in the Toy-MPO setup, all the values for \texttt{known\_SMILES} for GP-MOBO and GP-BO were within a similar range for all repeat experiments. An example of \texttt{known\_SMILES} and their respective $f_1$, $f_2$, and $f_3$ (\texttt{known\_Y}) are provided for the GP-MOBO setup is provided as well as \texttt{known\_SMILES}  and their scalarized geometric mean for the single-objective cases (\texttt{known\_Y}) are provided in Table \ref{tab:initial_training_set} and \ref{tab:known_smiles_gmean} (Appendix \ref{exampletrainingdtaset}) respectively. Therefore, there is a different explanation as to why GP-MOBO picks better SMILES than GP-BO from BO iteration 1. 

From the GUACAMOL setup (Section \ref{guacamolsection}), we notice that this trend does not extend to real-life drug discovery tasks. To understand this difference, we need to explore the role of the fingerprint dimensionality and its interaction with the specific objectives of the Toy MPO setup.

\begin{figure}[H]
    \centering
    \includegraphics[width=0.9\linewidth]{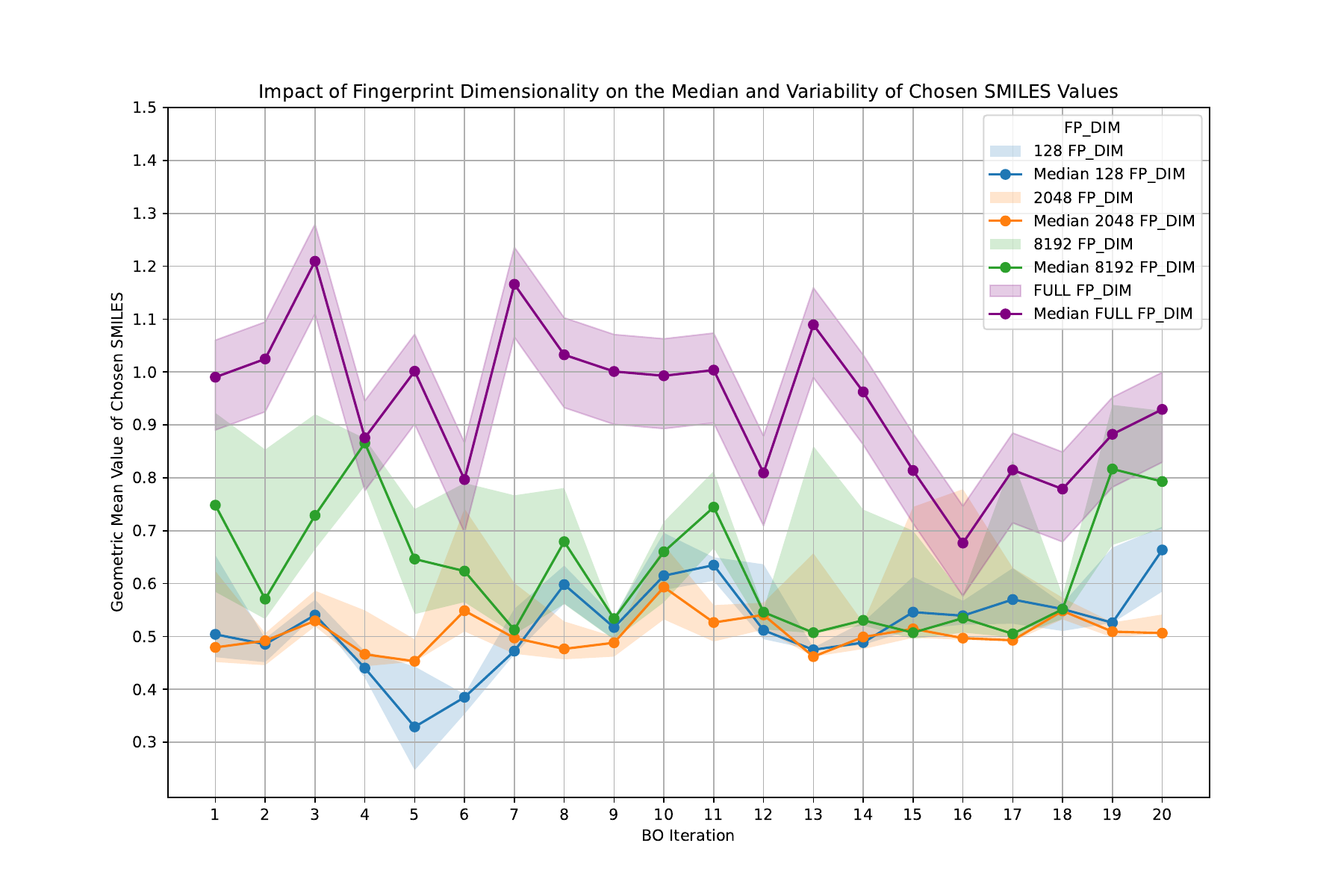}
    \caption[\textbf{Relationship of Fingerprint Dimensionality(\texttt{FP\_DIM}) with the Median and Variability of Chosen SMILES: }]{\textbf{Relationship of Fingerprint Dimensionality(\texttt{FP\_DIM}) with the Median and Variability of Chosen SMILES: }This compares the performance of different fingerprint dimensionalities (128, 2048, 8192, and FULL) by showing the median value and variability (25th to 75th percentile) of the geometric mean of chosen SMILES across 20 Bayesian Optimization (BO) iterations for Toy-MPO Objectives (Section \ref{toy-setup}).}
\label{fingerprint-dim-change}
\end{figure}

Figure \ref{fingerprint-dim-change} clearly demonstrates that, in the Toy MPO environment, higher fingerprint dimensionality \texttt{(FP\_DIM = FULL)} correlates strongly with improved geometric mean values for chosen SMILES. In contrast, lower dimensionalities (such as 2048 or 8192) do not capture the variability as effectively, leading to lower performance. However, we know from the GUACAMOL setup that \texttt{FP\_DIM = 2048} is sufficient for GP BO to sufficiently perform similarly to GP-MOBO. 

This result suggests that certain substructures, represented by the higher-dimensional fingerprint features, play a significant role in determining the objective values in the Toy MPO setup. 

\subsubsection*{Correlation Between Fingerprint Features and Performance of GP BO}
A key hypothesis emerging from our results is that the Toy MPO objectives benefit from specific substructures that are better captured by higher-dimensional fingerprints. These substructures, represented by the full fingerprint dimensionality directly correlate with the objectives of the Toy MPO. As a result, the models GP-MOBO and GP-BO (purple and blue lines in Figure \ref{fig:summary-results}), when using \texttt{KERN\_GP}, with full dimensionality, is able to explore and exploit these substructures more effectively, achieving better results compared to GP BO. 

Conversely, the standard dimensionality of 2048 (often used in real-world tasks for most models in molecular optimization) appears insufficient to fully capture the substructures, leading to the diminished performance in our toy MPO setup. The initial SMILES in Table \ref{tab:initial_training_set} and \ref{tab:known_smiles_gmean} (Appendix \ref{exampletrainingdtaset}) provide further evidence the only variable that has been changed in this setup is the fingerprint dimensionality and the single and multi-objective acquisition functions. This trend is consistent with the hypothesis that GP-MOBO benefits from the greater expressiveness provided by the full fingerprint representation.

In real-world tasks, where substructures are more complex and diverse, the full fingerprint dimensionality does not seem to confer to the same advantage. The variance in the complexity of drug-like molecules means that the relatively simplistic correlation between fingerprint dimensionality and objectives in the Toy MPO may not hold. In this regard, GP BO's approach, which operates effectively across multiple acquisition functions, shows competitive performance in the drug discovery task due to its robustness to different types of chemical representations and objective spaces. This would be left for further work to identify the correlation between the fingerprint dimensionality.

\subsection{Diversity on the Pareto Front}
The ability to explore diverse regions of the chemical latent space is critical in multi-objective optimization, particularly in drug discovery tasks where structural diversity can lead to novel compounds with improved or unique properties. In this context, GP-MOBO (KERN-GP-EHVI) demonstrates a clear advantage over GP BO (UCB-PT) by more effectively diversifying the chemical space it explores. We observe this habit in the Pareto plots in Figures \ref{pareto-points-fexofenadine}, \ref{pareto-points-perindopril}, \ref{pareto-points-amlodipine}, where yellow points (UCB-PT) are clustered around a similar region, not exploring the chemical space in contrast to the blue points (GP-MOBO) which explores the region and are fairly closer to the Pareto points (red).

\begin{figure}[H]
    \centering
    \includegraphics[width=1.1\linewidth]{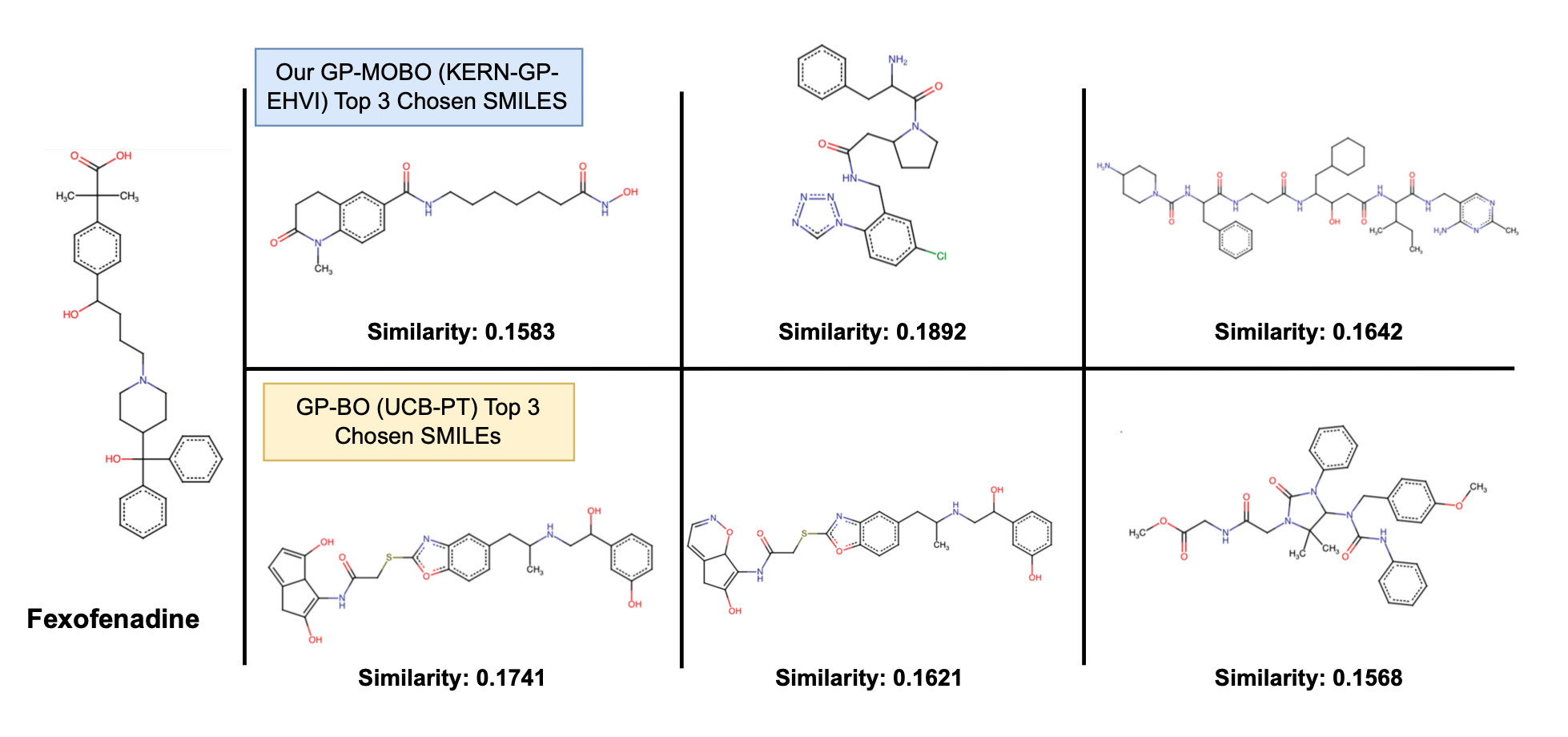}
    \caption[\textbf{Chemical Structure Comparison of Top 3 Chosen SMILES for Fexofenadine MPO by KERN-GP-EHVI and UCB-PT: }]{\textbf{Chemical Structure Comparison of Top 3 Chosen SMILES for Fexofenadine MPO by KERN-GP-EHVI and UCB-PT: }KERN-GP-EHVI (top row), selected SMILES strings with higher diversity and marginally better structural similarity (\texttt{TPSA\_score}) to Fexofenadine (similarity scores: 0.1583, 0.1892, 0.1642) compared to UCB-PT, whose chosen SMILES (bottom row) exhibit lower diversity and slightly lower similarity scores (0.1741, 0.1621, 0.1568).}
    \label{fig:similarity-chosen-smiles}
\end{figure}

As illustrated in Figure \ref{fig:similarity-chosen-smiles}, the SMILES selected by GP-MOBO show higher diversity compared to those chosen by GP-BO. The top row displays the three selected SMILES from GP-MOBO, with similarity scores of 0.1583, 0.1892, and 0.1642 relative to Fexofenadine. These lower similarity scores indicate that GP-MOBO explores a broader range of chemical structures, capturing a wider variety of substructures while maintaining marginally better structural alignment to the target compound (Fexofenadine).

In contrast, GP BO (UCB-PT), shown in the bottom row of Figure \ref{fig:similarity-chosen-smiles}, selects SMILES with higher similarity scores (0.1741, 0.1621, 0.1568). These SMILES are more closely clustered in the chemical space, indicating less diversity in the molecules GP-BO explores. This clustering behavior is consistent with the optimization strategy of UCB, which tends to focus on exploitation rather than exploration, resulting in a narrower chemical search space.

The greater diversity in the GP-MOBO results arises from its utilization of the Expected Hypervolume Improvement (EHVI) acquisition function, which balances both exploitation and exploration. This balance allows GP-MOBO to search unexplored regions of the chemical space more effectively, leading to the selection of structurally diverse molecules. The structural diversity of the selected SMILES is not only beneficial in terms of achieving better multi-objective performance but also increases the likelihood of discovering novel compounds with optimized properties for multiple objectives, such as Fexofenadine's MPO.

The observed behavior in this study is consistent with previous experiments using Amlodipine and Perindopril MPOs, where GP-MOBO exhibited greater diversity in the selected SMILES compared to GP-BO. This highlights GP-MOBO's superior capability in exploring broader chemical landscapes, a key factor in drug discovery applications where novelty and diversity are highly sought after.

\section{GP-MOBO's Prediction Evaluation}
After assessing the diversity of molecules generated by GP-MOBO (KERN-GP-EHVI), it is crucial to evaluate the model's predictive capabilities. While diversity on the Pareto front is key to exploring novel chemical structures, the effectiveness of these predictions directly impacts the success of optimization. Specifically, the Gaussian Process (GP) model plays a critical role in guiding the selection of SMILES through its predicted means and uncertainty quantification.
\begin{figure}[H]
    \centering
    \includegraphics[width=\linewidth]{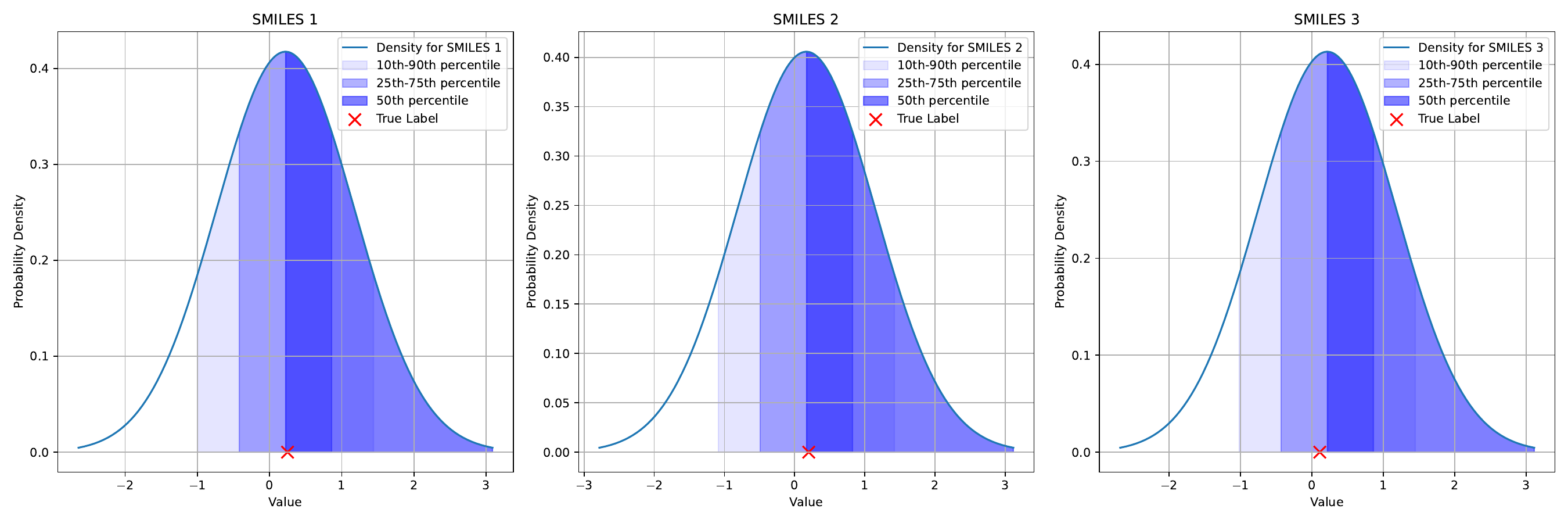}
    \caption[\textbf{Predictive Distributions of the Gaussian Process (GP) Model Across Fexofenadine MPO from Section \ref{subsec:fexofenadine-results} for Selected SMILES: }]{\textbf{Predictive Distributions of the Gaussian Process (GP) Model Across Fexofenadine MPO from Section \ref{subsec:fexofenadine-results} for Selected SMILES: } The shaded areas represent the $10^{th}$, $25^{th}$, and $50^{th}$ percentiles of the predicted distributions, while the red crosses mark the observed objective values (\texttt{KNOWN\_Y)} of Fexofenadine MPO.}
\label{gp-mobo-predictive-performance}
\end{figure}
Figure \ref{gp-mobo-predictive-performance} presents the predictive distributions for three selected SMILES from the Fexofenadine MPO task. These distributions help us understand how well the GP model predicts multi-objective outcomes and how closely the predicted means align with the observed values (\texttt{KNOWN\_Y)}). By analyzing the variance associated with each prediction, we can gauge the model's confidence, which is essential for making informed decisions in multi-objective optimization.

Table \ref{tab:gp_predictions} further reinforces this by comparing the actual objective values (\texttt{KNOWN\_Y)}) with the GP's predicted means and variances, as well as the Negative Log Predictive Density (NLPD)\cite{quionerocandela_2006_evaluating}, a performance metric that penalizes overconfident or under-confident predictions. Lower NLPD values indicate that the GP model is well-calibrated, suggesting that its predictions are reliable and the associated uncertainties are appropriately sized (see NLPD Definition in Appendix \ref{section:NLPD}).
\begin{table}[H]
\centering
\scriptsize  % Reduce font size to fit the table better
\caption{Table presents 3 SMILES from GUACAMOL's validation set (\texttt{guacamol\_v1\_valid.smiles}) corresponding to the molecules analyzed in Figure \ref{gp-mobo-predictive-performance}, along with their experimentally determined Fexofeandine MPO objective values (\texttt{KNOWN\_Y}) and the GP model's predicted means and variances, and performance metric for GP's prediction (NLPD).}
\label{tab:gp_predictions}
\begin{tabular}{|p{6cm}|c|c|c|c|}
\hline
\textbf{SMILES String} & \textbf{KNOWN\_Y} & \textbf{GP Mean} & \textbf{GP Variance} & \textbf{NLPD} \\ \hline
\texttt{CCCC(=O)NNC(=O)Nc1ccccc1} & 0.2489 & 0.2205 & 9.1320e-01 & 0.877 \\ \hline
\texttt{CC(=O)NC1CCC2(C)C(CCC3(C)C2\newline  C(=O)C=C2C4C(C)C(C)\newline CCC4(C)CCC23C)C1(C)C(=O)O} & 0.2008 & 0.1669 & 9.6686e-01 & 0.903 \\ \hline
\texttt{CC(=O)NC(C)Cc1ccc(C\#Cc2ccnc\newline(N3CCCC(F)C3)n2)cc1} & 0.1118 & 0.2159 & 9.3302e-01 & 0.890 \\ \hline
\end{tabular}
\end{table}
In this way, evaluating GP-MOBO's predictive accuracy is integral to understanding the model's overall performance in optimizing multi-objective tasks. The balance between exploration and exploitation not only depends on generating diverse SMILES but also on the GP's ability to accurately predict how those molecules will perform across objectives. This evaluation provides insight into the reliability of the GP predictions and opportunities for further model refinement.

\section{Monte Carlo Integration Error}
We concluded the GP-MOBO's prediction evaluation by demonstrating that the model provides good predictions with minimal uncertainty for chosen SMILES. A vital part of GP-MOBO's performance, especially when dealing with high-dimensional chemical spaces, is how accurately it estimates improvements across multiple objectives. This brings us to the importance of Monte Carlo (MC) integration methods in the Expected Hypervolume Improvement (EHVI) computation (discussed in Section \ref{subsection:EHVI}). 

The accuracy of EHVI is essential in the success of our GP-MOBO algorithm, especially when selecting molecules in high-dimensional chemical spaces. The MC integration method plays a pivotal role in this estimation, as it is used to approximate the EHVI, which directly influences the search for promising candidates. As Bayesian optimization relies on EHVI to guide the optimization process towards the Pareto front, the reliability of this estimate is crucial. 

\begin{figure}[H]
    \centering
    \includegraphics[width=0.8\linewidth]{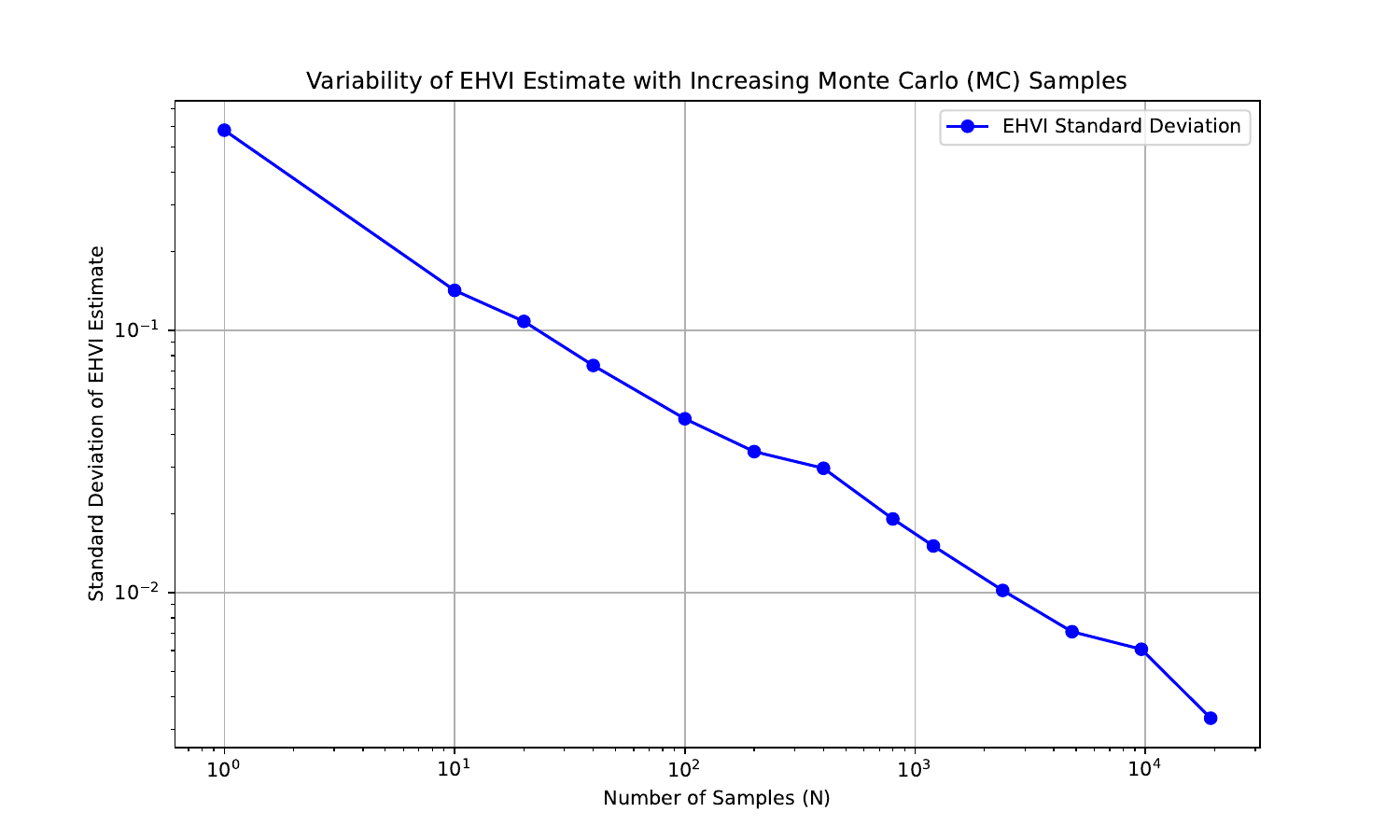}
    \caption[\textbf{Relationship between the number of Monte Carlo samples (N) and the variability of the Expected Hypervolume Improvement (EHVI) estimate in a Gaussian Process (GP) model:}]{\textbf{Relationship between the number of Monte Carlo samples (N) and the variability of the Expected Hypervolume Improvement (EHVI) estimate in a Gaussian Process (GP) model:} This convergence is important in the context of Bayesian Optimization, where accurate estimation of EHVI is critical for selecting the most promising candidates in high-dimensional spaces.}
    \label{fig:mc_estimation_error}
\end{figure}
As we increase the number of MC samples, the accuracy of EHVI improves, reducing noise and providing a clearer signal for selecting optimal candidates. However, this improvement comes at the cost of computational efficiency. To halve the error, we need to quadruple the number of MC samples, as demonstrated by the error scaling in Figure \ref{fig:mc_estimation_error}. Thus, a trade-off exists between computational effort and accuracy. In our experiments, we balanced this trade-off to ensure that the MC sample size provided a reasonable estimate of EHVI without excessively increasing computational time. The variability decreases as the number of MC samples increases, which is expected due to the central limit theorem as outlined by Lepage(1978)\cite{lepage_1978_central}. Specifically, the standard deviation of the EHVI estimate decreases proportionally to the inverse square root of the number of samples ($1/\sqrt{N}$). This convergence ensures that the optimization process remains reliable even when navigating high-dimensional chemical spaces.

By utilizing Monte Carlo integration effectively, we ensure that our GP-MOBO model reliably explores the Pareto front, especially in complex chemical spaces where exact analytical methods would be computationally prohibitive. This emphasizes the robustness of the GP-MOBO framework in handling real-world drug discovery tasks, making it an essential tool in the optimization of multi-objective problems. mWe leave for future work for testing different number of MC samples to further improve GP-MOBO's performance.

\section{Limitations and Further Work}
Bayesian Optimization (BO), especially when dealing with high-dimensional fingerprint vectors, poses significant computational challenges. The exact Gaussian Processes (GPs) used in this work scale poorly with the size of the dataset, exhibiting a computational complexity of $\mathcal{O}(N^3)$ and memory requirements as $\mathcal{O}(N^2)$, where $N$ is the number of data points. Such limitations make GPs less feasible for large-scale problems, especially in high-dimensional drug discovery applications\cite{cunningham_2023_actually}.

To mitigate these limitations, Sparse Gaussian Processes (SGPs)\cite{titsias_2009_variational} offer a viable alternative by approximating the full GP using a smaller set of inducing points or variables, as outlined by Michalis Titsias (2009)\cite{titsias_2009_variational}. SGPs can summarize the dataset efficiently while maintaining computational feasibility. Variants of SGPs, including Variational Fourier Features (VFF)\cite{hensman_2018_variational} and other inducing point methods, are commonly used to reduce the computational burden without sacrificing significant predictive performance. This makes SGPs particularly suited for high-dimensional input spaces such as molecular fingerprints, which are often sparse in nature. By leveraging the sparse nature of fingerprints, SGPs help reduce memory usage and accelerate computations, allowing for larger datasets to be processed.

Despite these advances, challenges still remain. For instance, in our GP-MOBO, the Improved Dimension-Sweep Algorithm (IDSA)\cite{fonseca_2006_an} was used to reduce the time complexity to  $\mathcal{O}(n^{d-2} \log n)$. However, this algorithm for hypervolume computation still faces scalability issues when the number of non-dominated points $n$ increases, or when dealing with higher dimensions. Furthermore, the recursive nature of the algorithm can lead to memory inefficiency in certain cases, particularly when processing large datasets. Additionally, the algorithm's performance is sensitive to the order in which objectives are processed, potentially leading to sub-optimal pruning, inflating computational costs.

Future work should explore the implementation of faster and more scalable EHVI algorithms. Additional methods like Multi-Objective Max-Value Entropy Search (MESMO)\cite{belakaria_2019_maxvalue} and Sequential Greedy Optimization \cite{ament_2024_unexpected} approaches have been shown to achieve comparable or superior performance while significantly reducing wall times in other optimization tasks. These methods provide promising avenues for further research, particularly for cases where higher-dimensional multi-objective problems arise, such as optimizing multiple properties of drug-like molecules. 

Additionally, future improvements include extending the current work to higher-dimensional MPO tasks such as GUACAMOL's Osimertinib MPO (4-dimensional), with more Bayesian optimization iterations. We could also optimize the performance by tuning the GP hyperparameters by minimizing the negative log marginal likelihood (NLML) (see Section \ref{section:NLML}). Further, investigating the integration of constraints into EHVI formulations to handle practical, real-world constraints that arise in drug discovery tasks is also important. We also leave the investigation of the relationship between fingerprint features and the DockSTRING objectives in the Toy MPO setup for future work to investigate the difference in performance as discussed above.

%%%%%%%%%%%%%%%%%%%%%%%%%%%%%%%%%%%%%%%%%%%%%%%%%%%%%%%%%%%%%%%%%%%%%%%%%%%%%%%%
\chapter{Conclusion} \label{Chap8}
% Clearly state the answer to the main research question
%Summarise and reflect on the research
%Make recommendations for future work on the topic
%Show what new knowledge you have contributed
In this work, we present GP-MOBO, which is developed and rigorously tested for multi-objective molecular optimization. This work compares with both scalarized single-objective methods (Expected Improvement and Upper Confidence Bound acquisition functions) and a multi-objective approach using the EHVI acquisition function. Our primary contributions include a detailed comparison of GP-MOBO with the current state-of-the-art model GP-BO model by Tripp \& Hernandez-Lobato(2024)\cite{tripp_2024_diagnosing}, Gao et al (2022)\cite{gao_2024_sample}, showing significant improvements in handling higher-dimensional fingerprint data and balancing conflicting objectives. 

Key findings from our experiments show that GP-MOBO consistently outperforms GP-BO in finding diverse and optimal SMILES across multiple tasks, particularly when maximizing hypervolume. The ability of GP-MOBO to better explore chemical space, as seen in the improved diversity on the Pareto front, demonstrates the superiority in identifying molecules that meet multiple objectives more effectively. The results are particularly pronounced in tasks such as Fexofenadine MPO, where GP-MOBO demonstrated enhanced exploration of structural diversity, resulting in better objective performance when compared to the current GP-BO model (UCB acquisition function and \texttt{FP\_DIM = 2048} were implemented) by Tripp \& Hernandez-Lobato(2024)\cite{tripp_2024_diagnosing}. 

However, there are limitations to our approach. While GP-MOBO excels in synthetic tasks, such as the Toy MPO setup, its advantages diminish in more complex real-world tasks like the GUACAMOL MPO setup. This is primarily due to the different role fingerprint dimensionality plays in these scenarios. In synthetic tasks, using full fingerprint dimensionality correlates strongly with improved performance, as the higher-dimensional representations capture specific substructures that are essential for optimizing the objectives. However, in real-world tasks, where the chemical space is more diverse and complex, the standard dimensionality of 2048 appears to be sufficient for GP-BO models, as it captures enough variability to achieve similar results. Our findings suggest that while higher-dimensional fingerprints offer advantages in synthetic tasks, they do not confer the same benefit in more realistic, diverse drug discovery tasks.

In future work, further investigation into the scalability of GP-MOBO for larger datasets, beyond 10,000 SMILES used in this work, would also be beneficial. Additionally, investigating why GP-MOBO outperforms GP-BO in the toy MPO setup needs to be explored. Finally, examining more diverse chemical spaces and different multi-objective acquisition functions could yield additional insights into optimizing molecular properties more robustly.
%%%%%%%%%%%%%%%%%%%%%%%%%%%%%%%%%%%%%%%%%%%%%%%%%%%%%%%%%%%%%%%%%%%%%%%%%%%%%%%%
\renewcommand{\bibname}{References}
\bibliographystyle{unsrt}
\bibliography{references.bib}
%%%%%%%%%%%%%%%%%%%%%%%%%%%%%%%%%%%%%%%%%%%%%%%%%%%%%%%%%%%%%%%%%%%%%%%%%%%%%%%%
% APPENDIX
\chapter{Appendix} 
\section{Source Code} 
Source code for all of the methods implemented in Chapter \ref{Chap4} and \ref{Chap5} for the project can be found in the GitHub repository: \url{https://github.com/anabelyong/GP-MOBO}. 

\section{Preliminary Mathematical Background}
\subsection{Cholesky Decomposition}
\label{section:cholesky}
The Cholesky decomposition, is useful in numerical methods including Gaussian Process (GP) regression. It is a specialization of the general LDU (lower-diagonal-upper) decomposition and is particularly applicable to symmetric, positive semi-definite matrices. 
\subsubsection*{Decomposing Symmetric Matrices}
Given a symmetric matrix A such that $A=A^T$, the Cholesky decomposition allows us to factor A as: 
\begin{equation*}
    A = LL^T
\end{equation*}
where L is a lower triangular matrix. This factorization is useful because it reduces the complexity of operations on A from $\mathcal{O}(n^3)$ to $\mathcal{O}(n^2)$ by leveraging the structure of triangular matrices. 
\subsubsection*{Connection to LDU Decomposition}
The LDU decomposition of a matrix A: 
\begin{equation*}
    A= LDU
\end{equation*}
where L is a lower triangular matrix, D is a diagonal matrix, and U is an upper triangular matrix. For symmetric matrices, it holds that $L = U^T$. Therefore, the LDU decomposition for a symmetric matrix can be rewritten as: 
\begin{equation*}
    A = LDL^T
\end{equation*}
This is where the Cholesky decomposition steps in. We can simplify D by further noting that D can be expressed as the square of a diagonal matrix $\sqrt{D}$, i.e:
\begin{equation*}
    D = \sqrt{D}\cdot \sqrt{D}^T
\end{equation*}
Substituting this into the LDU decomposition gives: 
\begin{equation*}
    A = L \sqrt{D} \cdot \sqrt{D}^T L^T = (L \sqrt{D})(L \sqrt{D})^T
\end{equation*}
Letting $L' = L \sqrt{D}$, we obtain the Cholesky decomposition: 
\begin{equation*}
    A = L'L'^T
\end{equation*}
L' is the Cholesky factor of matrix A.

\subsubsection*{Positive Semi-Definiteness and Cholesky Decomposition}
The Cholesky decomposition requires that the matrix \( A \) is positive semi-definite. This requirement ensures that all eigenvalues of \( A \) are non-negative, which in turn guarantees that the decomposition exists and is numerically stable.

The reason can be seen by considering the quadratic form.
For any nonzero vector $\boldsymbol{x}$,
$\boldsymbol{x}^\top A \boldsymbol{x}$ satisfies
\[
  \boldsymbol{x}^\top A \boldsymbol{x}
  = \boldsymbol{x}^\top L' {L'}^\top \boldsymbol{x}
  = \lVert {L'}^\top \boldsymbol{x} \rVert^2 \ge 0 .
\]
This confirms that $\boldsymbol{x}^\top A \boldsymbol{x} \ge 0$ if $A$ is positive semidefinite, so $A=L'{L'}^\top$.

\subsubsection*{Applications in Gaussian Processes}
In the context of Gaussian Processes, the Cholesky decomposition is particularly useful when sampling from a multivariate normal distribution. Given a multivariate Gaussian distribution \( \mathbb{X} \sim \mathcal{N}(\boldsymbol{\mu}, \mathbb{\Sigma}) \), where \( \mathbb{\Sigma} \) is the covariance matrix, we can express the distribution as:

\[
\mathbb{X} = \boldsymbol{\mu} + \mathbb{LZ}
\]

where \( \mathbb{L} \) is the Cholesky factor of \( \mathbb{\Sigma} \), and \( \mathbb{Z} \) is a vector of independent standard normal variables. This approach simplifies the sampling process and is essential for efficient GP regression implementations. The Cholesky decomposition not only simplifies matrix operations but also ensures numerical stability, making it a foundational tool in advanced statistical methods, including Gaussian Processes.

\subsection{Mercer's Theorem}
\label{mercertheorem}
\begin{definition}{\textbf{Mercer's Theorem: }}
Let $\mathcal{C}$ be a compact subset of $\mathbb{R}^n$. To ensure that a continuous symmetric kernel function $K(x_1, x_2)$ defined on $\mathcal{C}$ can be represented as an inner product in some feature space, the following expansion must hold:
\begin{equation*}
    K(x_1, x_2) = \sum_{k=1}^{\infty} \alpha_k \Phi_k(x_1) \Phi_k(x_2)
\end{equation*}
where $\alpha_k > 0$ are positive coefficients, and $\{\Phi_k(x)\}$ are the basis functions representing the implicit mapping from the input space $\mathcal{C}$ to the feature space. For the expansion to the valid, it is both necessary and sufficient that the kernel K is positive semi-definite, meaning that it satisfies the condition: 
\begin{equation*}
    \int_{\mathcal{C}} \int_{\mathcal{C}} g(x_1)g(x_2)K(x_1, x_2) \delta x_1 \delta x_2 \geq 0
\end{equation*}
for all square-integrable functions $g \in \mathcal{L}_2 (\mathcal{C})$
\end{definition}

\subsection{Positive Definite Kernel}
\label{section:PDK}
\begin{definition}{\textbf{Positive Definite Kernel:}}
A function $k:\mathcal{X} \times \mathcal{X} \rightarrow \mathbb{R}$ is called positive definite if, for any finite set of molecular fingerprints $x_1,...,x_n$ and any set of real numbers $\alpha_1,...,\alpha_n$, the following holds: 
\begin{equation*}
    \sum_{i=1}^n \sum_{j=1}^n \alpha_i \alpha_j k(x_i, x_j) \geq 0
\end{equation*}
This condition ensures that the associated Gram matrix (a matrix of kernel evaluations between pairs of data points) is positive semi-definite. 
\end{definition}

To establish that a function is positive semidefinite is central to applying the Mercer theorem. The integral in many cases, shown in Appendix, cannot be evaluated explicitly, making the proof of positive definiteness nontrivial. However, using the closer properties of positive definite functions defined below, we can show that this is a positive semidefinite symmetric kernel, ensuring that the kernel methods we are investigating can operate effectively within the RKHS framework. 

\begin{definition}{\textbf{Closure properties}}
\begin{itemize}
    \item Closure under a sum: For two positive semidefinite symmetric kernels $K_1, K_2: \mathcal{X} \times \mathcal{X} \rightarrow \mathbb{R}$, the sum becomes: 
    \begin{equation*}
        K = K_1 + K_2: \mathcal{X} + \mathcal{X} \rightarrow \mathbb{R}
    \end{equation*}
    is a positive semidefinite symmetric kernel. 
    \item Closure under a product: For two positive semidefinite symmetric kernels $K_1, K_2: \mathcal{X} \times \mathcal{X} \rightarrow \mathbb{R}$, the sum becomes: 
    \begin{equation*}
        K = K_1 \times K_2: \mathcal{X} \times \mathcal{X} \rightarrow \mathbb{R}
    \end{equation*}
    is a positive semidefinite symmetric kernel.
\end{itemize}
\end{definition}
Additionally, Aronszajn's theorem notes that any positive definite kernel corresponds to an inner product in some Hilbert space $\mathcal{H}$, with a mapping $\Phi: \mathcal{X} \rightarrow \mathcal{H}$ such that: 
\begin{equation*}
    k(x,x') = \langle \Phi(x), \Phi(x')\rangle_{\mathcal{H}}
\end{equation*}
This establishes the theoretical foundation for the kernel trick, as it guarantees that kernel methods can operate as if they were working in this high-dimensional space, even when space is not explicitly constructed. 

\subsection{Lebesgue Measure}
\label{Lebesgue-Measure}
The Lebesgue measure is a fundamental concept in measure theory, extending the intuitive notion of length, area, and volume to more complex sets beyond simple intervals. It was developed by Henri Lebesgue as a way to rigorously define the "size" of a set in a way that generalizes the concept of length to more abstract sets. Key properties of Lebesgue Measure are:
\begin{enumerate}
    \item \textbf{Extends Length:} For any interval $I = [a,b]$ in the real line $\mathbb{R}$, the Lebesgue measure $\mu(I)$ coincides with the length of the interval, i.e. $\mu(I) = b-a$.
    \item \textbf{Monotonicity: } If $A \subseteq B \subseteq \mathbb{R}$, then the Lebesgue measure is non-decreasing $\mu(A) \leq \mu(B)$. This ensures that larger sets have a greater or equal measure compared to their subsets. 
    \item \textbf{Translation Invariance: } For any set $A \subseteq \mathbb{R}$ and any real number $x_0$, the measure of A remains the same if the set is translated by $x_0$. Formally, $\mu(A+x_0)= \mu(A)$, where $A+x_0= \{x+x_0:x \in A\}$.
    \item \textbf{Countable Additivity: } is a countable collection of disjoint sets, then the measure of the union of these sets is the sum of their measures. Essentially, if $A_i \cap A_j = \emptyset$ for $i \neq j$, then: 
    \begin{equation*}
        \mu \left(\bigcup_{i=1}^{\infty} A_i\right) = \sum_{i=1}^{\infty} \mu(A_i)
    \end{equation*}
\end{enumerate}

\begin{definition}{\textbf{Lebesgue Outer Measure: }}
For any subset $E \subseteq \mathbb{R}$, the concept of Lebesgue outer measure $\mu^*(E)$. This is defined as: 
\begin{equation*}
    \mu^*(E) = \text{inf} \left\{\sum_{k=1}^{\infty}l(I_k): E \subseteq \bigcup_{k=1}^{\infty} I_k, \text{  where each $I_k$ is an interval}\right\}
\end{equation*}
Here, $\ell(I_k)$ denotes the length of the interval $I_k$, and the outer measure $\mu^*(E)$ is the infimum of sum of lengths of intervals covering the set E. 
\end{definition}

\begin{definition}{\textbf{Lebesgue Measurable Sets: }}
A set $E \subseteq \mathbb{R}$ is Lebesgue measurable if, for every set $A \subseteq \mathbb{R}$, the following holds:
\begin{equation*}
    \mu^*(A) = \mu^*(A \cap E) + \mu^*(A \cap E^C)
\end{equation*}
where $E^c$ denotes the complement of E. The Lebesgue measure $\mu(E)$ of a measurable set E is then defined as the outer measure $\mu^*(E)$.
\end{definition}
\textbf{HV Indicator Relevance: }These definitions and properties here are relevant for the Hypervolume Calculation, as it allows us to calculate this "volume" precisely, whether it is in 1 dimension (length), 2 dimensions(area) or higher dimensions (volume). The monotonicity property ensures that as the set of Pareto-optimal solutions expands, the hypervolume (measure of dominated region) increases or remains the same but never decreases. This property ensures that the hypervolume indicator correctly reflects improvements in the Pareto front. 

Countable additivity ensures that the total hypervolume is simply the sum of the measures of these individual regions. This property is fundamental when calculating the hypervolume, as it guarantees that the measure of the entire dominated region can be computed by adding up the measures of smaller, disjoint parts Conclusively, the hypervolume indicator can be thought of as a measure of the "outer" region dominated by the Pareto front, bounded by the reference point. The Lebesgue outer measure helps in defining this measure rigorously, ensuring that the hypervolume is calculated as the smallest possible "volume" that covers the entire dominated region.

\subsection{Klee's Measure Problem}
\label{section:KMP}
Klee's Measure Problem, a huge problem in computational geometry, involves determining the measure (such as length, area or volume) of the union of a collection of axis-aligned rectangles (or more generally, hyperrectangles) in d-dimensional space. The problem is stated as follows: 
\begin{definition}{\textbf{Klee's Measure Problem: }}
Let $R= \{R_1, R_2,...,R_n\}$ be a set of n axis-aligned hyperrectangles in $\mathbb{R}^d$. Each hyperrectangle $R_i$ is defined by its lower and upper bounds in each dimension. The objective of Klee's Measure Problem is to compute the volume of the union of these hyperrectangles, denoted by $V(R)$, where the volume is defined as the measure of the region covered by at least one hyperrectangle in R. This measure is expressed as: 
\begin{equation*}
    V(R) = \mu_d \left(\bigcup_{i=1}^n R_i\right)
\end{equation*}
where $\mu_d$ denotes the Lebesgue measure in d-dimensions.
\label{def:kleemeasure}
\end{definition}
The challenge in solving KMP arises from the potential overlap among hyperrectangles, as simply summing the volumes of the individual hyperrectangles would overestimate the total volume due to overlapping regions. The problem is known to have a complexity of $\mathcal{O}(n \log n + n^{d/2} \log n)$ in general d-dimensional space, making it computationally challenging for higher dimensions. 

\subsection{Gaussian Random Fields (GRFs)}
\label{def:GRFs}
A Gaussian random field (GRF is a collection of random variables indexed by a set of points in space, typically denoted as $Y(x)$, where x belongs to some spatial domain $\mathbb{R}^d$. Any finite collection of these random variables follows a multivariate Gaussian distribution. This property makes GRFs a powerful tool in modeling spatial phenomena where the underlying stochastic process is assumed to be Gaussian.
\begin{definition}{\textbf{GRFs: }}
Let $\{Y(x) : x \in \mathbb{R}^d\}$ be a random field and for any finite set of points $\{x_1, x_2,...,x_n\} \subset \mathbb{R}^d$, the joint distribution of $(Y(x_1), Y(x_2),...,Y(x_n))$ is a multivariate Gaussian. This can be expressed as: 
\begin{equation*}
    Y=(Y(x_1), Y(x_2),...,Y(x_n)) \sim \mathcal{N}(\mu, C)
\end{equation*}
where $\mu = (\mu(x_1), \mu(x_2),....,\mu(x_n))$ is the mean vector and C is the covariance matrix with entries $C_{ij} = Cov(Y(x_i), Y(x_j))$
\end{definition}

\subsubsection*{Covariance Structure and Homogeneity}
An important component of GRFs, is their covariance structure, which dictates how the values of the field are correlated across space. The covariance function $C(x_i, x_j)= Cov(Y(x_i), Y(x_j))$ captures the spatial dependence between two points $x_i$ and $x_j$. In the case of homogeneous fields, the covariance function depends only on the relative distance between the points $C(x_i, x_j) = C(r)$ where $r = ||x_i-x_j||$.

The covariance function is central to the GRF's smoothness properties and its related to the power spectrum $P(k)$ through the Fourier transform: 
\begin{equation*}
    C(r) = \int_{\mathbb{R}^d} P(k) e^{ik \cdot r} \delta k
\end{equation*}
where P(k) is the power spectral density function, which describes the distribution of variance as a function of spatial frequency k. 

\section{Molecular Objectives Definitions}
\label{molecularobjectivedefinition}
\begin{itemize}
    \item $f(m)= -\text{DockingScore}(\text{PPARD}, m)$, represents the negative docking score for the Peroxisome Proliferator-Activated Receptor Delta (PPARD) score. A higher value of $f_1$ indicates a stronger binding affinity between molecule m and PPARD target. 
    \item $f(m) = -\text{QED}(m)$ is Quantitative Estimate for Drug-likeness (QED) score. This metric evaluates how "drug-like" a molecule is, based on factors such as molecular weight, lipophilicity (logP) and number of hydrogen bond donors/acceptors. Higher QED values suggest molecule possesses properties commonly associated with effective drugs. 
    \item $f(m) = \text{sim}(m, \text{celecoxib})$ measures similarity of molecule m to Celecoxib, a well-known drug, using a fingerprint-based similarity metric. This is to ensure the molecules retain high degree fo structural similarity to an existing successful drug, maintaining potential efficacy. 
    \item $f(m) = \text{Fexofenadine MPO}$ measures multiple objectives for Fexofenadine by scalarizing the objectives below with geometric mean:
    \begin{itemize}
        \item $f(m) = \text{sim}(m, \text{fexofenadine, AP})$: Measures similarity to Fexofenadine based on aromatic properties (AP).
        \item $f(m) = \text{TPSA}(m)$: Topological Polar Surface Area, representing molecule polarity and influencing permeability and bioavailability.
        \item $f(m) = \log P(m)$: Logarithmic partition coefficient, reflecting molecule hydrophobicity or lipophilicity.
    \end{itemize}
    \item $f(m) = \text{Amlodipine MPO}$ measures multiple objectives for Amlodipine by scalarizing the objectives below with geometric mean:
    \begin{itemize}
        \item $f(m) = \text{sim}(m, \text{amlodipine, ECFP4})$: Measures similarity to Amlodipine using ECFP4 fingerprints.
        \item $f(m) = \text{NumberRings}(m)$: Measures the number of rings in the molecule, applying Gaussian smoothing to match the optimal number for drug-like properties.
    \end{itemize}
    \item $f(m) = \text{Perindopril MPO}$ measures multiple objectives for Perindopril by scalarizing the objectives below with geometric mean:
    \begin{itemize}
        \item $f(m) = \text{sim}(m, \text{perindopril, ECFP4})$: Measures similarity to Perindopril using ECFP4 fingerprints.
        \item $f(m) = \text{NumberAromaticRings}(m)$: Measures the number of aromatic rings in the molecule, applying Gaussian smoothing to reflect drug-like structures.
    \end{itemize}
\end{itemize}

\section{GP-MOBO Implementation Details}
\subsection{Oracle Utility Function Example}
\label{oracle}
\begin{figure}[H]
    \centering
    \includegraphics[width=0.7\textwidth]{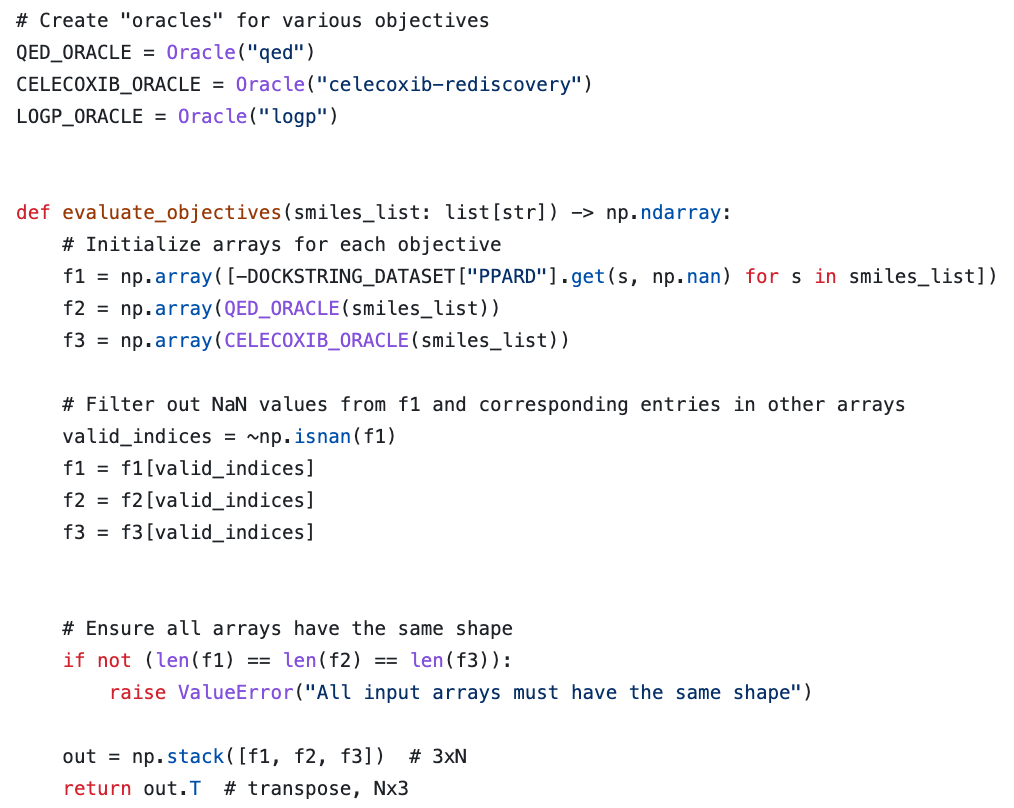}
    \caption{Oracle Function for Toy MPO Experimental Setup}
\end{figure}

\subsection{Hypervolume Computation Test Cases}
\label{ehvi-test-cases-section}
The test cases for Hypervolume Indicator and Expected Hypervolume Improvement were available in BoTorch's \url{https://github.com/pytorch/botorch/blob/main/test/utils/multi_objective/test_hypervolume.py}. The results from our implementation treating the data as a numpy array, instead of using tensorial data such as BoTorch is as shown: 
\begin{figure}[H]
    \centering
    \includegraphics[width=0.8\linewidth]{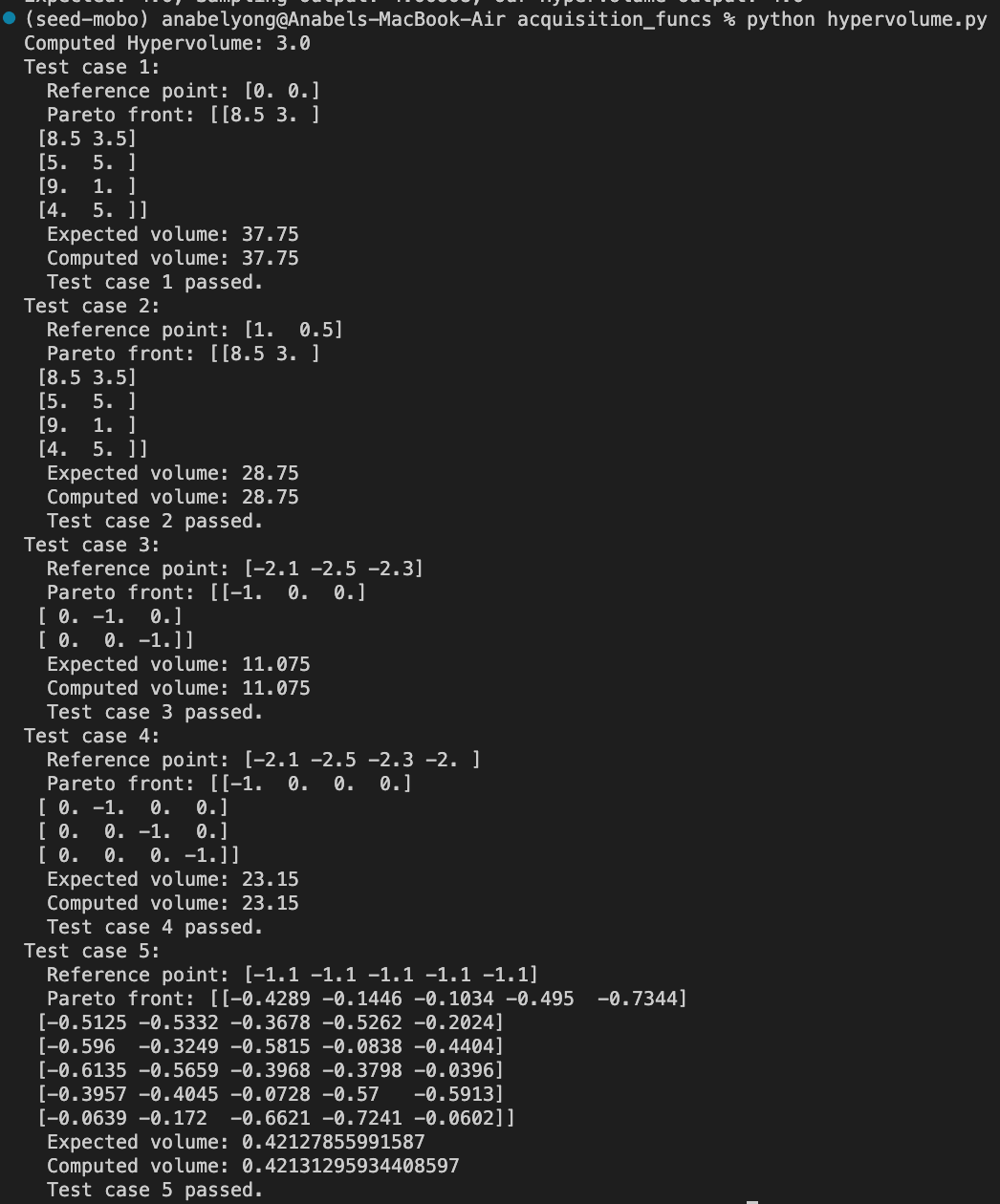}
    \caption{\textbf{Hypervolume Test Cases available from BoTorch passed by our EHVI implementation}}
    \label{fig:ehvi-test-cases}
\end{figure}

\subsection{Negative Log Predictive Density (NLPD)}
\label{section:NLPD}
The predictive density of a test observation $\tilde{y}$ given training data (x,y) and test data $\tilde{x}$ can be expressed as: 
\begin{equation*}
    p(\tilde{y}|\tilde{x}, x, y) = \int p(\tilde{y}, \tilde{x}, \theta) \cdot p(\theta|x,y)\delta\theta
\end{equation*}
where $\theta$ represents the GP's parameters. This integral is typically computed through Monte Carlo methods: 
\begin{equation*}
    p(\tilde{y}|\tilde{x}, x, y) \approx \frac{1}{M}\sum_{m=1}^M p(\tilde{y}|\tilde{x}, \theta^{m})
\end{equation*}
where $\theta^{m}$ being draws from the posterior distribution given the training data. Taking the logarithm of the predictive density provides the log predictive density, which is averaged over all test cases: 
\begin{equation*}
    \log p(\tilde{y}|\tilde{x}, x, y) \approx -\log M + \log \sum_{m=1}^M \exp(\log  p(\tilde{y}, \tilde{x}, \theta^{(m)}))) 
\end{equation*}
The NLPD is the negative of this average, giving us a measure of how well the model's predictive distribution captures the true outcomes.

\section{Additional Results}
\subsection{Example of Training Dataset for Both GP-MOBO and GP-BO}
\label{exampletrainingdtaset}
\begin{table}[ht]
\centering
\scriptsize 
\begin{tabular}{|p{7.5cm}|c|c|c|c|}
\hline
\textbf{Known SMILES} & \textbf{$f_1$} & \textbf{$f_2$} & \textbf{$f_3$} & \textbf{GMean} \\ \hline
O=C(NC1=C2C(NC=C2)=NC=C1)\newline C3CCC(CC3)C(N)C & 8.3 & 0.8107 & 0.1102 & 0.9051 \\ \hline
S(=O)(=O)(/N=C/1/C=C(C2C(=O)CC(CC2=O)(C)C)\newline C(=O)C=3C1=CC=CC3)C4=CC=C(C=C4)C & 9.5 & 0.6628 & 0.2823 & 1.2114 \\ \hline
O=C1N(C2CCCCCC2)CC(=O)N(C1C3=CC(OC)\newline =C(OC)C=C3)CC4=CC=CC=C4 & 9.0 & 0.6324 & 0.1643 & 0.9779 \\ \hline
O(C=1C=C(CNC=2N(C=3C(N2)=CC=CC3)\newline CCN4CCCCC4)C=CC1)C & 9.0 & 0.6840 & 0.1615 & 0.9981 \\ \hline
O1N=C(C=C1CC(C)C)C(=O)NCCC2=CC=CC=C2 & 8.8 & 0.8794 & 0.1327 & 1.0089 \\ \hline
ClC1=CC=C(N2C(=NN=C2SC(C(=O)\newline NC3=CC=4OCOC4C=C3)C)C(N(C)C)C)C=C1 & 8.7 & 0.5094 & 0.2015 & 0.9630 \\ \hline
O[C@]1([C@@]2([C@H]([C@H]3[C@H]([C@@H](O)\newline C2)[C@@]4(C(=CC3)CC(=O)C=C4)C)CC1)C)\newline C(=O)COC(=O)C & 7.8 & 0.5540 & 0.0563 & 0.6243 \\ \hline
O=N(=O)C1=CC(/C(=N/NC=2N=C(C(=NN2)\newline C=3C=CC=CC3)C4=CC=CC=C4)/C)=CC=C1 & 10.2 & 0.2738 & 0.1926 & 0.8133 \\ \hline
ClC1=CC=C(S(=O)(=O)C=2C(=CC(=NC2NC)\newline C)C=C1 & 8.2 & 0.9453 & 0.2755 & 1.2878 \\ \hline
ClC1=CC(NC=2N=C(N=C(N2)N)\newline CN3CCN(CC3)CC4=CC=5OCOC5C=C4)=CC=C1C & 10.2 & 0.5653 & 0.1517 & 0.9564 \\ \hline
\end{tabular}
\caption{Initial Training Set: Known SMILES and Corresponding Objective Values ($f_1$, $f_2$, $f_3$) for multi-objective GP-MOBO setup and their Geometric Mean of $f_1$, $f_2$, $f_3$.}
\label{tab:initial_training_set}
\end{table}

\begin{table}[ht]
\centering
\scriptsize 
\begin{tabular}{|p{7.5cm}|c|}
\hline
\textbf{Known SMILES} & \textbf{GMean} \\ \hline
C1=C(C2=C(C=C1O)\newline OC(C(C2=O)=O)C3=CC=C(C(=C3)O)O)O & 0.9159 \\ \hline
O=S(=O)(N1CCNCCC1)\newline C2=CC=CC=3C2=CC=NC3 & 1.2550 \\ \hline
C=1C=C2S/C(/N(CC)\newline C2=CC1OC)=CC(=O)C & 1.0040 \\ \hline
C=1(N=C(C=2C=NC=CC2)\newline C=CN1)NC=3C=C(NC(C4=CC=C(CN5CCN(CC5)\newline C)C=C4)=O)C=CC3C & 0.9521 \\ \hline
C1=CC=2C(=CNC2C=C1)\newline C=3C=CN=CC3 & 1.0327 \\ \hline
N1(C2=C(C(N)=NC=N2)\newline C=N1)C3=CC=CC=C3 & 0.8999 \\ \hline
C1(=C2C(C=CC=C2)=NC=N1)\newline NC3=CC(OC)=CC=C3 & 0.9412 \\ \hline
N1C(N(C(C2=CC=CC=C12)=O)\newline CCN3CCC(CC3)=C(C=4C=CC(=CC4)F)\newline C=5C=CC(=CC5)F)=S & 1.1289 \\ \hline
C1(O[C@@H](CC(C(=CC([C@H]([C@H](C([C@@H] \newline (C[C@@H](C=CC=CC=C([C@H](C[C@H]2O[C@]\newline (C(C(N3[C@H]1CCCC3)=O)=O)(O)[C@@H](CC2)C)OC) \newline C)C)=O)OC)O)C)C)=O)[C@@H](C[C@H]4C[C@@H](OC)\newline [C@H](O)CC4)C)=O & 0.9731 \\ \hline
O=C1C=2C=3C(=NNC3C=CC2)\newline C4=C1C=CC=C4 & 0.9411 \\ \hline
\end{tabular}
\caption{Initial Training Set: Known SMILES and Corresponding Geometric Mean Values provided for the single-objective GP BO setup}
\label{tab:known_smiles_gmean}
\end{table}
\subsection{Dataset BEST SMILES (Top 20 SMILES) in Toy MPO Setup}
\label{subsec:dataset_best_toy}
\begin{table}[H]
    \centering
    \scriptsize  % Reduces font size to help fit the table on the page
    \renewcommand{\arraystretch}{1.2}  % Adjusts the row height
    \begin{tabular}{|p{12cm}|p{3cm}|}  % Specifies column width
        \hline
        \textbf{Dataset Best SMILES for Toy MPO DockSTRING setup} & \textbf{Value of Best SMILES} \\ \hline
        S(C1=CC=C(C=C1)N2C(C3=CC=C(C)C=C3)=CC(C(F)(F)F)=N2)(N)(=O)=O & 1.9478373773604283 \\ \hline
        FC(F)(F)C1=CC(N2N=CC(=C2N)C=3C=CC(=CC3)C)=CC=C1 & 1.567558036817745 \\ \hline
        ClC1=CC=C(C=2C(=O)N(NC=3C=CC(=CC3)C)C(=O)C2)C=C1 & 1.466675340058482 \\ \hline
        N1(N=C(C=2C=CC(=CC2)C)C=C1N)C3=CC=C(C=C3)C & 1.4641374959896714 \\ \hline
        S(=O)(=O)(NNC1=NC=2C(N=C1C(F)(F)F)=CC=CC2)C3=CC=C(C=C3)C & 1.4512344549267704 \\ \hline
        S(=O)(=O)(N)C1=CC=C(NC(=O)NC=2C=CC(=CC2)C(F)(F)F)C=C1 & 1.439286157552866 \\ \hline
        FC(F)(F)C1=CC(N2CCN(CC2)C(=O)CC3=CC=C(C=C3)C)=CC=C1 & 1.4249646907867288 \\ \hline
        S(=O)(=O)(NNC=1C=CC(=CC1)C)C2=CC=C(C=C2)C & 1.421004073331502 \\ \hline
        C1(=CC=C(S(NC2=CC=C(C=C2)C(NC3=NOC(=C3)C)=O)(=O)=O)C=C1)C & 1.4127307356041288 \\ \hline
        CCNC(=O)C=1C=CC(=CC1)N2C(=CC(=N2)C)C3=CC=CC=C3 & 1.4116668811297421 \\ \hline
        S(=O)(=O)(N1N=C(N)C(=C1)C2=CC=C(F)C=C2)C3=CC=C(OCC)C=C3 & 1.4061813943772048 \\ \hline
        S(=O)(=O)(N)C1=CC=C(C=2C(=NOC2C)C=3C=CC=CC3)C=C1 & 1.4043521001657435 \\ \hline
        O=C(N1N=C(N=C1N)C=2C=CC(=CC2)C)CC3=CC=CC=C3 & 1.3989038673335519 \\ \hline
        S(=O)(=O)(N1CCN(CC1)C=2C(=CC=CC2)C(F)(F)F)C3=CC=C(C=C3)C & 1.3957869324271062 \\ \hline
        ClC1=CC(CN2CC(=NS(=O)(=O)C3=CC=C(C=C3)C)C=CC2=O)=CC=C1Cl & 1.3941200508505909 \\ \hline
        S(=O)(=O)(C=1C(=CC(=NC1NC)C)C)C2=CC=C(C=C2)C & 1.3939248840045841 \\ \hline
        S(=O)(=O)(NCC)C=1C=CC(NC(=O)C=2N(N=C(C2)C(F)(F)F)C)=CC1 & 1.3908884961286663 \\ \hline
        S(=O)(=O)(NCCN1C=2C(C=C1C)=CC=CC2)C3=CC=C(C=C3)C & 1.3878057287883785 \\ \hline
        FC(F)(F)C1=CN(CC=2C=CC(=CC2)C(=O)NC3=CC(=CC=C3)C)C(=O)C=C1 & 1.3767026956086061 \\ \hline
        S(=O)(=O)(NC1=C2CCCC2=NC(O)=C1)C3=CC=C(C=C3)C & 1.3749559507452085 \\ \hline
    \end{tabular}
    \caption{Dataset Best SMILES and their corresponding Values of Best SMILES in Toy MPO Setup}
    \label{tab:dataset_best_smiles}
\end{table}

\subsection{Dataset BEST SMILES (Top 20 SMILES) in Fexofenadine MPO}
\begin{table}[H]
    \centering
    \scriptsize  % Adjust font size to make the table fit better
    \renewcommand{\arraystretch}{1.2}  % Adjust row height for better readability
    \begin{tabular}{|p{13cm}|p{3cm}|}  % Adjusted column widths
        \hline
        \textbf{Best 20 SMILES for Fexofenadine MPO from GUACAMOL} & \textbf{Best SMILES Value} \\ \hline
        O=C(CCC(=O)NC(CO)C(O)c1ccc([N+](=O)[O-])cc1)NCCCNCCCCN(Cc1ccccc1)Cc1ccccc1 & 0.72892475 \\ \hline
        O=C(O)CC(O)(CSCCCCCCc1ccc2ccccc2c1)C(=O)O & 0.72478167 \\ \hline
        CNC(=O)N1CCC(NC(=O)c2ccc(Oc3ccc(C\#CC4(O)CN5CCC4CC5)cc3)cc2)CC1 & 0.72041918 \\ \hline
        CN1C(=O)C(C(O)C2CCCCC2)NC(=O)C12CCN(Cc1ccc(Oc3ccc(C(=O)O)cc3)cc1)CC2 & 0.71171583 \\ \hline
        O=C(Cc1cc2ccccc2[nH]1)N1CCC(Nc2ncc(C(O)=NO)cn2)(c2ccccc2)CC1 & 0.70899797 \\ \hline
        COc1ccc2nccc(C(O)CN3CCC(NCc4cc5cccnc5[nH]4)CC3)c2n1 & 0.70524978 \\ \hline
        CC(C)(C)NC(=O)C1CN(Cc2cccnc2)CCN1C[S+](O-)CC(Cc1ccccc1)C(=O)NC1c2ccccc2CC1O & 0.70424961 \\ \hline
        O=C(CN(c1cccc([N+](=O)[O-])c1)S(=O)(=O)c1ccccc1)N1CCCCC1 & 0.70213893 \\ \hline
        O=C(c1ccc(O)cc1OCC(O)CN1CCC2(CC1)Cc1cc(Cl)ccc1O2)N1CCOCC1 & 0.70144999 \\ \hline
        O=C(O)c1ccc(-c2noc(C3CCN(C(=O)NC4CC4c4ccccc4)CC3)n2)cc1 & 0.70001582 \\ \hline
        CCC(=O)N(c1ccccc1)C1(C(=O)OC)CCN(CCn2c(=O)c3ccccc3n(CC)c2=O)CC1 & 0.69978857 \\ \hline
        CC(O)C1C(=O)N2C(C(=O)[O-])=C(c3ccc(C[n+]4ccc(N5CCCCC5)cc4)cc3)CC12 & 0.69792347 \\ \hline
        O=C(c1ccc(NS(=O)(=O)c2cccc3c2OCCO3)cc1)N1CCC(O)(Cc2ccccc2)CC1 & 0.69737852 \\ \hline
        O=C(O)CC1c2ccccc2C(=O)N(CC(=O)NCCCCNc2nc3ccccc3[nH]2)c2ccccc21 & 0.69677083 \\ \hline
        O=C(O)CNC(C(=O)N1CCCC1C(=O)NCC\#Cc1c[nH]cn1)C(c1ccccc1)c1ccccc1 & 0.69641193 \\ \hline
        NCc1ccc(Cl)cc1CNC(=O)C1CCCN1C(=O)C1(O)c2ccccc2-c2c1ccc[n+]2[O-] & 0.69526927 \\ \hline
        Cn1c(=O)c(C(=O)NCC2CCN(Cc3ccccc3)CC2)c(O)c2cc(O)c(O)cc21 & 0.69161915 \\ \hline
        O=C(C=Cc1ccc([N+](=O)[O-])cc1)Nc1ccc(N2CCN(CC(O)(Cn3cncn3)c3ccc(F)cc3F)CC2)c(F)c1 & 0.68930971 \\ \hline
        CCOCCN(CC(O)CN1CCCC2(CC(=O)c3cc(O)ccc3O2)C1)S(=O)(=O)c1c(C)cccc1C & 0.68897279 \\ \hline
        O=C(O)CN1CCC(c2c(C=Cc3ccc4ccccc4n3)nc3c(N4CCOCC4)ccnn23)CC1 & 0.68839601 \\ \hline
    \end{tabular}
    \caption{Best 20 SMILES and their corresponding Values for Fexofenadine MPO}
    \label{tab:best_20_smiles-fexofenadine}
\end{table}

\subsection{Dataset BEST SMILES (Top 20 SMILES) in Amlodipine MPO}
\begin{table}[H]
    \centering
    \scriptsize  % Adjust font size to make the table fit better
    \renewcommand{\arraystretch}{1.2}  % Adjust row height for better readability
    \begin{tabular}{|p{13cm}|p{3cm}|}  % Adjusted column widths
        \hline
        \textbf{Best 20 SMILES for Amlodipine MPO from GUACAMOL} & \textbf{Best SMILES Value} \\ \hline
        COC(=O)C1=C(C)NC(C)=C(C(=O)OCc2cccc(F)c2)C1c1cccc([N+](=O)[O-])c1 & 0.61237244 \\ \hline
        CCOC(=O)C1=C(C)N=C(C)C(=C(O)OCC)C1c1nc2ccccc2[nH]1 & 0.58747999 \\ \hline
        CCOC(=O)C1=C(C)NC(C)=C(C(=O)OCC)C1c1cc(C(C)CC)c2oc(=O)c(C(=O)OC)cc2c1 & 0.58387421 \\ \hline
        COc1cc(C2NC(=O)NC(C)=C2C(C)=O)ccc1OCc1ccccc1Cl & 0.57983351 \\ \hline
        CCOC(=O)C1=C(C)NC(=S)NC1c1ccc(NC(=O)Nc2ccc(OC)cc2)cc1 & 0.5585696 \\ \hline
        CCOC(=O)c1c(C)[nH]c(C)c1C(=O)CSc1nnnn1-c1ccccc1 & 0.5547002 \\ \hline
        COC(=O)c1c(SCC(=O)Nc2cccc(Cl)c2C)[nH]c2ccccc2c1=O & 0.5500191 \\ \hline
        CCOC(=O)C1=C(C)OC(N)=C(C(=O)OCC)C12C(=O)Nc1ccc(Br)cc12 & 0.54461929 \\ \hline
        COC(=O)C1=C(C)N=C(C)C(=C(O)OC)C1C1=CCN(C(=O)Oc2ccccc2)C=C1 & 0.54189556 \\ \hline
        COC(OC)C1=C(C(=O)OCC=Cc2ccccc2)C(c2ccc(Cl)c(Cl)c2)C(C(=O)O)=CN1 & 0.54151 \\ \hline
        CCC1c2ccccc2CN1CNC(=O)c1cc(Cl)c(N)cc1OC & 0.53802759 \\ \hline
        CCOC(=O)c1cnc2c(ccc3ccccc32)c1Cl & 0.53708616 \\ \hline
        CCOC(=O)c1ccc(OCc2nc3ccccc3n(C)c2=O)cc1 & 0.53568323 \\ \hline
        CCOC(=O)C1=C(Nc2ccccc2)CC(C)(O)C(C(=O)OCC)C1c1ccc(Br)cc1 & 0.53555738 \\ \hline
        CCOC(=O)C1C(C(=O)OCC)C12C(=O)N(C)c1ccccc12 & 0.53329511 \\ \hline
        COC(=O)c1c(NC(=O)CCCOc2ccccc2)sc2c1CCC(C)C2 & 0.53229065 \\ \hline
        CCOc1ccccc1C1CC(=O)Nc2cc(OC)c(OC)cc21 & 0.53215208 \\ \hline
        CCOC(=O)C1=C(C)N(c2cccc(C(F)(F)F)c2)C(=O)N(C)C1c1ccc(C\#N)cc1C(=O)N(C)CCCOC & 0.52812079 \\ \hline
        CCOC(=O)c1c[nH]c2c(ccc3nc(Cl)cc(C)c32)c1=O & 0.52704628 \\ \hline
        CCOC(=O)c1cnc2ccc(OC)cc2c1NCCc1ccccc1 & 0.52660319 \\ \hline
    \end{tabular}
    \caption{Best 20 SMILES and their corresponding Values for Amlodipine MPO}
    \label{tab:best_20_smiles_amlodipine}
\end{table}

\subsection{Dataset BEST SMILES (Top 20 SMILES) in Perindopril MPO}
\begin{table}[H]
    \centering
    \scriptsize  % Adjust font size to make the table fit better
    \renewcommand{\arraystretch}{1.2}  % Adjust row height for better readability
    \begin{tabular}{|p{14.5cm}|p{2.5cm}|}  % Adjusted column widths
        \hline
        \textbf{Best 20 SMILES for Perindopril MPO from GUACAMOL} & \textbf{Best SMILES Value} \\ \hline
        CCCC(NC(=O)C(N)Cc1ccc(O)cc1)C(=O)N1CCCC1C(=O)NCC(=O)NC(Cc1ccccc1)C(=O)N1CCCC1C(=O)O & 0.48795004 \\ \hline
        CCOC(=O)C(Cc1ccc(O)cc1)NC(=O)C1(NC(=O)C(SC(=O)c2ccccn2)C(C)C)CCCC1 & 0.48760869 \\ \hline
        CCOC(=O)C(CNC(C)=O)c1cn(C(=O)OCC)c2ccccc12 & 0.47756693 \\ \hline
        CC(C)CC(C(=O)NC1CCCCC1)N(Cc1cccs1)C(=O)c1snc(C(N)=O)c1N & 0.47050403 \\ \hline
        CCOC(=O)C(C)(C)Oc1ccc(N(CC2CCCC2)C(=O)Nc2nccs2)cc1 & 0.46770717 \\ \hline
        COC(C)(C)C(O)C(=O)N1C(C(=O)NCc2cc(Cl)ccc2-n2cnnn2)CC2CC21 & 0.46589083 \\ \hline
        CCOCCOC(=O)Nc1cc2nc(C3CCCCC3)[nH]c2cc1N(C)C & 0.46164354 \\ \hline
        CCOC(=O)C(CCCCNC(=O)C(C)n1c([N+](=O)[O-])cnc1C)NC(=O)Cn1cc([N+](=O)[O-])nc1C & 0.4612656 \\ \hline
        CCCc1cc(=O)oc2c(C)c(OCC(=O)N3CCC(C(=O)O)CC3)ccc12 & 0.45976311 \\ \hline
        CC(C)C(O)CC(O)C(CC1CCCCC1)NC(=O)C(Cc1c[nH]cn1)NC(=O)c1ccc[nH]1 & 0.45841567 \\ \hline
        CC(=O)NC(Cc1ccccc1)C(=O)N1CCCC1C(=O)NC(CCCn1ccnc1)B1OC2CC3CC(C3(C)C)C2(C)O1 & 0.45812285 \\ \hline
        CCOC(=O)c1c(OC2CCCCC2C)cc(CCc2ccccc2)[nH]c1=O & 0.45607017 \\ \hline
        CC(C)C(=O)N1CCC(C(=O)NC(C(=O)NC(CCCCN)C(=O)OC(C)(C)C)C(C)c2c[nH]c3ccccc23)CC1 & 0.45584231 \\ \hline
        CC(C)(C)OC(=O)NCCCCC1NC(=O)C2Cc3c([nH]c4ccccc34)C(C3CCCCC3)N2C1=O & 0.45329841 \\ \hline
        CCOC(=O)N1CCC(NC(=O)C2CCN(Cc3nc(-c4ccc(CC)cc4)oc3C)CC2)CC1 & 0.45191299 \\ \hline
        COCC(C)n1c(SCC(=O)NC2CCCCC2)nc2ccccc2c1=O & 0.45083482 \\ \hline
        CCN(CC)S(=O)(=O)c1cccc(-c2nnc(SCC(=O)NC3CCCCC3C)n2N)c1 & 0.45056356 \\ \hline
        CCOC(=O)C=CC(=O)Nc1ccccc1CCCN1CCC23CCCCC2C1Cc1ccc(O)cc13 & 0.45022517 \\ \hline
        CCCCCOC(=O)N1CCN(C(=O)C(CCC(=O)O)NC(=O)c2cc(OCC3CCN(C)CC3)nc(-c3ccccc3)n2)CC1 & 0.44986771 \\ \hline
        CCOC(=O)C(Cc1ccccc1)NC(=O)C(C)(C)C(CC(C)C)NC(=O)c1ccc(C\#N)cc1 & 0.44881939 \\ \hline
    \end{tabular}
    \caption{Best 20 SMILES and their corresponding Values for Perindopril MPO}
    \label{tab:best_20_smiles_3}
\end{table}

%%%%%%%%%%%%%%%%%%%%%%%%%%%%%%%%%%%%%%%%%%%%%%%%%%%%%%%%%%%%%%%%%%%%%%%%%%%%%%%%

% INDEX?

\end{document}